\documentclass[10pt,journal,compsoc]{IEEEtran}
\ifCLASSOPTIONcompsoc
  \usepackage[nocompress]{cite}
\else
  \usepackage{cite}
\fi
\ifCLASSINFOpdf
   \usepackage[pdftex]{graphicx}
\else
\fi
\usepackage{amsmath}
\usepackage{amsfonts}
\usepackage{multirow}
\usepackage{amsthm}

\newcounter{definition}[section]
\newenvironment{definition}[1][]{\refstepcounter{definition}\par\medskip
   \textbf{Definition~\thedefinition. #1} \rmfamily}{\medskip}
\DeclareMathOperator*{\minimize}{minimize}
\DeclareMathOperator*{\maximize}{maximize}

\usepackage{mathtools}
\DeclarePairedDelimiter\floor{\lfloor}{\rfloor}

\usepackage{xcolor}
\usepackage{floatrow}
\usepackage{algorithmic}

\usepackage{array}

\ifCLASSOPTIONcompsoc
 \usepackage[caption=false,font=footnotesize,labelfont=sf,textfont=sf]{subfig}
\else
 \usepackage[caption=false,font=footnotesize]{subfig}
\fi
\usepackage{dblfloatfix}

\ifCLASSOPTIONcaptionsoff
 \usepackage[nomarkers]{endfloat}
\let\MYoriglatexcaption\caption
\renewcommand{\caption}[2][\relax]{\MYoriglatexcaption[#2]{#2}}
\fi
\usepackage{url}
\usepackage{hyperref}

\renewenvironment{IEEEbiography}[1]
  {\IEEEbiographynophoto{#1}}
  {\endIEEEbiographynophoto}

\begin{document}
%

\title{Resisting Adversarial Attacks in Deep Neural Networks using Diverse Decision Boundaries\vspace{-0.2cm}}
%
%
%
%

\author{Manaar~Alam, Shubhajit~Datta, Debdeep~Mukhopadhyay, Arijit~Mondal, and Partha~Pratim~Chakrabarti\vspace{-0.2cm}
\IEEEcompsocitemizethanks{\IEEEcompsocthanksitem M.~Alam, D.~Mukhopadhyay, and P.~P.~Chakrabarti are with the Department of Computer Science and Engineering, Indian Institute of Technology Kharagpur, India. E-mail: alam.manaar@gmail.com, debdeep.mukhopadhyay@gmail.com, ppchak@cse.iitkgp.ac.in\protect\\
\IEEEcompsocthanksitem S.~Datta and A.~Mondal are with the Centre of Excellence in Artificial Intelligence, Indian Institute of Technology, Kharagpur, India. E-mail: arijit@iitp.ac.in, shubhajitdatta1988@gmail.com\protect\\
\IEEEcompsocthanksitem P.~P.~Chakrabarti is jointly associated with the Centre of Excellence in Artificial Intelligence, Indian Institute of Technology, Kharagpur.
}
}

\IEEEtitleabstractindextext{%
\begin{abstract}
The security of deep learning (DL) systems is an extremely important field of study as they are being deployed in several applications due to their ever-improving performance to solve challenging tasks. Despite overwhelming promises, the deep learning systems are vulnerable to crafted adversarial examples, which may be imperceptible to the human eye, but can lead the model to misclassify. Protections against adversarial perturbations on ensemble-based techniques have either been shown to be vulnerable to stronger adversaries or shown to lack an end-to-end evaluation. In this paper, we attempt to develop a new ensemble-based solution that constructs defender models with diverse decision boundaries with respect to the original model. The ensemble of classifiers constructed by (1) transformation of the input by a method called {\em Split-and-Shuffle}, and (2) restricting the significant features by a method called {\em Contrast-Significant-Features} are shown to result in diverse gradients with respect to adversarial attacks, which reduces the chance of transferring adversarial examples from the original to the defender model targeting the same class. We present extensive experimentations using standard image classification datasets, namely MNIST, CIFAR-10 and CIFAR-100 against state-of-the-art adversarial attacks to demonstrate the robustness of the proposed ensemble-based defense. We also evaluate the robustness in the presence of a stronger adversary targeting all the models within the ensemble simultaneously. Results for the overall false positives and false negatives have been furnished to estimate the overall performance of the proposed methodology.\vspace{-0.2cm}
\end{abstract}

}

\maketitle

\IEEEdisplaynontitleabstractindextext

%
\IEEEpeerreviewmaketitle

\IEEEraisesectionheading{\section{Introduction}\label{sec:introduction}}

%
%
%
%

 

\IEEEPARstart{D}{eep} learning algorithms have seen rapid growth in recent years because of their unprecedented successes with near-human accuracies in a wide variety of challenging tasks starting from image classification~\cite{DBLP:conf/cvpr/SzegedyVISW16}, speech recognition~\cite{DBLP:conf/icml/AmodeiABCCCCCCD16}, natural language processing~\cite{DBLP:journals/corr/WuSCLNMKCGMKSJL16}, to self-driving cars~\cite{DBLP:journals/corr/BojarskiTDFFGJM16}. Deep learning algorithms have even shown to surpass human intelligence at games like Go~\cite{DBLP:journals/nature/SilverHMGSDSAPL16}. While deep learning algorithms are extremely efficient in solving complicated classification tasks, they are vulnerable to an adversary who aims to fool the classifier. Szegedy \textit{et~al.}~\cite{DBLP:journals/corr/SzegedyZSBEGF13} first demonstrated the existence of \textit{adversarial examples} in the image classification domain. They have shown that it is possible for an adversary to slightly perturb a valid example with a visually imperceptible noise to make the classifier alter its decision from the original class.

The widely-studied phenomenon of adversarial examples among the research community has produced several attack methodologies with varied complexity and efficient fooling strategy~\cite{DBLP:journals/corr/GoodfellowSS14,DBLP:conf/iclr/KurakinGB17a,DBLP:conf/iclr/MadryMSTV18,DBLP:conf/cvpr/Moosavi-Dezfooli16,DBLP:conf/ccs/PapernotMGJCS17}. A wide range of defenses against such attacks has been proposed in the literature, which generally fall into two categories. The first category improves the training of neural networks to make them less vulnerable to adversarial examples by training the networks with different kinds of adversarially perturbed training data~\cite{DBLP:conf/nips/BastaniILVNC16,DBLP:journals/corr/HuangXSS15,DBLP:journals/corr/JinDC15,DBLP:conf/cvpr/ZhengSLG16} or changing the training procedure like gradient masking, defensive distillation, etc.~\cite{DBLP:journals/corr/GuR14,DBLP:conf/sp/PapernotM0JS16,DBLP:conf/cvpr/RozsaRB16,DBLP:journals/corr/ShahamYN15}. However, developing such defenses has been shown to be extremely challenging, as demonstrated by Athalye \textit{et al.}~\cite{DBLP:conf/icml/AthalyeC018} and Carlini and Wagner~\cite{DBLP:conf/sp/Carlini017}. The authors demonstrated that these defenses are not generalized for all types of adversarial attacks but are constrained to specific classes. Moreover, the changes in training procedures provide a false sense of security. The second category aims to \textit{detect} adversarial examples by simply flagging them~\cite{DBLP:journals/corr/BhagojiCM17,DBLP:journals/corr/FeinmanCSG17,DBLP:journals/corr/GongWK17,DBLP:journals/corr/GrosseMP0M17,DBLP:conf/iclr/MetzenGFB17,DBLP:conf/iclr/HendrycksG17a,DBLP:conf/iccv/LiL17}. However, even detection of adversarial examples can be quite a complicated task, as shown by Carlini and Wagner~\cite{DBLP:conf/ccs/Carlini017}. The authors illustrated with several experimentations that these detection techniques could be efficiently bypassed by an intelligent adversary having partial or complete knowledge of their internal working procedure.

In this paper, we aspire to detect adversarial examples using ensembles of classifiers instead of a single model. The idea of using ensembles to increase the robustness of a classifier against adversarial examples has recently been explored in the research community. The primary motivation of using an ensemble-based defense is that if multiple neural network models with similar decision boundaries perform the same task, the \textit{transferability} of adversarial examples makes it easier for an adversary to deceive all the models simultaneously. However, it will be difficult for an adversary to deceive multiple models simultaneously if they have diverse decision boundaries. Strauss \textit{et~al.}~\cite{DBLP:journals/corr/abs-1709-03423} used various ad-hoc techniques such as different random initializations, different neural network structures, bagging the input data, adding Gaussian noise while training to create multiple diverse classifiers, and finally combining them as the primary ensemble to detect adversarial examples. Adam \textit{et~al.}~\cite{DBLP:journals/corr/abs-1808-06645} proposed a stochastic method to add Variational Autoencoders between layers as a noise removal operator for creating combinatorial ensembles to limit the transferability of adversarial attacks. Tram{\`{e}}r \textit{et al.}~\cite{DBLP:conf/iclr/TramerKPGBM18} proposed \textit{Ensemble Adversarial Training} that incorporates perturbed inputs transferred from other pre-trained models during adversarial training to decouple adversarial example generation from the parameters of the trained model. Grefenstette \textit{et al.}~\cite{DBLP:journals/corr/abs-1811-09300} demonstrated that ensembling two models and then adversarially training them performs better than single-model adversarial training and ensemble of two separately adversarially trained models. Kariyappa and Qureshi~\cite{DBLP:journals/corr/abs-1901-09981} proposed \textit{Diversity Training} of an ensemble of models with uncorrelated loss functions using \textit{Gradient Alignment Loss} metric to reduce the dimension of adversarial sub-space shared between different models. Pang \textit{et al.}~\cite{DBLP:conf/icml/PangXDCZ19} proposed \textit{Adaptive Diversity Promoting} regularizer to train an ensemble of neural networks that encourages the non-maximal predictions in each member in the ensemble to be mutually orthogonal, making it challenging to transfer adversarial examples among all the models in the ensemble.

In order to detect adversarial examples, these ensemble-based approaches operate either by training the ensembles with pre-computed adversarial examples or mutually interacting among the models in the ensemble. In this work, we propose a methodology to incorporate \emph{diversity} among the models in an ensemble by training each model independently and promoting \emph{lower-level features and relatively less important components} in the training dataset. An input to a classifier consists of multiple features, among which some are more significant for correct classification, and some are less significant. In this work, we concentrate on the features which are less significant and term those as lower-level features. We propose two approaches for promoting the lower-level features, each having a different notion of these lower-level features. In the first approach, we reduce the correlation among significant features to promote features that are not highly correlated using suitable input transformation. In the second approach, we restrict the influence of significant features in a classifier to promote features that are not significant. The diversity among the models trained in such a way helps to degenerate the transferability of adversarial examples, thereby increasing the robustness of the ensemble. Also, we do not use pre-computed adversarial examples to train the models, which helps us proposing a defense by not restricting it to a particular class of adversarial attacks. Moreover, He \textit{et al.}~\cite{DBLP:conf/woot/HeWCCS17} showed that an adversary could evade an ensemble of weak defenses by targeting all the models within the ensemble simultaneously. We also evaluate our proposed methodology in a similar strong attack scenario.\vspace{-0.2cm}

\subsection*{Motivation behind the Proposed Approach}

\begin{figure}[!t]
\centering
\subfloat[]{\includegraphics[width=0.14\linewidth]{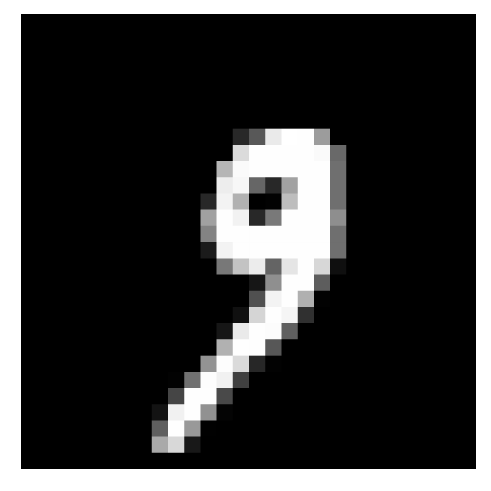}%
\label{mfig:original_image}}
\subfloat[]{\includegraphics[width=0.14\linewidth]{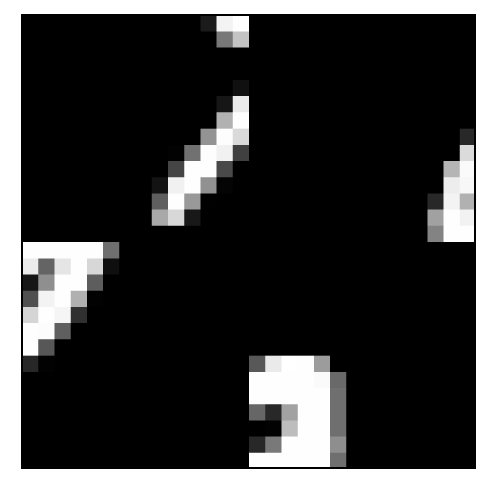}%
\label{mfig:transformed_image}}
\subfloat[]{\includegraphics[width=0.14\linewidth]{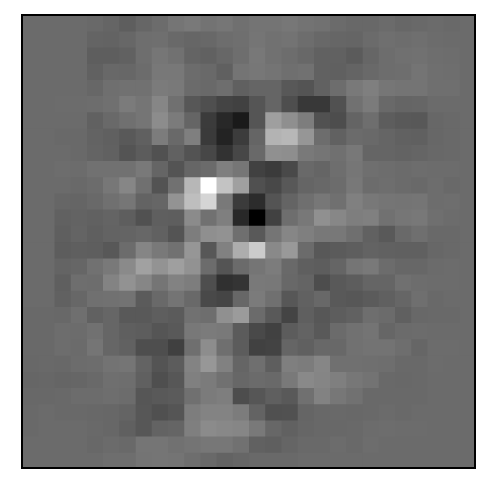}%
\label{mfig:original_gradient}}
\subfloat[]{\includegraphics[width=0.14\linewidth]{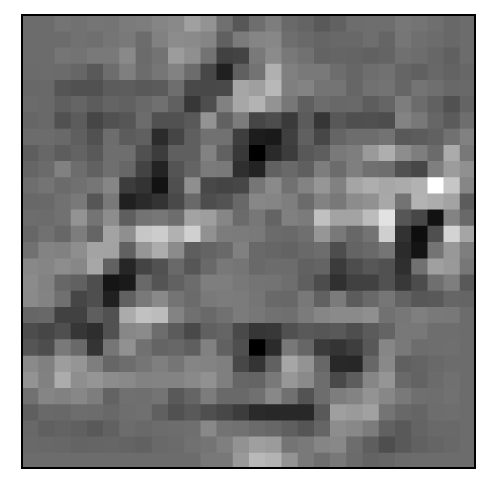}%
\label{mfig:transformed_gradient}}
\subfloat[]{\includegraphics[width=0.14\linewidth]{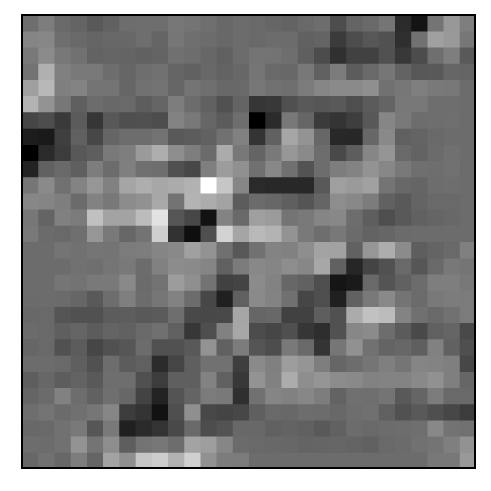}%
\label{mfig:transformed_gradient_inverse}}\vspace{-0.2cm}

\subfloat[]{\includegraphics[width=0.14\linewidth]{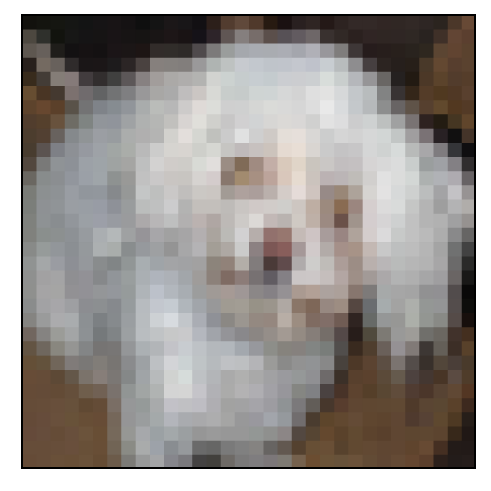}%
\label{fig:original_image}}
\subfloat[]{\includegraphics[width=0.14\linewidth]{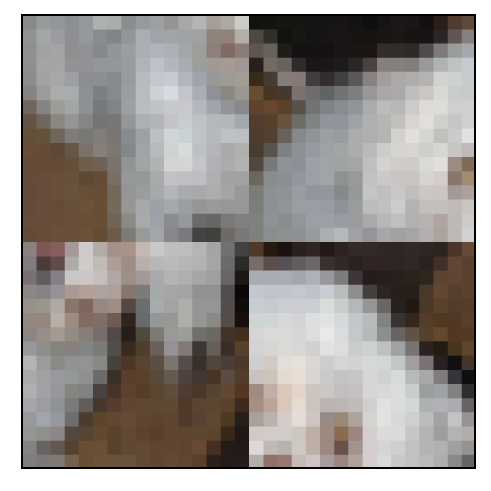}%
\label{fig:transformed_image}}
\subfloat[]{\includegraphics[width=0.14\linewidth]{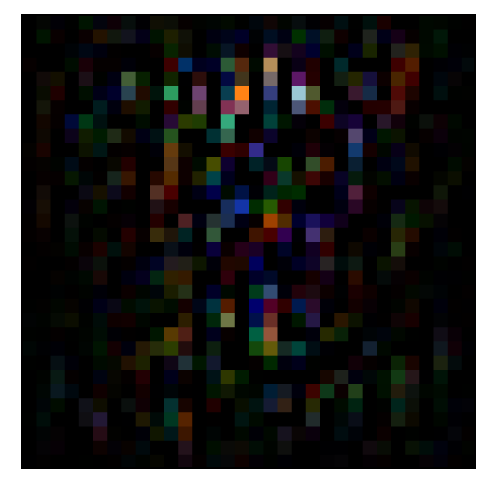}%
\label{fig:original_gradient}}
\subfloat[]{\includegraphics[width=0.14\linewidth]{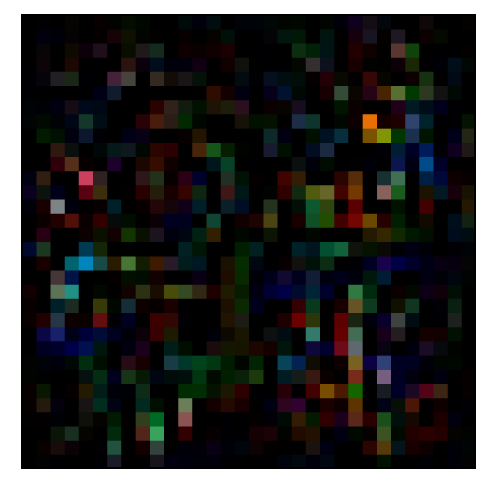}%
\label{fig:transformed_gradient}}
\subfloat[]{\includegraphics[width=0.14\linewidth]{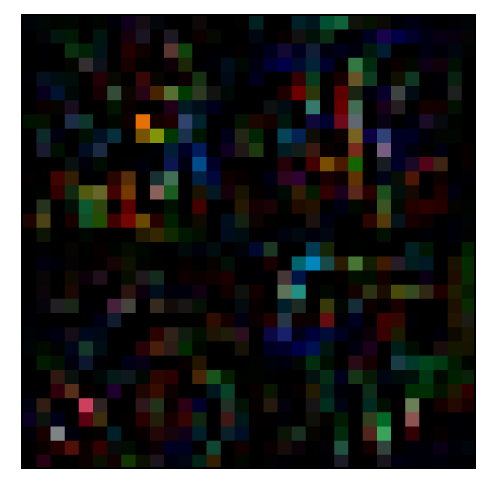}\label{fig:transformed_gradient_inverse}}\vspace{-0.2cm}

\subfloat[]{\includegraphics[width=0.14\linewidth]{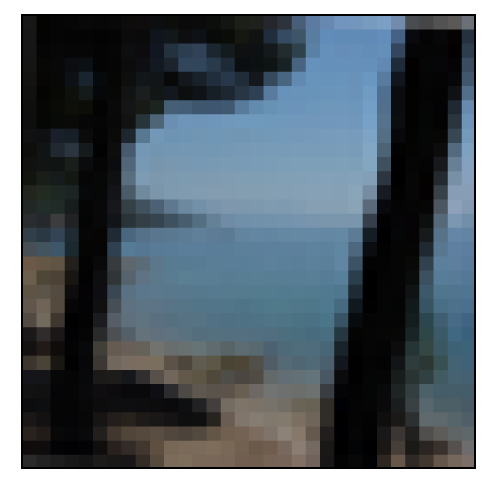}%
\label{cfig:original_image}}
\subfloat[]{\includegraphics[width=0.14\linewidth]{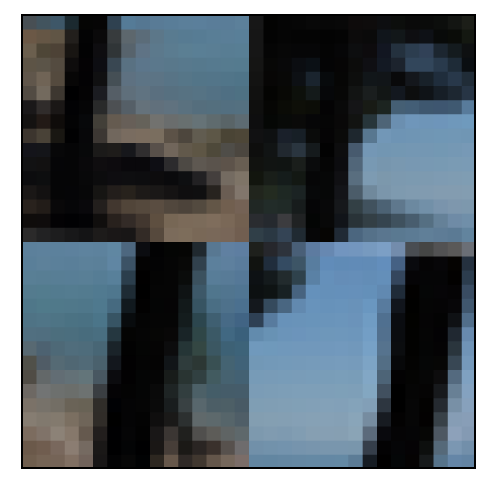}%
\label{cfig:transformed_image}}
\subfloat[]{\includegraphics[width=0.14\linewidth]{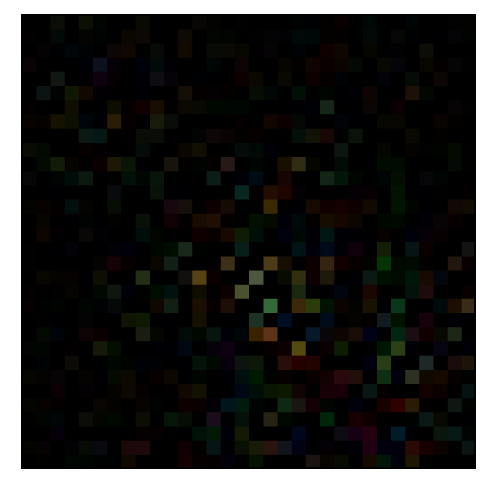}%
\label{cfig:original_gradient}}
\subfloat[]{\includegraphics[width=0.14\linewidth]{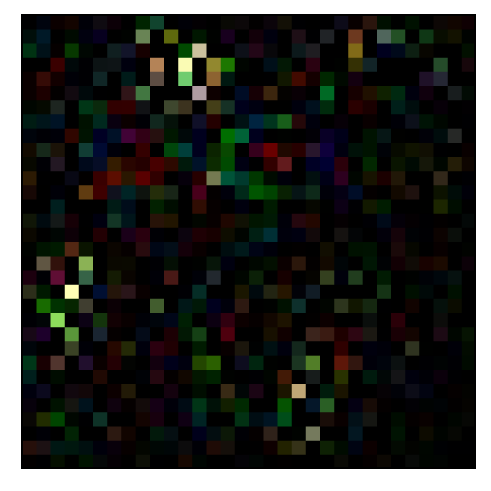}%
\label{cfig:transformed_gradient}}
\subfloat[]{\includegraphics[width=0.14\linewidth]{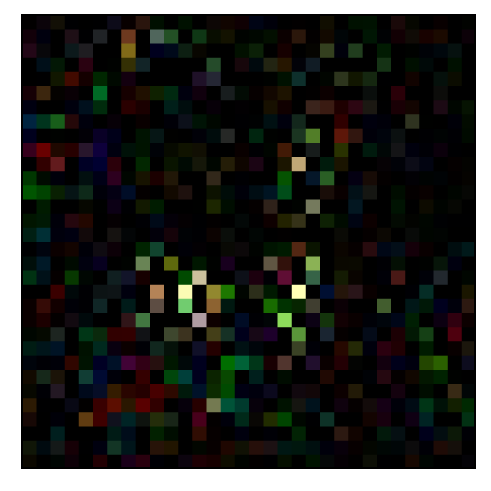}\label{cfig:transformed_gradient_inverse}}
\caption{Motivation for using lower level features to train an image classifier for increasing resiliency against adversarial attacks. Case studies using one MNIST (top row), one CIFAR-10 (middle row), and one CIFAR-100 (bottom row) image. (a,~f,~k) Original Image, (b,~g,~l) Transformed Image, (c,~h,~m) Gradients in original domain, (d,~i,~n) Gradients in transformed domain, (e,~j,~o) Transformed gradients in original domain. The gradient of loss with respect to the MNIST image is computed for a target image class of \texttt{four}, the same is computed for CIFAR-10 image for a target image class of \texttt{cat} and for CIFAR-100 image for a target image class of \texttt{cloud}. The transformation used is simply splitting the image into four equal blocks and applying a random shuffle to the blocks\vspace{-0.3cm}}
\label{fig:motivation_mnist}
\end{figure}

In this section, we outline the intuition behind developing defender models that aid in detecting adversarial examples in the original model. In classical ensemble-based detection, the individual models are either distinct machine learning models or operate on different feature sets. We aim to construct secondary models with diverse decision boundaries from the original model. The idea can be explained by Fig.~\ref{fig:motivation_mnist}, which shows image classifiers being built for standard image classification datasets, namely MNIST, CIFAR-10 and CIFAR-100. Without loss of generality, Fig.~\ref{mfig:original_image} shows a figure of `9' from MNIST being considered as an input to the original classifier. We build a secondary classifier for a {\em different} MNIST, which is {\em diverse} from the first model with respect to its decision boundary. The difference is illustrated by splitting the input image into multiple blocks and performing an arbitrary shuffling for the same, which is shown in Fig.~\ref{mfig:transformed_image}. If these two images are fed as an input to two models for classification, it turns out that even if their decisions are the same, their decision boundaries are quite contrasting. This can be observed in Fig.~\ref{mfig:original_gradient} and Fig.~\ref{mfig:transformed_gradient}, which show the gradients of loss computed with respect to both the original and transformed images from their respective models targeting the class `4'. The gradient of the loss with respect to the transformed image in the original image domain is shown in Fig.~\ref{mfig:transformed_gradient_inverse}. It may be observed to be significantly distinct from the gradient of the loss with respect to the original image. This observation implies that it becomes exceedingly difficult for an attacker to introduce perturbations in the original image classified into the same target class (say `4' in this case) by the two models with {\em different} decision boundaries and hence distinct gradients. This principle of operating on models working with varying decision boundaries is expected to lead to increased {\em strength} of the overall scheme against adversarial attacks. Likewise, Fig.~\ref{fig:original_image} and Fig.~\ref{fig:transformed_image} provide a CIFAR-10 image of a `dog' and its transformed counterpart respectively. The Fig.~\ref{fig:original_gradient} and Fig.~\ref{fig:transformed_gradient} show the gradients of loss for a target class of `cat' in an adversarial attack respectively. Fig.~\ref{fig:transformed_gradient_inverse} provides the gradient of loss with respect to the transformed image in the original image domain. Comparing Fig.~\ref{fig:original_gradient} and Fig.~\ref{fig:transformed_gradient_inverse}, one can observe the differences to comprehend why it is harder for both the models to result in the same target class under an adversarial perturbation. Similar observations are shown from Fig.~\ref{cfig:original_image} to Fig.~\ref{cfig:transformed_gradient_inverse} for a CIFAR-100 image of a `sea'. The gradient of loss is computed for a target class of `cloud'.\vspace{-0.2cm}

\subsection*{Our Contributions}
The primary contributions of this paper are as follows:\vspace{-0.1cm}
\begin{enumerate}
    \item We develop a methodology for detecting adversarial perturbations using an ensemble of classifiers. The classifiers are ensured to have diversity in the decision boundaries. 
    \item We propose two methods for designing such varying decision boundaries, namely \textbf{(1)} Transforming the inputs by a technique we call {\em Split-and-Shuffle}, and \textbf{(2)} Restricting the significant features by a method called {\em Contrast-Significant-Features}.
    \item We evaluated the robustness of the proposed ensemble-based methodology on with extensive experimentation on benchmark datasets and its effect on overall false positives and false negatives. The {\em strength} of the method, banking on the idea of {\em differences}, is achieved by evaluating it against several state-of-the-art adversarial attacks and those that target both the original model and the detector model simultaneously.\vspace{-0.1cm}
\end{enumerate}

The paper is organized as follows: Section~\ref{sec:prelim} presents a preliminary discussion on the generation of adversarial examples using different methodologies. Section~\ref{sec:threat_model} discusses the threat model considered in this paper, followed by an overview of the proposed approach in Section~\ref{sec:overview}. Section~\ref{sec:input_transformation} and Section~\ref{sec:orthogonal-feature} present two techniques for feature prioritization by input transformation and diverse feature selection, respectively. Section~\ref{sec:results} shows detailed experimental results to evaluate the robustness of the proposed methodology. Finally, we conclude the work in Section~\ref{sec:conclusion} with a scope of future research direction.\vspace{-0.2cm}

\section{Preliminaries on Adversarial Example Generation}\label{sec:prelim}
Let us consider a data point $x$, classified into class $\mathcal{C}_i$ by a classifier $\mathcal{F}$. An adversarial attack tries to add visually imperceptible perturbation to $x$ and creates a new data point $x_{adv}$ such that $\mathcal{F}$ misclassifies $x_{adv}$ into another class $\mathcal{C}_j$ other than $\mathcal{C}_i$. The definition of each symbol used throughout this section are mentioned in Table~\ref{table:sym_def}.\vspace{-0.1cm}

\begin{table}[!b]
\centering
\caption{Definition of symbols used for preliminaries on adversarial attacks}\label{table:sym_def}
\resizebox{\linewidth}{!}{
\begin{tabular}{|c|l|}
\hline
\textbf{Symbol} & \multicolumn{1}{c|}{\textbf{Definition}} \\ \hline
$d$ & number of features in a datapoint \\ \hline
$x$ & datapoint in $\mathbb{R}^d$ \\ \hline
$x_{init}$ & initial $x$ on which adversarial attack is performed \\ \hline
$r$ & adversarially generated perturbation in $\mathbb{R}^d$ \\ \hline
$x^{(i)}$ & datapoint after $i^{th}$ iteration of adversarial attack\\ \hline
$x_{adv}$ & adversarially perturbed datapoint \\ \hline
$g_i(\cdot)$ & discriminant function for class $\mathcal{C}_i$ \\ \hline
$\nabla_x$ & gradient of a function with respect to $x$ \\ \hline
$J(x, w)$ & loss function for $x$ of a classifier with parameters $w$ \\ \hline
$\eta$ & parameter controlling magnitude of adversarial attack \\ \hline
$\alpha$ & step size of each iteration of adversarial attack \\ \hline
$clip_{\eta}(\cdot)$ & function to restrict adversarial examples within $\eta$-ball\\ \hline
$\lambda$ & parameter controlling emphasis of adversarial perturbation \\ \hline
$z(\cdot)$ & output of logit layer of a classifier \\ \hline
$\kappa$ & parameter controlling confidence of fooling a classifier \\ \hline
\end{tabular}}
\end{table}

\begin{definition}
Let $x \in \mathbb{R}^d$ is a data point classified into class $\mathcal{C}_i$. An \textbf{\textit{adversarial attack}} is a linear mapping that adds perturbation $r \in \mathbb{R}^d$ to $x$ creating $x_{adv} = x + r$ such that $x_{adv}$ is misclassified into a class $\mathcal{C}_j \neq \mathcal{C}_i$.
\end{definition}\vspace{-0.1cm}

Let us consider a multi-class scenario where we have $m$ classes $\mathcal{C}_1, \mathcal{C}_2 \dots, \mathcal{C}_m$. Let us also assume that the decision boundaries of these $m$ classes are specified by $m$ discriminant functions $g_1(\cdot), g_2(\cdot), \dots, g_m(\cdot)$. If a data point $x$ is classified into a class $\mathcal{C}_k$, then all the discriminant functions should satisfy the following condition\vspace{-0.1cm}
\begin{equation*}
    g_k(x) > g_l(x) \text{\hspace{0.4cm}} \forall l \neq k
\end{equation*}
Hence the class $\mathcal{C}_k$ will have a discriminant value $g_k(x)$ greater than all other classes
in the classifier. Since, $x_{adv}$ is misclassified into $\mathcal{C}_j$, it will satisfy the following inequality\vspace{-0.1cm}
\begin{equation*}
    g_j(x_{adv}) > \max_{l \neq j}\{g_l(x_{adv})\} \Leftrightarrow \max_{l \neq j}\{g_l(x_{adv})\} - g_j(x_{adv}) < 0
\end{equation*}
Thus, the goal of any adversarial attack is to find $x_{adv}$ such that the above inequality holds for \textit{any} $j$ in case of \textit{untargeted attacks} and for a \textit{fixed} $j$ in case of \textit{targeted attacks}\footnote{In untargeted attacks, goal of an adversary is to misclassify a data point to any class different from the original class. However, in targeted attacks, adversary tries to misclassify a data point to a particular class.}.

The adversarial attacks discussed in the literature can be broadly classified into the following three definitions based on the generation of adversarial perturbations~\cite{adversarial_definitions}.\vspace{-0.1cm}

\begin{definition}
The \textbf{\textit{minimum norm attack}} finds a perturbed data point $x$ from an initial data point $x_{init}$ by solving the optimization\vspace{-0.1cm}
\begin{align*}
    \displaystyle{\minimize_{x}} & \text{\hspace{0.3cm}} \|x - x_{init} \| \\
    \text{such that} & \text{\hspace{0.3cm}} \max_{l \neq j}\{g_l(x)\} - g_j(x) < 0
\end{align*}
where $\| \cdot \|$ can be any \textit{norm} specified by the attacker.
\end{definition}\vspace{-0.1cm}

The goal of the minimum norm attack is to minimize the magnitude of perturbation while ensuring the new data $x$ is misclassified into $\mathcal{C}_j$ from $\mathcal{C}_i$. The \textit{DeepFool} attack~\cite{DBLP:conf/cvpr/Moosavi-Dezfooli16} follows the principle of minimum norm attack. The iterative approach to obtain adversarial example with \textit{DeepFool} attack using $L_2$-norm for a two-class problem can be written as\vspace{-0.1cm}
\begin{equation*}
    x^{(k+1)} = x^{(k)} - \left(\frac{g(x^{(k)})}{\|\nabla_x g(x^{(k)})\|_2}\right)\cdot \nabla_x g(x^{(k)})
\end{equation*}
where $g(\cdot) = g_i(\cdot) - g_j(\cdot)$ is the discriminant function for the decision boundary between class $\mathcal{C}_i$ and $\mathcal{C}_j$, $x^{(0)} = x_{init}$, and $\nabla_x$ is the gradient of a function with respect to its input. The exact formulation used for multi-class problems is rather sophisticated. We request interested readers to refer to the original work.\vspace{-0.1cm}

\begin{definition}
The \textbf{\textit{maximum loss attack}} finds a perturbed data point $x$ from an initial data point $x_{init}$ by solving the optimization\vspace{-0.1cm}
\begin{align*}
    \displaystyle{\maximize_{x}} & \text{\hspace{0.3cm}} g_j(x) - \max_{l \neq j}\{g_l(x)\} \\
    \text{such that} & \text{\hspace{0.3cm}} \|x - x_{init} \| \leq \eta
\end{align*}
where $\| \cdot \|$ can be any \textit{norm} specified by the attacker, and $\eta > 0$ is a parameter controlling the magnitude of attack.
\end{definition}\vspace{-0.1cm}

The goal of maximum loss attack is to find new data point $x$ such that the objective function $g_j(x) - \max_{l \neq j}\{g_l(x)\}$ is maximized while ensuring that the magnitude of the perturbation is upper bounded by $\eta$. The \textit{FGSM} attack~\cite{DBLP:journals/corr/GoodfellowSS14}, BIM attack~\cite{DBLP:conf/iclr/KurakinGB17a}, and PGD attack~\cite{DBLP:conf/iclr/MadryMSTV18} follow the principle of maximum loss attack. The one-shot approach to obtain adversarial examples with \textit{FGSM} attack using $L_{\infty}$-norm can be written as\vspace{-0.1cm}
\begin{equation*}
    x = x_{init} + \eta \cdot sign(\nabla_x J(x_{init}, w))
\end{equation*}
where $J(x, w)$ is a loss function evaluating the amount of loss incurred by a classifier, parameterized by $w$, while classifying the data point $x$. The iterative approach to obtain adversarial examples with \textit{BIM} attack using $L_{\infty}$-norm can be written as\vspace{-0.1cm}
\begin{equation*}
    x^{(k+1)} = x^{(k)} + clip_{\eta}(\alpha \cdot sign(\nabla_x J(x^{(k)}, w)))
\end{equation*}
where $\alpha$ is a small step size and $clip_{\eta}(\cdot)$ is used to generate adversarial examples within $\eta$-ball of the original image $x^{(0)} = x_{init}$. The PGD attack is a stronger variant of the BIM attack. The PGD attack generates adversarial examples in the same way as the BIM attack; however, instead of starting from the data sample, it randomly starts within the $L_{\infty}$ ball of a data sample.\vspace{-0.1cm}

\begin{definition}
The \textbf{\textit{regularization-based attack}} finds a perturbed data point $x$ from an initial data point $x_{init}$ by solving the optimization\vspace{-0.1cm}
\begin{equation*}
    \displaystyle{\minimize_{x}} \text{\hspace{0.3cm}} \|x - x_{init} \| + \lambda \cdot (\max_{l \neq j}\{g_l(x)\} - g_j(x))
\end{equation*}
where $\| \cdot \|$ can be any \textit{norm} specified by the attacker, and $\lambda > 0$ is a regularization parameter controlling the emphasis of two terms.
\end{definition}\vspace{-0.1cm}

The CW attack~\cite{DBLP:conf/sp/Carlini017} follows the principle of regularization-based attack. The adversarial examples using iterative approach of \textit{CW} attack with $L_{2}$-norm can be obtained by solving the following optimization problem\vspace{-0.1cm}
\begin{equation*}
    \displaystyle{\minimize_{x}} \text{\hspace{0.3cm}} \|x - x_{init} \|_{2}^{2} + \lambda \cdot l(x)
\end{equation*}
where $l(x) = max(max\{z(x)_i: i \neq t\} - z(x)_t, - \kappa)$, $z(\cdot)$ is the output of the \textit{logit} layer of the classifier, $x_{init}$ belongs to class $\mathcal{C}_t$, and $\kappa$ is a parameter capable of fooling a classifier with a high confidence rate.

It can be shown that the three optimizations defined by minimum norm attack, maximum loss attack, and regularization-based attack are equivalent in the sense that the solutions are identical for appropriately chosen $\eta$ and $\lambda$.\vspace{-0.3cm}

\section{Threat Model}\label{sec:threat_model}
We consider the following two threat models in this paper while generating the adversarial examples, which is in line with the works presented by Carlini \textit{et al.}~\cite{DBLP:conf/ccs/Carlini017} and Biggio \textit{et~al.}~\cite{DBLP:conf/pkdd/BiggioCMNSLGR13}.
\begin{itemize}
    \item \emph{Zero Knowledge Adversary $(\mathcal{A}_\mathcal{Z})$:} The adversary $\mathcal{A}_\mathcal{Z}$ is unaware that a defense $\mathcal{M}_\mathcal{D}$ is in place for the unsecured neural network model $\mathcal{M}_\mathcal{U}$. We term $\mathcal{A}_\mathcal{Z}$ as a \textit{black-box} adversary\footnote{However, the authors in~\cite{DBLP:conf/ccs/Carlini017} have not used any categorization for this scenario.}. The adversary $\mathcal{A}_\mathcal{Z}$ generates adversarial examples for $\mathcal{M}_\mathcal{U}$. The detector $\mathcal{M}_\mathcal{D}$ is considered to be successful if it can detect the adversarial examples.
    \item \emph{Perfect Knowledge Adversary $(\mathcal{A}_\mathcal{P})$:} The adversary $\mathcal{A}_\mathcal{P}$ is a stronger adversary than $\mathcal{A}_\mathcal{Z}$ who is aware that the neural network model $\mathcal{M}_\mathcal{U}$ is secured with a given detection scheme $\mathcal{M}_\mathcal{D}$. The adversary $\mathcal{A}_\mathcal{P}$ also knows the parameters used by $\mathcal{M}_\mathcal{D}$, and can generate adversarial examples considering both $\mathcal{M}_\mathcal{U}$ and $\mathcal{M}_\mathcal{D}$.  We term $\mathcal{A}_\mathcal{P}$ as a \textit{white-box} adversary. The adversary $\mathcal{A}_\mathcal{P}$ is considered to be successful if it can evade both $\mathcal{M}_\mathcal{U}$ and $\mathcal{M}_\mathcal{D}$ simultaneously.\vspace{-0.3cm}
\end{itemize}

\section{Overview of the Proposed Methodology}\label{sec:overview}
In this section, we provide a brief description of the proposed methodology used in this paper to detect adversarially perturbed examples using an ensemble of classifiers. We use two classifiers in the ensemble -- \textbf{(1)} Unprotected model $\mathcal{M}_\mathcal{U}$: trained with the original dataset $\mathcal{D}$, and \textbf{(2)} Detector model $\mathcal{M}_\mathcal{D}$: trained with the same dataset $\mathcal{D}$ but prioritizing importance to lower-level features. We use two methods to train $\mathcal{M}_\mathcal{D}$ with lower-level feature prioritization\vspace{-0.1cm}
\begin{itemize}
    \item \textit{Transforming the inputs}: We use a \textit{split-and-shuffle} transformation for each image in the dataset $\mathcal{D}$. The transformation splits an image into multiple segments and randomly shuffles all the segments to remove the spatial correlation among the lower-level features that existed in the original image. We describe the transformation and the combined robustness of $\mathcal{M}_\mathcal{U}$ and $\mathcal{M}_\mathcal{D}$ against adversarial examples in detail in Section~\ref{sec:input_transformation}.
    \item \textit{Restricting significant features}: We trained a new model with the same architecture of the unprotected model, but by restricting the important features of the unprotected model. Our target is to design a new model such a way that the significant features of the first
model should not be significant in second model. By, doing so we can establish the diversity between the models. We have detailed the methods in Section~\ref{sec:orthogonal-feature}.\vspace{-0.1cm}
\end{itemize}

\begin{figure}[!t]
\centering
\includegraphics[width=0.8\linewidth]{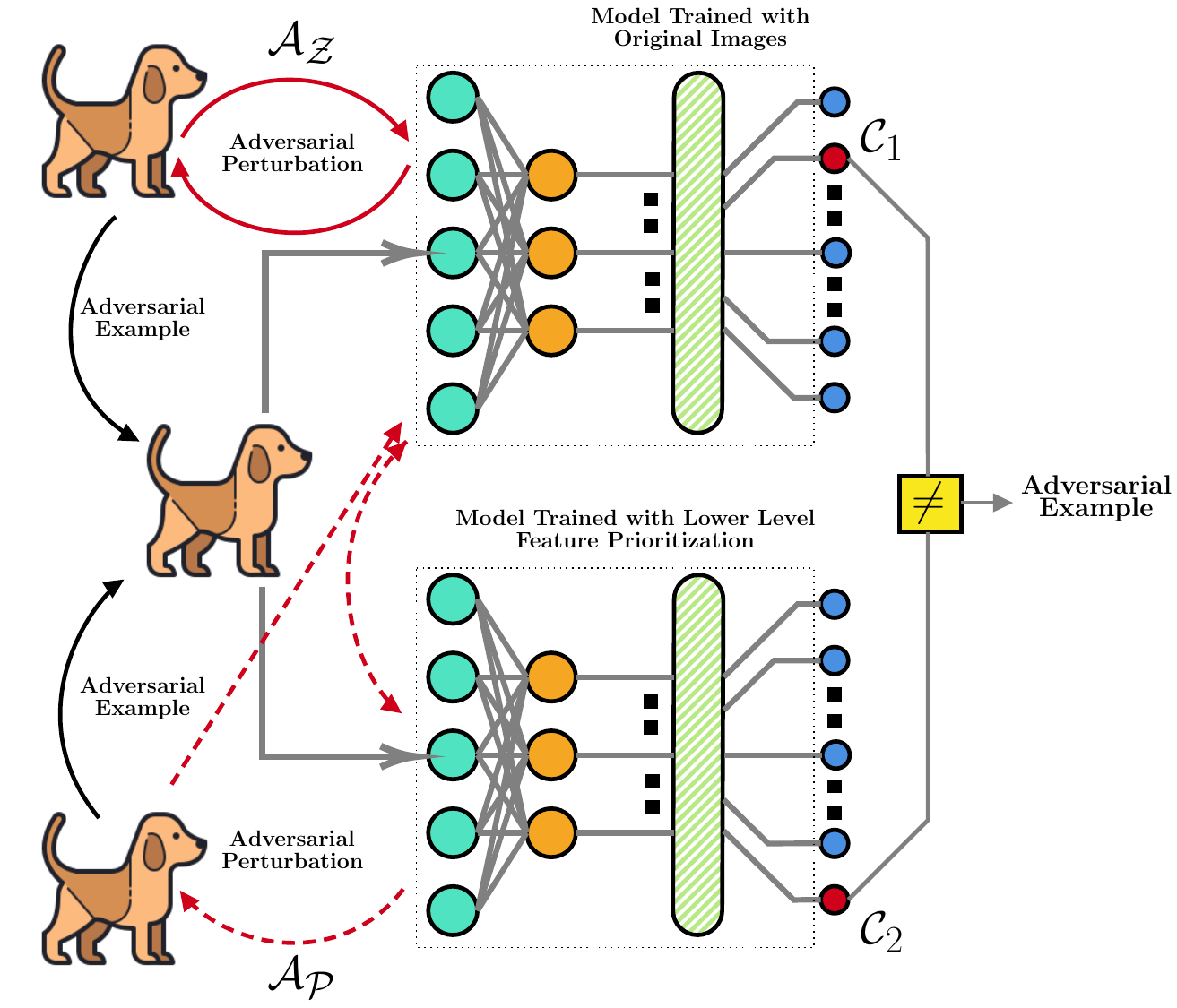}\vspace{-0.1cm}
\caption{\textit{Overview of the Proposed Methodology:} The adversaries $\mathcal{A}_{\mathcal{Z}}$ and $\mathcal{A}_{\mathcal{P}}$ can generate adversarial examples from a given input image. The perturbed image is detected as adversarial example when it is classified into different classes on both the models or it is visually perceptible to the human eye\vspace{-0.3cm}}
\label{fig:overview}
\end{figure}

The primary argument behind the proposed detection methodology is that $\mathcal{M}_\mathcal{U}$ and $\mathcal{M}_\mathcal{D}$ will have dissimilar decision boundaries but not significantly different accuracies, as the lower-level features are not hindered. Hence, a genuine example classified as class $\mathcal{C}_i$ in $\mathcal{M}_\mathcal{U}$ will also be classified as $\mathcal{C}_i$ in $\mathcal{M}_\mathcal{D}$. Consequently, because of the diversity in decision boundaries between $\mathcal{M}_\mathcal{U}$ and $\mathcal{M}_\mathcal{D}$, the adversarial examples generated by a \textit{zero knowledge adversary} $(\mathcal{A}_\mathcal{Z})$ based on $\mathcal{M}_\mathcal{U}$ will have a different impact on $\mathcal{M}_\mathcal{D}$, i.e., the \textit{transferability} of adversarial examples will be challenging. The ensemble detects an adversarial example when it produces two different classes in both the models. The overview of the methodology is presented in Fig.~\ref{fig:overview}. Moreover, a \textit{perfect knowledge adversary} $(\mathcal{A}_{\mathcal{P}})$ can generate adversarial examples based on both $\mathcal{M}_\mathcal{U}$ and $\mathcal{M}_\mathcal{D}$. In this scenario, there is a high chance that $\mathcal{A}_{\mathcal{P}}$ will generate adversarial examples that can produce the same misclassification in both the models. However, the input image perturbation will be significantly higher as the dissimilar perturbations from both the models need to be added to the input image, making it visually perceptible to the human eye. We used the $L_2$-norm of the difference of an adversarially perturbed image with its original version as a measure of the amount of perturbation. Higher values of $L_2$-norm signifies the perturbation is perceptible to the human eye. We evaluated our proposed defense, considering both the adversaries, and presented the results in Section~\ref{sec:results}. We want to stress that prioritizing lower-level features will impact the accuracy of model $\mathcal{M}_\mathcal{D}$ depending on the level of prioritization. Hence, to maintain the original accuracy of $\mathcal{M}_\mathcal{U}$, we use $\mathcal{M}_\mathcal{D}$ only for detecting adversarial examples and report accuracy of ensemble from $\mathcal{M}_\mathcal{U}$.

\textit{What is the success rate of an adversary?} As discussed, we have an ensemble $\mathcal{E}$ of $2$ classification models, i.e., $\mathcal{E} = \{\mathcal{M}_\mathcal{U}, \mathcal{M}_\mathcal{D}\}$, where both are independently trained classifiers. An adversary is considered successful if she can fool both the models in $\mathcal{E}$ in the same fashion. If the adversary has a set of adversarial examples, the success rate for the adversary is defined as the portion of examples yielding the same incorrect misclassification from both the models in $\mathcal{E}$. Formally, let us assume a test set $\mathcal{T}$ of inputs $\{x_1, x_2, \dots, x_t\}$ with respective ground truth labels as $\{y_1, y_2, \dots, y_t\}$. Both the models in the ensemble $\mathcal{E}$ correctly classify all $x_i$'s, i.e., $\mathcal{M}_\mathcal{U}(x_i) = y_i$ and $\mathcal{M}_\mathcal{D}(x_i) = y_i$ for all $i = 1\dots t$. An adversary takes an input $x_i \in \mathcal{T}$ and perturbs them to create adversarial examples $x^{*}_i$. The attack success rate with respect to the ensemble $\mathcal{E}$ and the test set $\mathcal{T}$ is defined as\vspace{-0.1cm}
\begin{equation*}
    \mathcal{S}(\mathcal{E}, \mathcal{T}) = \frac{|\{x_i \in \mathcal{T}: \mathcal{M}_\mathcal{U}(x^*_i) = \mathcal{M}_\mathcal{D}(x^*_i) \neq y_i\}|}{|\mathcal{T}|}
\end{equation*}

\section{Building Detector Model $\mathcal{M}_{\mathcal{D}}$ with Input Transformation}\label{sec:input_transformation}
In this section, we provide a detailed discussion on training a detector model while prioritizing the lower-level features by applying a transformation on the input dataset. First, we explain the transformation used in the proposed method, followed by a formal approach to demonstrate its robustness against adversarial examples.\vspace{-0.2cm}

\begin{figure}[!t]
\centering
\includegraphics[width=0.9\linewidth]{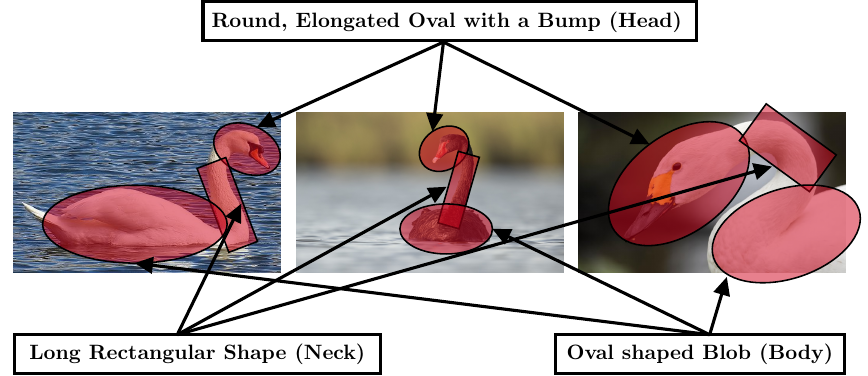}
\caption{A swan has specific spatial characteristic features that can be used for its recognition -- an oval head followed by a rectangular neck followed by an oval body\vspace{-0.3cm}}
\label{fig:spatial_correlation}
\end{figure}

\subsection{Notion behind the Input Transformation}
A classifier is trained in an image classification problem by leveraging the spatial correlation among the image features. For example, a swan can be recognized by detecting its spatial characteristics, like an oval head followed by a rectangular neck followed by an oval body, as shown in Fig.~\ref{fig:spatial_correlation}. In the proposed method, we introduce a \textit{split-and-shuffle} transformation to remove such correlation from the input images. Applying such a transformation provides priority to the lower-level features (like oval head, rectangular neck, and oval body) instead of their order of appearances while training a classifier. Also, the transformation does not depend on any particular content of any image. The disarrangements of spatial features will be different for images belonging to different categories. This approach aims to obtain a classifier with dissimilar decision boundary without adversely affecting the accuracy. The accuracy of the transformed images will be close to the original one as the features are kept unimpaired. However, removing the correlation will impact accuracy, which we have observed is not significant. We have used \textit{Central Kernel Alignment} (CKA) analysis proposed by Kornblith \textit{et al.}~\cite{DBLP:conf/icml/Kornblith0LH19} as a metric to measure the similarity between decision boundaries learned by both the models. The results are discussed later in Section~\ref{sec:results}.

Training the detector model $\mathcal{M}_{\mathcal{D}}$ with \textit{split-and-shuffle} transformation on the input data provides two advantages\vspace{-0.1cm}
\begin{itemize}
    \item The \textit{dissimilar decision boundaries} ensure that adversarial examples generated on a model trained with correlated features $(\mathcal{M}_{\mathcal{U}})$ will have a different impact on the model trained by prioritizing the lower-level features $(\mathcal{M}_{\mathcal{D}})$. The intuition behind the argument is that while generating adversarial examples in $\mathcal{M}_{\mathcal{U}}$, the attack algorithms will also consider the spatial correlation among the features. Hence, such adversarial examples will impact $\mathcal{M}_{\mathcal{D}}$ differently where the spatial correlation is not considered, i.e., the misclassified classes in both models will be different if there is a misclassification in $\mathcal{M}_{\mathcal{D}}$ at all.
    \item The \textit{similar accuracies} in both $\mathcal{M}_{\mathcal{U}}$ and $\mathcal{M}_{\mathcal{D}}$ ensure that the false positive in the detection remains low. For a significantly lower accuracy in $\mathcal{M}_{\mathcal{D}}$, it would not have been apparent whether the misclassification is due to the adversarial perturbation or improper training error.\vspace{-0.1cm}
\end{itemize}

Next we discuss in details the \textit{split-and-shuffle} transformation used in this paper.\vspace{-0.2cm}

\subsection{Split-and-Shuffle Transformation}\label{sec:split_and_transform}
In this section, we discuss two different types of split-and-shuffle transformations -- \textbf{(i)} Non-overlapping Transformation and \textbf{(ii)} Overlapping Transformation. The details of both these transformations are discussed as follows:\vspace{-0.2cm}

\subsubsection{Non-overlapping Transformation}
Let us consider an image $\mathcal{I}$ of dimension $m \times m$. The \textit{non-overlapping} \textit{split-and-shuffle} transformation divides $\mathcal{I}$ into equal-sized non-overlapping segments and shuffles them with a random permutation. However, the order of the shuffle is fixed for all images in a dataset. Based on the number of equal-sized segments, we have considered two split operations for the input dataset.\vspace{-0.1cm}
\begin{itemize}
    \item $\mathcal{T}_4$: Image $\mathcal{I}$ is divided into \textit{four} equal-sized segments with dimension $\floor*{\frac{m}{2}} \times \floor*{\frac{m}{2}}$.
    \item $\mathcal{T}_9$: Image $\mathcal{I}$ is divided into \textit{nine} equal-sized segments with dimension $\floor*{\frac{m}{3}} \times \floor*{\frac{m}{3}}$.\vspace{-0.1cm}
\end{itemize}

\begin{figure}[!t]
\centering
\includegraphics[width=0.5\linewidth]{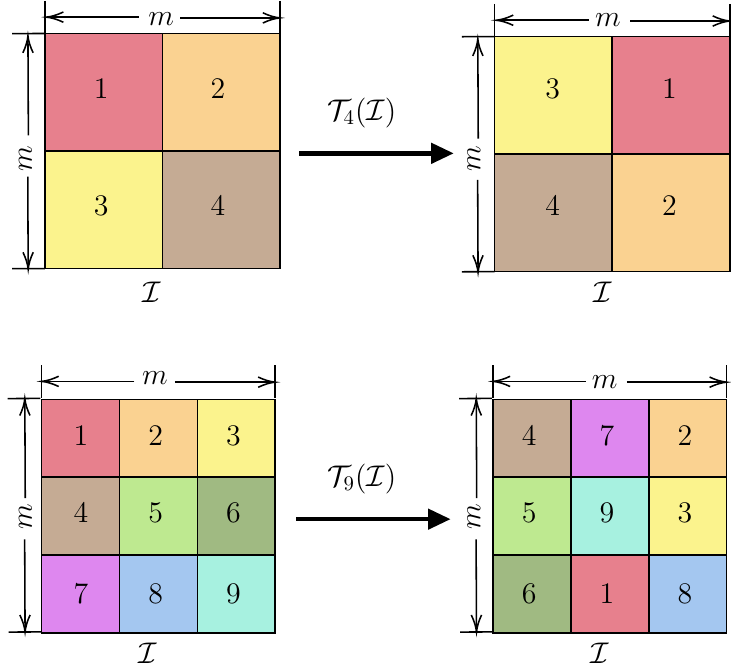}\vspace{-0.1cm}
\caption{\textit{Non-Overlapping Split-and-Shuffle Transformation}: An image $\mathcal{I}$ is divided into specific number of non-overlapping segments, and are shuffled with a randomly chosen fixed order\vspace{-0.4cm}}
\label{fig:transformation}
\end{figure}

\begin{figure}[!t]
\centering
\subfloat[]{\includegraphics[width=0.3\linewidth]{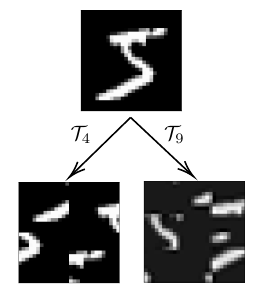}}
\subfloat[]{\includegraphics[width=0.3\linewidth]{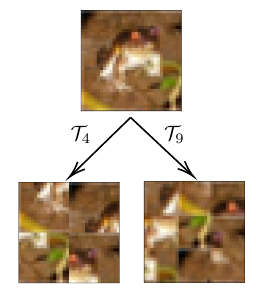}}
\subfloat[]{\includegraphics[width=0.3\linewidth]{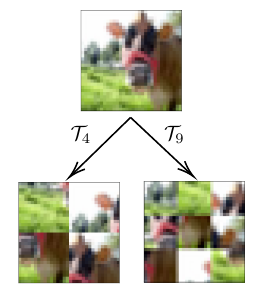}}\vspace{-0.1cm}
\caption{Transforming \textbf{(a)} MNIST, \textbf{(b)} CIFAR-10, and \textbf{(c)} CIFAR-100 dataset considering non-overlapping $\mathcal{T}_4(\cdot)$ and $\mathcal{T}_9(\cdot)$ split-and-shuffle transformation\vspace{-0.3cm}}
\label{fig:dataset_transformation}
\end{figure}

The overview of the shuffle operation for a randomly chosen fixed order is provided in Fig.~\ref{fig:transformation}. A sample resultant image from each of the MNIST, CIFAR-10 and CIFAR-100 datasets after applying \textit{non-overlapping split-and-shuffle} transformation is provided in Fig.~\ref{fig:dataset_transformation}. We want to mention that a similar transformation of $\mathcal{T}_{16}$ will further divide the inputs into more small-sized segments, which will affect the accuracy adversely. Thus such a transformation is not desired as a detector since it will result in high false positives.\vspace{-0.2cm}

\subsubsection{Overlapping Transformation}\label{sec:overlap_split_and_shuffle}
The \textit{overlapping split-and-shuffle} transformation works in the same way as the non-overlapping transformation. However, in this case, the image $\mathcal{I}$ is divided into multiple overlapping segments. Like the non-overlapping transformation, in this scenario, the segments are shuffled with a random permutation of fixed order for all images in a dataset. Like the non-overlapping transformation, we have considered two split operations for the input dataset based on the number of segments.\vspace{-0.1cm}
\begin{itemize}
    \item $\mathcal{T}_4$: Image $\mathcal{I}$ is divided into \textit{four} segments like the non-overlapping transformation. However, in this scenario, all the segments are extended along their sides to include more image features within the segments than the non-overlapping counterparts. In this paper, without loss of generality, we have considered extending the segments 50\% along the sides. The dimension of each segment, in this scenario, is $(\floor*{\frac{m}{2}} + \floor*{\frac{m}{4}}) \times (\floor*{\frac{m}{2}} + \floor*{\frac{m}{4}})$.
    \item $\mathcal{T}_9$: Image $\mathcal{I}$ is divided into \textit{nine} segments like the non-overlapping transformation. However, like the $\mathcal{T}_4$ transformation, all the segments are extended along their sides to include more image features within the segments than the non-overlapping counterparts. In this paper, we have considered extending the segments 50\% along the sides without loss of generality. The dimension of each segment, in this scenario, is not the same. Hence, before combining all the segments, each segment is reshaped into a dimension of $(\floor*{\frac{m}{3}} + \floor*{\frac{m}{6}}) \times (\floor*{\frac{m}{3}} + \floor*{\frac{m}{6}})$.\vspace{-0.1cm}
\end{itemize}

\begin{figure}[!b]
\centering
\includegraphics[width=0.75\linewidth]{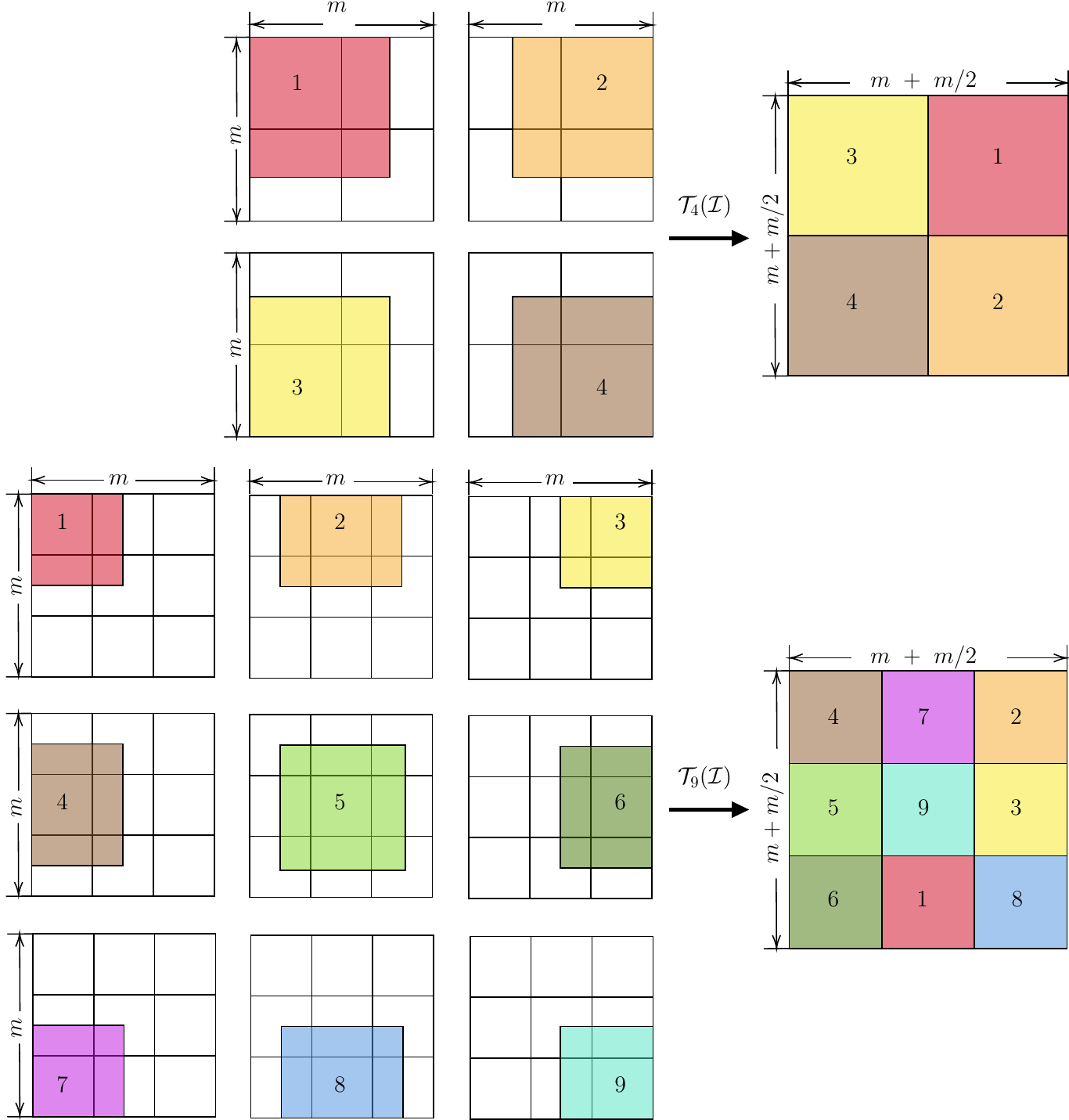}\vspace{-0.1cm}
\caption{\\textit{Overlapping Split-and-Shuffle Transformation}: An image $\mathcal{I}$ is divided into specific number of overlapping segments, and are shuffled with a randomly chosen fixed order\vspace{-0.4cm}}
\label{fig:transformation_2}
\end{figure}

\begin{figure}[!b]
\centering
\subfloat[]{\includegraphics[width=0.3\linewidth]{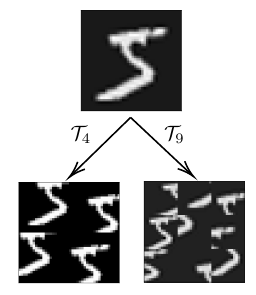}}
\subfloat[]{\includegraphics[width=0.3\linewidth]{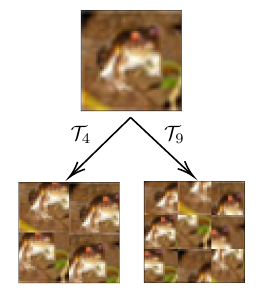}}
\subfloat[]{\includegraphics[width=0.3\linewidth]{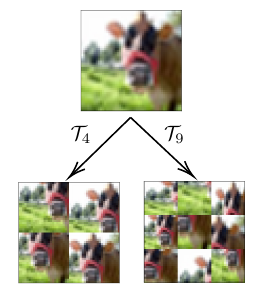}}\vspace{-0.1cm}
\caption{Transforming \textbf{(a)} MNIST, \textbf{(b)} CIFAR-10, and \textbf{(c)} CIFAR-100 dataset considering overlapping $\mathcal{T}_4(\cdot)$ and $\mathcal{T}_9(\cdot)$ split-and-shuffle transformation}
\label{fig:dataset_transformation_2}
\end{figure}

The overview of the overlapping split-and-shuffle transformation for a randomly chosen fixed order is provided in Fig.~\ref{fig:transformation_2}. A sample resultant image from each of the MNIST, CIFAR-10 and CIFAR-100 datasets after applying \textit{overlapping split-and-shuffle} transformation is provided in Fig.~\ref{fig:dataset_transformation_2}. We can observe that the images in Fig.~\ref{fig:dataset_transformation_2} have more features than the images in Fig.~\ref{fig:dataset_transformation}. The inclusion of more features aids the training process to achieve better accuracy than the non-overlapping transformations. Better accuracy in the detector models helps to deal with the false positives more efficiently.\vspace{-0.2cm}

\subsection{A Formal Approach Explaining the Detection}
In this section, we try to formally analyze why the transferability of adversarial examples generated from $\mathcal{M}_{\mathcal{U}}$ to $\mathcal{M}_{\mathcal{D}}$ is challenging because of the diversity in decision boundaries. Without loss of generality, we perform our analysis using the minimum norm attack. As discussed in Section~\ref{sec:prelim}, if an adversary employs minimum norm attack to generate adversarial example $x_{*}$ from a clean data point $x$ to move it from class $\mathcal{C}_i$ to the decision boundary between $\mathcal{C}_i$ and $\mathcal{C}_j$, we can write the following equation\vspace{-0.1cm}
\begin{equation*}
    x_{*} = x - \frac{g(x)}{\|\nabla_{x} g(x)\|_{2}}\cdot \nabla_{x} g(x)
\end{equation*}
where $g(\cdot) = g_i(\cdot) - g_j(\cdot)$ is the discriminant function for the decision boundary between class $\mathcal{C}_i$ and $\mathcal{C}_j$, $g_i(\cdot)$ and $g_j(\cdot)$ are the discriminant functions for individual classes $\mathcal{C}_i$ and $\mathcal{C}_j$ respectively.

Let us consider linear decision boundaries for our study. The discriminant functions for class $\mathcal{C}_i$ and $\mathcal{C}_j$ in model $\mathcal{M}_{\mathcal{U}}$ can be written as\vspace{-0.1cm}
\begin{equation*}
    g_i(x) = w_i^T\cdot x + b_i; \text{\hspace{0.3cm}} g_j(x) = w_j^T\cdot x + b_j
\end{equation*}
where $w_i$, $b_i$ are the parameters for the $g_i(\cdot)$ and $w_j$, $b_j$ are the parameters for the $g_j(\cdot)$. Hence, we can write\vspace{-0.1cm}
\begin{align*}
    x_{*} &= x - \frac{(w_i^T - w_j^T)\cdot x + (b_i - b_j)}{\|w_i^T - w_j^T\|_{2}}\cdot (w_i - w_j)\\
    &= x - \frac{w^T\cdot x + b}{\|w\|_{2}}\cdot w
\end{align*}
where $w = w_i - w_j$ and $b = b_i - b_j$.

The discriminant functions for class $\mathcal{C}_i$ and $\mathcal{C}_j$ in model $\mathcal{M}_{\mathcal{D}}$ can be written as\vspace{-0.1cm}
\begin{equation*}
    \overline{g}_i(\overline{x}) = \overline{w}_i^T\cdot \overline{x} + \overline{b}_i; \text{\hspace{0.3cm}} \overline{g}_j(\overline{x}) = \overline{w}_j^T\cdot \overline{x} + \overline{b}_j
\end{equation*}
where $\overline{w}_i$, $\overline{b}_i$ are the parameters for the $\overline{g}_i(\cdot)$, $\overline{w}_j$, $\overline{b}_j$ are the parameters for the $\overline{g}_j(\cdot)$, and $\overline{x} = T_{n}(x)$ is the transformed data point after the application of split-and-shuffle transformation. We aim to analyse the effect of $\overline{x}_* = T_n(x_*)$ in $\mathcal{M}_\mathcal{D}$. Now, we can write\vspace{-0.1cm}
\begin{equation*}
    \overline{x}_* = \overline{x} - \sigma \cdot T_n(w)
\end{equation*}
where $\sigma = \frac{w^T\cdot x + b}{\|w\|_{2}}$ is a scalar quantity responsible for creating the adversarial example. Next, we compute\vspace{-0.1cm}
\begin{align*}
    \overline{g}_i(\overline{x}_*) - \overline{g}_j(\overline{x}_*) &= (\overline{w}_i - \overline{w}_j)^T \cdot \overline{x}_* + (\overline{b}_i - \overline{b}_j) \\
    &= \overline{w}^T \cdot \overline{x}_* + \overline{b} \text{\hspace{0.3cm}}[\overline{w} = \overline{w}_i - \overline{w}_j; \overline{b} = \overline{b}_i - \overline{b}_j]\\
    &= \overline{w}^T \cdot (\overline{x} - \sigma \cdot T_n(w)) + \overline{b}\\
    &= \overline{w}^T\cdot\overline{x} + \overline{b} - \sigma \cdot \overline{w}^T \cdot T_n(w)
\end{align*}

If we do not apply any transformation on input data points, the models $\mathcal{M}_\mathcal{U}$ and $\mathcal{M}_\mathcal{D}$ will produce similar decision boundaries. Hence, $\overline{w} \approx w$, $\overline{b} \approx b$, and $T_n(w) = w$. Then,\vspace{-0.1cm}
\begin{equation*}
    \overline{g}_i(\overline{x}_*) - \overline{g}_j(\overline{x}_*) \approx w^T \cdot x + b - \frac{w^T\cdot x + b}{\|w\|_{2}} \cdot \|w\|_2 \approx 0
\end{equation*}

Thus, $x_*$ will also lie on the decision boundary between class $\mathcal{C}_i$ and $\mathcal{C}_j$ of model $\mathcal{M}_\mathcal{D}$ explaining the transferability of adversarial examples between models having similar decision boundaries. More dissimilarity in decision boundaries between the unprotected and detector model will ensure a more diverse impact of adversarial examples (generated from the unprotected model) on the detector model. For non-linear decision boundaries, we can similarly explain by considering $\sigma$ as $\frac{g(x)}{\|\nabla_{x} g(x)\|_{2}}$, $w$ as $\nabla_{x} g(x)$, and $\overline{w}$ as $\nabla_{x} \overline{g}(\overline{x})$, where $g(\cdot)$ and $\overline{g}(\cdot)$ are the discriminant functions between class $\mathcal{C}_i$ and $\mathcal{C}_j$ in $\mathcal{M}_\mathcal{U}$ and $\mathcal{M}_\mathcal{D}$ respectively.\vspace{-0.2cm}

\section{Building Detector Model $\mathcal{M}_{\mathcal{D}}$ with Diverse Feature Selection}\label{sec:orthogonal-feature}

In this section, we discuss another method to detect adversarial attacks by prioritizing relatively less important components in training data. The main objective of this method is to train another neural network model with a diverse feature set than the original model. First, we explain the notion of selecting diverse features, followed by a detailed discussion on the methodology.\vspace{-0.3cm}

\begin{figure}[!b]
\centering
\floatbox[{\capbeside\thisfloatsetup{capbesideposition={right},capbesidewidth=3.6cm}}]{figure}[\FBwidth]
{\caption{Importance of diversity between the unprotected and the detector model to detect adversarial examples. The left-side of classifiers $\mathcal{X}$ is class $\mathcal{C}_1$ and right-side is class $\mathcal{C}_2$. The same decision for classifier $\mathcal{Y}$}
\label{fig:classifier-motivation}}
{\includegraphics[width=\linewidth]{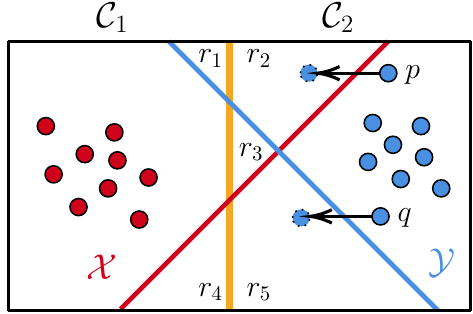}}\vspace{-0.2cm}
\end{figure}

\subsection{Notion behind the Construction}
The primary objective of this approach is to establish diversity with respect to decision boundaries between the original model and the detector model. In order to illustrate the importance of model diversity, let us consider Fig.~\ref{fig:classifier-motivation} showing a two-class classification by two classifiers. The golden line denotes the golden boundary between the classes, the red line denotes the boundary made by classifier $\mathcal{X}$, and the blue line represents the boundary made by classifier $\mathcal{Y}$. The region to the left-side of the golden boundary belongs to class $\mathcal{C}_1$, and the region to the right-side of the golden boundary belongs to class $\mathcal{C}_2$. In the figure, red and blue samples are well classified by the classifiers $\mathcal{X}$ and $\mathcal{Y}$. Let an adversary perturb a clean sample $p$ and moves it to region $r_2$ in the input space. The sample $p$ still belongs to class $\mathcal{C}_2$ according to the golden decision boundary, but $\mathcal{X}$ classifies it as belonging to class $\mathcal{C}_1$. Similarly, let the adversary perturb a clean sample $q$ and moves it to region $r_5$ in the input space. The region $r_5$ belongs to class $\mathcal{C}_2$ according to the golden decision boundary, but classifier $\mathcal{Y}$ will classify perturbed $q$ as belonging to class $\mathcal{C}_1$. Let us consider an ensemble of both the classifiers $\mathcal{X}$ and $\mathcal{Y}$. In this scenario, region $r_2$ will be classified as class $\mathcal{C}_1$ by classifier $\mathcal{X}$ but as class $\mathcal{C}_2$ by classifier $\mathcal{Y}$. Similarly, region $r_5$ will be classified as class $\mathcal{C}_1$ by classifier $\mathcal{Y}$, but as class $\mathcal{C}_2$ by classifier $\mathcal{X}$. So, as the outputs of the classifiers differ for both scenarios, we can denote the perturbed samples as adversarial examples, which promotes the use of classifiers with good accuracy but diverse decision boundaries to detect adversarial attacks. Hence, to detect adversarial examples, the primary objective should be -- the significant features regarding one model for moving clean samples in the feature space to generate adversarial examples should not be significant features in another model. 

In order to achieve the objective, we propose a diverse feature selection approach, namely \emph{contrast-significant-features}. The diversity between the original model $\mathcal{M}_\mathcal{U}$ and the detector model $\mathcal{M}_\mathcal{D}$ is achieved by assigning importance to different features for different models. We used the gradient-based visualization method~\cite{aggarwal2018neural} for computing the most significant parameters of model $\mathcal{M}_\mathcal{U}$. The overview of the contrast-significant-features method is as follows\vspace{-0.1cm}

\begin{itemize}
    \item \textit{Select significant parameters:} We identify the pixels of input images, which mostly affect the output of $\mathcal{M}_\mathcal{U}$, and then we identify the parameters of hidden layers that describe those pixels.
    \item \textit{Contrast the parameters while training:} We train the detector model $\mathcal{M}_\mathcal{D}$ with the same architecture as $\mathcal{M}_\mathcal{U}$ but restraining the significant parameters.\vspace{-0.1cm}
\end{itemize}

The method produces a detector model $\mathcal{M}_\mathcal{D}$ with the same structure but with different significant parameters, thereby incorporating diversity between decision boundaries. We examine the diversity by measuring the CKA similarity between the models and discuss the results later in Section~\ref{sec:results}. Like the split-and-shuffle transformation mentioned previously, the contrast-significant-features method also generates diverse decision boundaries without adversely affecting the accuracy, which is useful for reducing both false negatives and false positives from the detection. Next, we discuss the working of \textit{contrast-feature-selection} methodology in detail.\vspace{-0.4cm} 


\subsection{Contrast-Significant-Features Method}\label{sec:csf}
We first train an unprotected model $\mathcal{M}_\mathcal{U}$ and compute the gradient of $\mathcal{M}_\mathcal{U}$'s output with respect to each pixel of an input image. We select a set of pixels $\mathcal{P}$, having gradients with higher magnitude values. The pixels in $\mathcal{P}$ mostly dominate the output of $\mathcal{M}_\mathcal{U}$. Next, we compute the gradient of different neuron's output in the intermediate layers with respect to the pixels in $\mathcal{P}$ and select a set of neurons $\mathcal{N}$ having gradients with higher magnitude values. The pixels in $\mathcal{P}$ and parameters $\mathcal{W}$ associated to the neurons in $\mathcal{N}$ are the significant pixels and parameters for the model $\mathcal{M}_\mathcal{U}$. The mathematical details of this process are given below:

Let $N$ be the number of input samples, $O$ be the output of $\mathcal{M}_\mathcal{U}$ and $u(\cdot)$ be the unit step function. Let, $g_{i}=|\frac{\partial O}{\partial x_{i}}|$ and $g_{i}^{jk}=|\frac{\partial h^{jk}}{\partial x_{i}}|$ where $x_{i}$ is the $i^{th}$ pixel of the input image, $g_{i}$ is the magnitude of gradient of $O$ with respect to $x_{i}$, $h^{jk}$ is the output of the $j$th neuron of the $k^{th}$ layer, and $g_{i}^{jk}$ is the magnitude of gradient of $h^{jk}$ neuron with respect to $x_{i}$. Let us define the following thresholds, $\theta_{1}$: gradient threshold for output, $\theta_{2}$: gradient threshold for neurons, $\theta_{3}$: importance threshold for neuron.
$\theta_{4}$: significant threshold for pixels. It defines the number of samples for which the pixels are significant. {$\theta_{5}$: significant threshold for neurons. It defines the number of samples for which the neurons are significant. The value of $\theta_{4}$ and $\theta_{5}$ lie between $0$ to $N$}.
These thresholds act as hyperparameters and are selected empirically based on the observation obtained from $\mathcal{M}_\mathcal{U}$.

Now for a particular input sample $s$, let $I_{s,i} = u(g_{s,i}-\theta_{1})$ denote significance of the output for sample  with respect to $x_{i}$ pixel. Here, $g_{s,i}$ is the value of $g_{i}$ for sample $s$. We say a pixel $x_{i}$ is important if the following condition holds\vspace{-0.1cm}
\begin{equation}
\left(\sum_{s=1}^{N}I_{s,i}\right)>\theta_{4}\label{equ:imp-pixel}
\end{equation}

\vspace{-0.2cm}
Let $\mathcal{P}$ be the set of all important pixels that dictates the decision of $\mathcal{M}_\mathcal{U}$ mostly and is obtained based on the Eq.~(\ref{equ:imp-pixel}). Next, we identify the important neurons $\mathcal{N}$ in a given layer. Let us consider the $j^{th}$ neuron of the $k^{th}$ layer i.e., $h^{jk}$. Let us define $F_{s,i}^{j,k} = u(g_{s,i}^{jk}-\theta_{2})$ as the importance of the $h^{jk}$ neuron with respect to $x_{i}$ pixel for sample $s$. Here, $g_{s,i}^{jk}$ is the value of $g_{i}^{jk}$ for sample s. We compute $F_{s,i}^{jk}$ with respect to all the pixels $x_{i}\in \mathcal{P}$. Let $M_{s}^{jk} = u(\sum_{x_{i}\in \mathcal{P}}F_{s,i}^{jk}-\theta_{3})$ denote the importance of a neuron with respect to all important pixels. A neuron is considered as significant if the following condition holds\vspace{-0.1cm}
\begin{equation}
\left(\sum_{s=1}^{N}M_{s}^{jk}\right)>\theta_{5}\label{equ:imp-neuron}
\end{equation}

\vspace{-0.2cm}
The operation can be performed for all the neurons for any number of layers. In our case we perform it for the last two layers. All the parameters associated with the neurons satisfying Eq.~(\ref{equ:imp-neuron}) are considered as significant parameters of $\mathcal{M}_\mathcal{U}$ and are represented by $\mathcal{W}$.

%
%

Next, we train the detector model $\mathcal{M}_\mathcal{D}$ with the same architecture as $\mathcal{M}_\mathcal{U}$ but restricting the significant parameters $\mathcal{W}$ during training. The restriction is imposed by forcing the parameters in $\mathcal{W}$ to zero at each training epoch of model $\mathcal{M}_\mathcal{D}$. Hence, the significant parameters generated after training of $\mathcal{M}_\mathcal{D}$ will be entirely different from the parameters in $\mathcal{W}$.\vspace{-0.3cm}

\section{Experimental Evaluation}\label{sec:results}
In this section, we evaluate and validate the proposed methodology first in the presence of a \textit{zero knowledge adversary} $(\mathcal{A}_\mathcal{Z})$, where the adversary crafts adversarial examples with the original images on the unprotected model and then in the presence of a \textit{perfect knowledge adversary} $(\mathcal{A}_\mathcal{P})$, where the adversary crafts adversarial examples optimizing for both the detector and unprotected model. The threat models for both these adversaries are previously discussed in Section~\ref{sec:threat_model}. We considered Convolutional Neural Network (CNN) architectures for all our experiments in this paper. The details of the CNN architectures and hyper-parameters used to train three standard image classification datasets, namely MNIST, CIFAR-10, and CIFAR-100 are mentioned in Table~\ref{table:cnn_arch}. More specifically, we used VGG-16 architecture to train the CIFAR-10 dataset and VGG-19 architecture to train the CIFAR-100 dataset. We used `adam' optimization and `categorical crossentropy' loss for training all the models. We also used batch normalization at each layer and dropout to prevent the models from overfitting and stabilize the learning process.\vspace{-0.3cm}

\begin{table}[!t]
\centering
\caption{Architectures of different CNN models used to train MNIST, CIFAR-10 and CIFAR-100. conv$i$-$j$ signifies $j$ convolution filters of size $i$ each. Both convolution and maxpool filters use stride size 2 and zero padding. fc-$k$ signifies fully connected layer with $k$ neurons.}\label{table:cnn_arch}
\resizebox{\linewidth}{!}{
\begin{tabular}{c|l|}
\cline{2-2}
 & \multicolumn{1}{c|}{\textbf{CNN Architecture}} \\ \hline
\multicolumn{1}{|c|}{\textbf{MNIST}} & \begin{tabular}[c]{@{}l@{}}input, conv5-8, maxpool, conv3-16, maxpool,\\ fc-10, fc-10, softmax\end{tabular} \\ \hline
\multicolumn{1}{|c|}{\textbf{CIFAR-10}} & \begin{tabular}[c]{@{}l@{}}input, conv3-64, conv3-64, maxpool,\\ conv3-128, conv3-128, maxpool,\\ conv3-256, conv3-256, conv3-256, maxpool,\\ conv3-512, conv3-512, conv3-512, maxpool,\\ conv3-512, conv3-512, conv3-512, maxpool,\\ fc-4096, fc-4096, fc-10, softmax\end{tabular} \\ \hline
\multicolumn{1}{|c|}{\textbf{CIFAR-100}} & \begin{tabular}[c]{@{}l@{}}input, conv3-64, conv3-64, maxpool,\\ conv3-128, conv3-128, maxpool,\\ conv3-256, conv3-256, conv3-256, conv3-256, maxpool,\\ conv3-512, conv3-512, conv3-512, conv3-512, maxpool,\\ conv3-512, conv3-512, conv3-512, conv3-512, maxpool,\\ fc-4096, fc-4096, fc-100, softmax\end{tabular} \\ \hline
\end{tabular}}\vspace{-0.2cm}
\end{table}

\subsection{Evaluation in the presence of zero knowledge adversary $\mathcal{A}_\mathcal{Z}$}

\subsubsection{Evaluation for split-and-shuffle input transformation}\label{sec:results_input}
We consider both non-overlapping and overlapping split-and-shuffle transformation, as discussed in Section~\ref{sec:split_and_transform}, for our initial observation. The $\mathcal{T}_4(\cdot)$ and $\mathcal{T}_9(\cdot)$ transformations produce separate training datasets using MNIST, CIFAR-10, and CIFAR-100 for each non-overlapping and overlapping transformations. We train a neural network model $\mathcal{M}_{\mathcal{U}}$ with the original dataset, a neural network model $\mathcal{M}_{\mathcal{D}}^{\mathcal{T}_4}$ with the dataset produced from $\mathcal{T}_4(\cdot)$ transformation, and a neural network model $\mathcal{M}_{\mathcal{D}}^{\mathcal{T}_9}$ with the dataset produced from $\mathcal{T}_9(\cdot)$ transformation. The individual classification accuracies for each of the models and for each transformations are mentioned in Table~\ref{table:accuracy}. We can observe from the table that the removal of spatial correlation from training images does not adversely affect the overall accuracy of a neural network model. Also, the overlapping split-and-shuffle transformations produce detector models with better accuracies than the non-overlapping transformations, as already speculated in Section~\ref{sec:overlap_split_and_shuffle}. \textit{Hence, in all our future experiments, we consider models trained with overlapping split-and-shuffle transformations to reduce the number of false positives.} Moreover, as discussed previously, the detector model $\mathcal{M}_{\mathcal{D}}$ is used only for detecting adversarial examples. The overall accuracy of the ensemble is reported from the model $\mathcal{M}_{\mathcal{U}}$, which is trained with original training images. Thus the proposed ensemble-based approach does not compromise the overall accuracy of a dataset. Moreover, as discussed previously, the detector model $\mathcal{M}_{\mathcal{D}}$ is used only for detecting adversarial examples. The overall accuracy of the ensemble is reported from the model $\mathcal{M}_{\mathcal{U}}$, which is trained with original training images. Thus the proposed ensemble-based approach does not compromise the overall accuracy for a dataset.


\begin{table}[!t]
\caption{Accuracy of different neural network models trained with original datasets and different split-and shuffle input transformations for MNIST, CIFAR-10 and CIFAR-100\label{table:accuracy} }
\resizebox{\linewidth}{!}{
\begin{tabular}{cc|c|c|c|c|}
\cline{3-6}
 &  & \multicolumn{2}{c|}{\textbf{\begin{tabular}[c]{@{}c@{}}Non-Overlapping\\ Transformation\end{tabular}}} & \multicolumn{2}{c|}{\textbf{\begin{tabular}[c]{@{}c@{}}Overlapping\\ Transformation\end{tabular}}} \\ \cline{2-6} 
\multicolumn{1}{c|}{} & $\mathcal{M}_{\mathcal{U}}$ & $\mathcal{M}_{\mathcal{D}}^{\mathcal{T}_4}$ & $\mathcal{M}_{\mathcal{D}}^{\mathcal{T}_9}$ & $\mathcal{M}_{\mathcal{D}}^{\mathcal{T}_4}$ & $\mathcal{M}_{\mathcal{D}}^{\mathcal{T}_9}$ \\ \hline
\multicolumn{1}{|c|}{\textbf{MNIST}} & 98.84 & 98.74 & 98.65 & 98.79 & 98.68 \\ \hline
\multicolumn{1}{|c|}{\textbf{CIFAR-10}} & 91.98 & 84.89 & 77.68 & 89.48 & 86.64 \\ \hline
\multicolumn{1}{|c|}{\textbf{CIFAR-100}} & 86.90 & 68.73 & 67.12 & 83.30 & 79.21 \\ \hline
\end{tabular}}\vspace{-0.2cm}
\end{table}

\begin{figure}[!t]
\centering
\includegraphics[width=\linewidth]{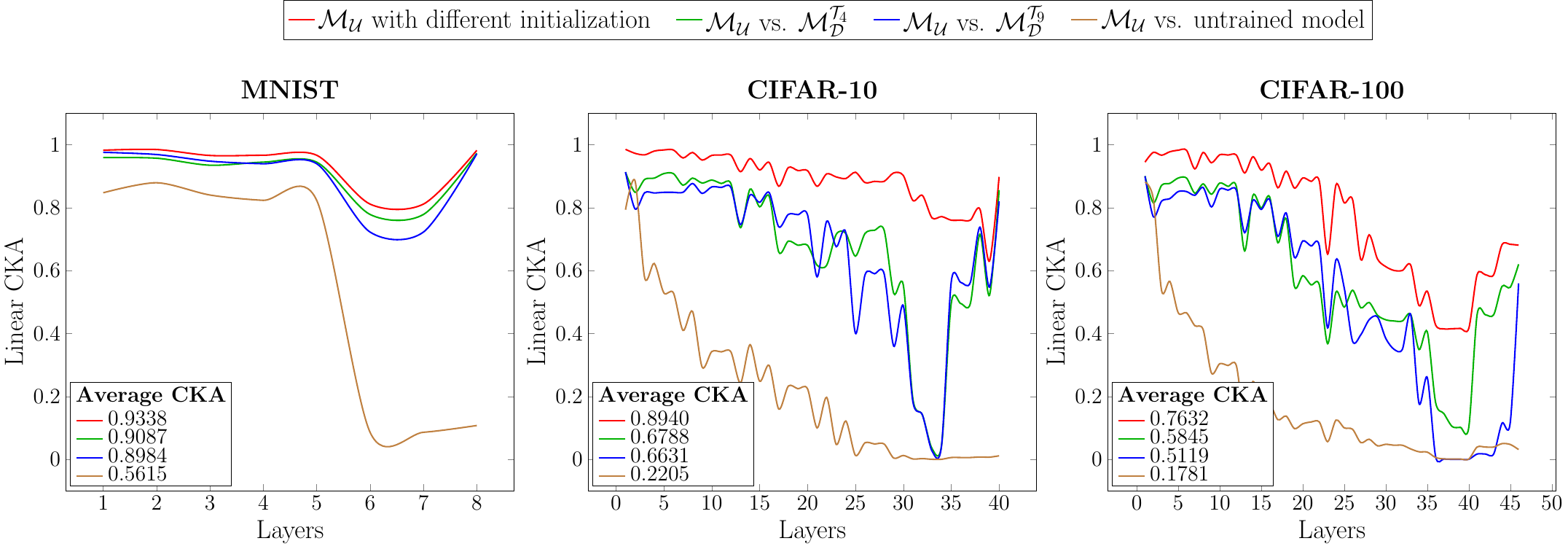}
\caption{Linear CKA plots showing the similarities of different neural network models trained with different split-and-shuffle input transformations, different random initialization, and untrained model with respect to the neural network model trained with the original inputs for MNIST, CIFAR-10 and CIFAR-100. The Linear CKA values in the last layer are high for both $\mathcal{M}_{\mathcal{D}}^{\mathcal{T}_4}$ and $\mathcal{M}_{\mathcal{D}}^{\mathcal{T}_9}$ as the classification problem is the same and the last layer outputs for appropriately trained models are equivalent\vspace{-0.3cm}}
\label{fig:cka_plots}
\end{figure}

We use Linear Central Kernel Alignment (CKA) analysis proposed by Kornblith \textit{et al.}~\cite{DBLP:conf/icml/Kornblith0LH19} to measure the similarity between decision boundaries learned by the models $\mathcal{M}_{\mathcal{U}}$, $\mathcal{M}_{\mathcal{D}}^{\mathcal{T}_4}$, and $\mathcal{M}_{\mathcal{D}}^{\mathcal{T}_9}$. We compute the layer-wise linear CKA values of four different neural network models with respect to $\mathcal{M}_{\mathcal{U}}$ -- \textbf{(i)} $\mathcal{M}_{\mathcal{U}}$ but trained with a different random initialization, \textbf{(ii)} $\mathcal{M}_{\mathcal{D}}^{\mathcal{T}_4}$, \textbf{(iii)} $\mathcal{M}_{\mathcal{D}}^{\mathcal{T}_9}$, and \textbf{(iv)} an untrained neural network model\footnote{We considered the neural network architectures used in Table~\ref{table:cnn_arch} with all the trainable parameters initialized to random values as the untrained neural network models, i.e., the models are not trained using any error feedback.} for MNIST, CIFAR-10 and CIFAR-100. Fig.~\ref{fig:cka_plots} shows the layer-wise Linear CKA plots along with average CKA values over all the layers. We can observe that the neural network models trained with different random initialization produce similar decision boundaries than the rest of the scenario, as the CKA values are significantly higher in this case. Hence a higher CKA value between two neural network models implies a good transferability of adversarial examples. The most dissimilar decision boundary occurs when $\mathcal{M}_{\mathcal{U}}$ is compared with an untrained neural network model. We can observe that $\mathcal{M}_{\mathcal{D}}^{\mathcal{T}_9}$ produces more dissimilar decision boundaries than $\mathcal{M}_{\mathcal{D}}^{\mathcal{T}_4}$ with respect to $\mathcal{M}_{\mathcal{U}}$.

\begin{figure}[!t]
\centering
\includegraphics[width=\linewidth]{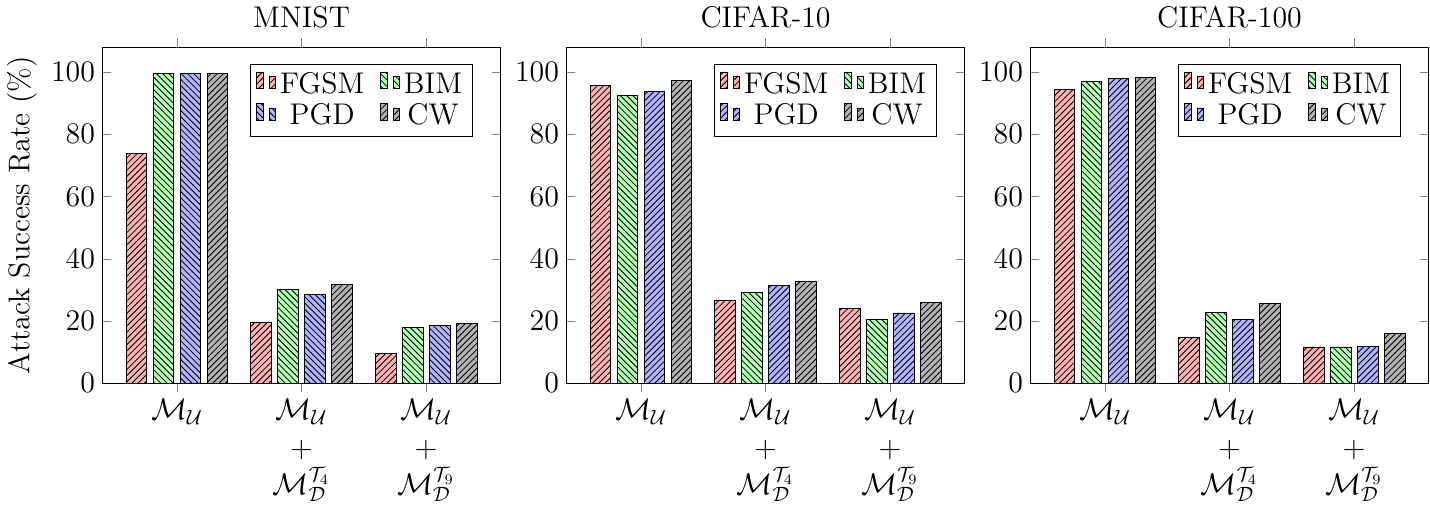}
\caption{The success rate of different adversarial attacks on unprotected neural network model, and for ensembles trained with different split-and-shuffle input transformations for MNIST, CIFAR-10 and CIFAR-100. The attack parameters are -- FGSM ($\eta = 0.3$), BIM and PGD ($\eta = 0.4, \alpha = 1$, iteration=50), CW ($c = 0.1$, $\kappa = 0.2$, iteration=1000) for MNIST; FGSM ($\eta = 0.02$), BIM and PGD ($\eta = 0.03, \alpha = 1$, iteration=50), CW ($c = 0.1$, $\kappa = 0.2$, iteration=1000) for both CIFAR-10 and CIFAR-100\vspace{-0.2cm}}
\label{fig:attack_accuracy_2}
\end{figure}

We consider four adversarial attacks, namely FGSM, BIM, PGD and CW, to evaluate the proposed ensemble-based defense strategy. The equations for generating adversarial examples using these attack strategies are previously discussed in Section~\ref{sec:prelim}. We consider three scenarios to evaluate the attack success rate (discussed previously in Section~\ref{sec:overview}) -- \textbf{(i)} attacking only $\mathcal{M}_{\mathcal{U}}$, \textbf{(ii)} attacking the ensemble of $\mathcal{M}_{\mathcal{U}}$ and $\mathcal{M}_{\mathcal{D}}^{\mathcal{T}_4}$, and \textbf{(iii)} attacking the ensemble of $\mathcal{M}_{\mathcal{U}}$ and $\mathcal{M}_{\mathcal{D}}^{\mathcal{T}_9}$. In all three cases, the adversarial examples are generated from $\mathcal{M}_{\mathcal{U}}$. Fig.~\ref{fig:attack_accuracy_2} show the attack success rate for all three attacks considering MNIST, CIFAR-10 and CIFAR-100 with different attack parameters. We can observe that the attack success rate is maximum when considering only $\mathcal{M}_{\mathcal{U}}$. However, the success rate is lower in the presence of the detector $\mathcal{M}_{\mathcal{D}}^{\mathcal{T}_4}$ and even lower in the presence of the detector $\mathcal{M}_{\mathcal{D}}^{\mathcal{T}_9}$. This is due to the fact that the decision boundary of $\mathcal{M}_{\mathcal{U}}$ is more dissimilar to $\mathcal{M}_{\mathcal{D}}^{\mathcal{T}_9}$ than $\mathcal{M}_{\mathcal{D}}^{\mathcal{T}_4}$, as shown in Fig.~\ref{fig:cka_plots}. The observation validates our argument that the dissimilar decision boundaries produced during the training by prioritizing lower-level features make it challenging to transfer adversarial examples between models.


In addition to measuring the success rate of adversarial attacks, it is also essential to evaluate the true positive rate ($TPR$) and false positive rate ($FPR$) of the proposed detection-based defense methodology. The $TPR$ and $FPR$ are useful metrics that estimate how many adversarial examples are detected by the proposed method and how many natural images are considered as adversarial examples, respectively. The $TPR$ and $FPR$ are computed as

\begin{equation*}
    TPR = \frac{TP}{TP+FN}; \text{\hspace{0.3cm}} FPR = \frac{FP}{FP+TN}
\end{equation*}
where $TP$: number of adversarial examples detected as adversarial examples, $FN$: number of adversarial examples detected as clean examples, $FP$: number of clean examples detected as adversarial examples, and $TN$: number of clean examples detected as clean examples. We compute $TPR$ and $FPR$ for the ensembles $\mathcal{M}_{\mathcal{U}} + \mathcal{M}_{\mathcal{D}}^{\mathcal{T}_4}$ and $\mathcal{M}_{\mathcal{U}} + \mathcal{M}_{\mathcal{D}}^{\mathcal{T}_9}$ considering FGSM, BIM, PGD and CW attacks for MNIST, CIFAR-10 and CIFAR-100. The results are provided in Table~\ref{table:tpr_fpr}. The parameters of each attack is same as mentioned in the caption of Fig.~\ref{fig:attack_accuracy_2}. We can observe that the drop in accuracy for $\mathcal{M}_{\mathcal{D}}^{\mathcal{T}_9}$, as shown in Table~\ref{table:accuracy}, increases the $FPR$ of the proposed detection. However, the ensemble $\mathcal{M}_{\mathcal{U}} + \mathcal{M}_{\mathcal{D}}^{\mathcal{T}_9}$ is more robust than the ensemble $\mathcal{M}_{\mathcal{U}} + \mathcal{M}_{\mathcal{D}}^{\mathcal{T}_4}$ against adversarial examples. One can consider this as a performance-robustness trade-off for the proposed ensemble-based detection methodology.\vspace{-0.2cm}

\begin{table}[!t]
\centering
\caption{True Positive Rate and False Positive Rate of the proposed methodology using ensembles trained with different split-and-shuffle input transformation for FGSM, BIM, PGD, and CW attacks considering MNIST, CIFAR-10 and CIFAR-100\label{table:tpr_fpr}}
\resizebox{\linewidth}{!}{
\begin{tabular}{cc|c|c|c|c|c|c|c|c|}
\cline{3-10}
 &  & \multicolumn{2}{c|}{\textbf{FGSM}} & \multicolumn{2}{c|}{\textbf{BIM}} & \multicolumn{2}{c|}{\textbf{PGD}} & \multicolumn{2}{c|}{\textbf{CW}} \\ \cline{3-10} 
 &  & \textbf{TPR}     & \textbf{FPR}    & \textbf{TPR}    & \textbf{FPR}    & \textbf{TPR}    & \textbf{FPR}    & \textbf{TPR}    & \textbf{FPR}   \\ \hline
\multicolumn{1}{|c|}{\multirow{2}{*}{\textbf{MNIST}}}     & $\mathcal{M}_{\mathcal{U}} + \mathcal{M}_{\mathcal{D}}^{\mathcal{T}_4}$ & 0.74 & 0.08 & 0.69 & 0.06 & 0.71 & 0.06 & 0.72 & 0.07 \\ \cline{2-10} 
\multicolumn{1}{|c|}{}                                    & $\mathcal{M}_{\mathcal{U}} + \mathcal{M}_{\mathcal{D}}^{\mathcal{T}_9}$ & 0.87 & 0.19 & 0.82 & 0.11 & 0.84 & 0.12 & 0.86 & 0.11 \\ \hline
\multicolumn{1}{|c|}{\multirow{2}{*}{\textbf{CIFAR-10}}}  & $\mathcal{M}_{\mathcal{U}} + \mathcal{M}_{\mathcal{D}}^{\mathcal{T}_4}$ & 0.72 & 0.24 & 0.69 & 0.22 & 0.63 & 0.17 & 0.71 & 0.19 \\ \cline{2-10} 
\multicolumn{1}{|c|}{}                                    & $\mathcal{M}_{\mathcal{U}} + \mathcal{M}_{\mathcal{D}}^{\mathcal{T}_9}$ & 0.75 & 0.26 & 0.78 & 0.31 & 0.77 & 0.28 & 0.79 & 0.26 \\ \hline
\multicolumn{1}{|c|}{\multirow{2}{*}{\textbf{CIFAR-100}}} & $\mathcal{M}_{\mathcal{U}} + \mathcal{M}_{\mathcal{D}}^{\mathcal{T}_4}$ & 0.84 & 0.34 & 0.76 & 0.26 & 0.79 & 0.29 & 0.82 & 0.28 \\ \cline{2-10} 
\multicolumn{1}{|c|}{}                                    & $\mathcal{M}_{\mathcal{U}} + \mathcal{M}_{\mathcal{D}}^{\mathcal{T}_9}$ & 0.88 & 0.37 & 0.87 & 0.37 & 0.87 & 0.36 & 0.88 & 0.37 \\ \hline
\end{tabular}}\vspace{-0.2cm}
\end{table}

\subsubsection{Evaluation for contrast-significant-features method}
\begin{figure}[!t]
\centering
\includegraphics[width=\linewidth]{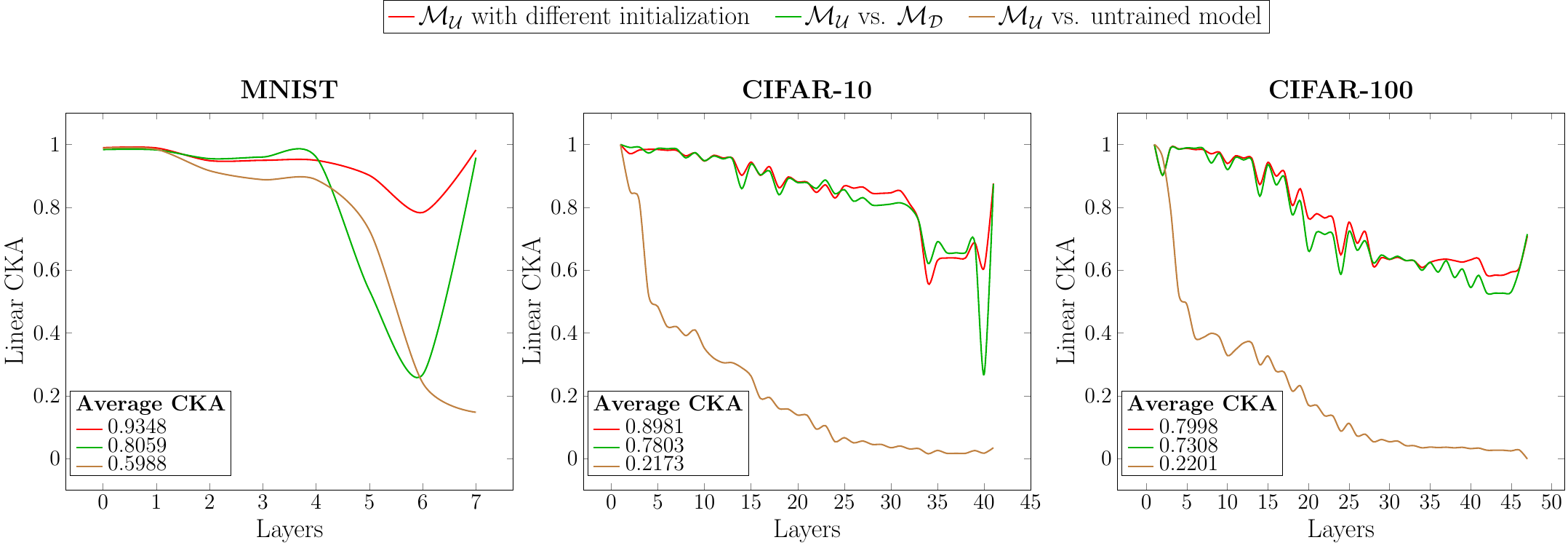}
\caption{Linear CKA plots showing the similarities of different neural network models trained with contrast-significant-features method, different random initialization, and untrained model with respect to the neural network model trained with the original inputs for MNIST, CIFAR-10 and CIFAR-100. The Linear CKA values in the last layer are high for both different initialization and $\mathcal{M}_{\mathcal{D}}$ as the classification problem is the same, and the last layer outputs for appropriately trained models are equivalent\vspace{-0.4cm}}
\label{fig:cka_plots_2}
\end{figure}

In order to evaluate the contrast-significant-features method, we follow the same experiments as performed in Section~\ref{sec:results_input}. In this scenario, we train a neural network model $\mathcal{M}_{\mathcal{U}}$ with the original dataset and a neural network model $\mathcal{M}_{\mathcal{D}}$ with the contrast-significant-features method. The individual classification accuracies for each of the models are mentioned in Table~\ref{table:accuracy_second}. We use Linear CKA analysis to measure the similarity between decision boundaries learned by the models $\mathcal{M}_{\mathcal{U}}$ and $\mathcal{M}_{\mathcal{D}}$. The layer-wise linear CKA values of three different neural network models with respect to $\mathcal{M}_{\mathcal{U}}$ -- \textbf{(i)} $\mathcal{M}_{\mathcal{U}}$ but trained with a different random initialization, \textbf{(ii)} $\mathcal{M}_{\mathcal{D}}$, and \textbf{(iii)} an untrained neural network model for MNIST, CIFAR-10 and CIFAR-100 is shown in Fig.~\ref{fig:cka_plots_2}, along with average CKA values over all the layers. The figure shows the diversity achieved by the contrast-significant-features methodology.

\begin{table}[!t]
\centering
\caption{Accuracy of the original model and the model trained with contrast-significant-features method for MNIST, CIFAR-10 and CIFAR-100}\label{table:accuracy_second}
\resizebox{0.45\linewidth}{!}{
\begin{tabular}{c|c|c|}
\cline{2-3}
 & $\mathcal{M}_{\mathcal{U}}$ & $\mathcal{M}_{\mathcal{D}}$ \\ \hline
\multicolumn{1}{|c|}{\textbf{MNIST}} & 98.19 & 96.69 \\ \hline
\multicolumn{1}{|c|}{\textbf{CIFAR-10}} & 89.50 & 88.17 \\ \hline
\multicolumn{1}{|c|}{\textbf{CIFAR-100}} & 83.66 & 81.27 \\ \hline
\end{tabular}}
\end{table}

\begin{figure}[!t]
    \centering
    \subfloat[\label{fig:varrying hyperparameter_a}]{\includegraphics[width=0.66\linewidth]{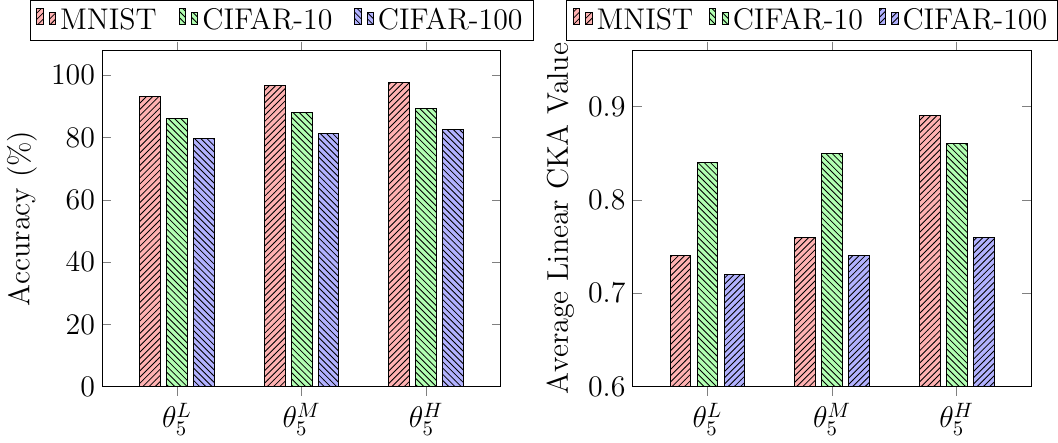}}
    \subfloat[\label{fig:varrying hyperparameter_b}]{\includegraphics[width=0.33\linewidth]{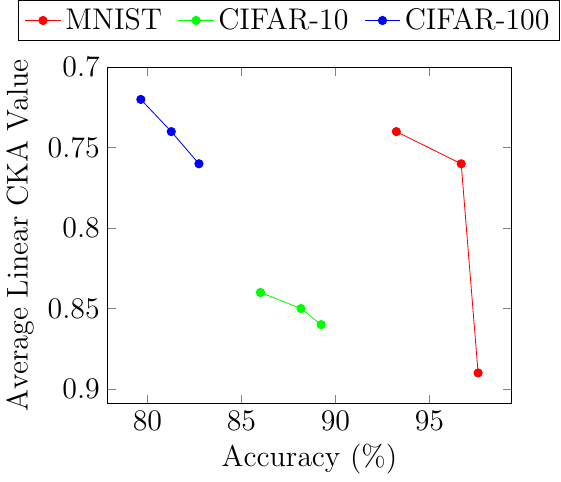}}\vspace{-0.1cm}
    \caption{(a) Accuracy and Average Linear CKA Values for different values of $\theta_{5}$ considering MNIST, CIFAR-10 and CIFAR-100, (b) Relationship between accuracy and average linear CKA values with respect to $\theta_{5}$, where the $y$-axis is drawn in reverse order\vspace{-0.3cm}}
    \label{fig:varrying hyperparameter}
\end{figure}

We provide the effect on model accuracy and the diversity of the detector model for different values of hyper-parameter $\theta_{5}$ (as discussed in Section~\ref{sec:csf}) in Fig.~\ref{fig:varrying hyperparameter}. Fig.~\ref{fig:varrying hyperparameter_a} shows the accuracy of the detector model and its average linear CKA value with respect to the unprotected model for different values of $\theta_{5}$. Equation~\ref{equ:imp-neuron} indicates that if any neuron is significant for at least $\theta_{5}$ number of samples, only that neuron will be called a significant neuron. We assume that we have a total of $N$ input samples. Now, let $\theta_{5} = k \times N$, where $k$ varies from 0 to 1. So, the value of $\theta_{5}$ will lie between 0 and $N$. If we select a $k$ closer to 1, then very few neurons will become significant. Alternatively, if we chose a $k$ closer to 0, a large number of neurons will become significant. The training of the detector model will restrict parameters associated with these significant neurons. So, it is obvious that if we restrict more parameters, we will get a diverse model, but with a compromisation in the accuracy. However, if we restrict a smaller number of parameters, accuracy will not be affected, but the detector model may not achieve the desired diversity. $\theta_{5}^{M}$ indicates the value of $\theta_{5}$, which we choose during the training of our detector models. In this case, we choose the values of $k$ as 0.5 for MNIST, 0.55 for CIFAR-10, and 0.25 for CIFAR-100 datasets to find the significant parameters. To illustrate the effect of different values of $\theta_{5}$, we choose two different values of $\theta_{5}$ for each dataset which is denoted by $\theta_{5}^{L}$ and $\theta_{5}^{H}$. For MNIST, $\theta_{5}^{L} = 0.45 \times N$, $\theta_{5}^{H} = 0.55 \times N$, for CIFAR-10, $\theta_{5}^{L} = 0.4 \times N$, $\theta_{5}^{H} = 0.75 \times N$, and for CIFAR-100, $\theta_{5}^{L} = 0.15 \times N$, $\theta_{5}^{H} = 0.35 \times N$. For each value of $\theta_{5}$, the accuracy of the model and the average linear CKA value of the detector model with respect to the unprotected model is shown in Fig.~\ref{fig:varrying hyperparameter_a}. Fig.~\ref{fig:varrying hyperparameter_b} shows the effect of $\theta_{5}$ on the accuracy vs. robustness trade-off, i.e., with an increase in accuracy, the average linear CKA value increases, and as a result, the robustness of the model against adversarial examples will decrease. Also, with a decrease in accuracy, the average linear CKA value decreases, and as a result, the robustness of the model will increase. The leftmost point in each line of Fig.~\ref{fig:varrying hyperparameter_b} represents the accuracy and average linear CKA value corresponding to $\theta_5^{L}$, the middle point represents these values for $\theta_5^{M}$, and the rightmost point represents these values for $\theta_5^{H}$ from Fig.~\ref{fig:varrying hyperparameter_a}.

\begin{figure}[!t]
\centering
\includegraphics[width=\linewidth]{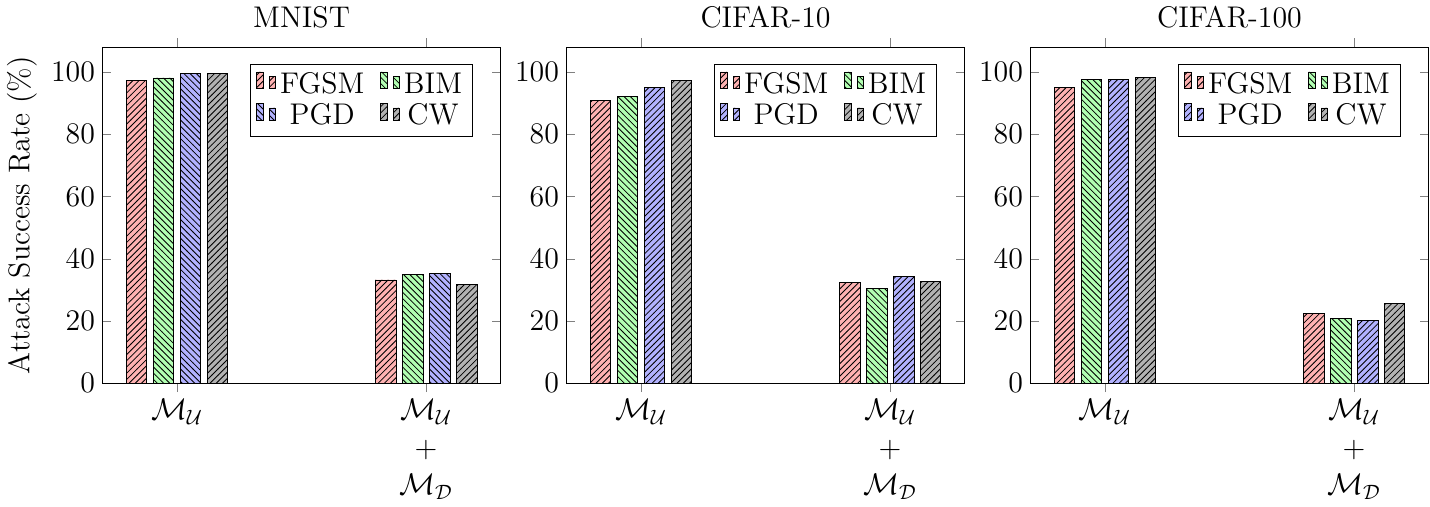}
\caption{The success rate of different adversarial attacks on unprotected neural network model and the ensemble trained with contrast-significant-features method for MNIST, CIFAR-10 and CIFAR-100. The attack parameters are -- FGSM ($\eta = 0.3$), BIM and PGD ($\eta = 0.4, \alpha = 1$, iteration=50), CW ($c = 0.1$, $\kappa = 0.2$, iteration=1000) for MNIST; FGSM ($\eta = 0.02$), BIM and PGD ($\eta = 0.03, \alpha = 1$, iteration=50), CW ($c = 0.1$, $\kappa = 0.2$, iteration=1000) for CIFAR-10 and CIFAR-100\vspace{-0.2cm}}
\label{fig:attack_accuracy_second_2}
\end{figure}

Next, we provide Fig.~\ref{fig:attack_accuracy_second_2} to show the performance of individual model $\mathcal{M}_{\mathcal{U}}$ and the ensemble $\mathcal{M}_{\mathcal{U}}+\mathcal{M}_{\mathcal{D}}$ in the presence of FGSM, BIM, CW and PGD attacks for both MNIST, CIFAR-10 and CIFAR-100. The attack parameters are mentioned in the caption of the figure. We can observe that the success rate is lower in the presence of the detector $\mathcal{M}_{\mathcal{D}}$ because of the diversity between decision boundaries. The $TPR$ and $FPR$ of the ensemble for MNIST, CIFAR-10 and CIFAR-100 is shown in Table~\ref{table:tpr_fpr_second} with the same attack parameters as discussed in the caption of Fig.~\ref{fig:attack_accuracy_second_2}.\vspace{-0.2cm}




\begin{table}[!t]
\centering
\caption{True Positive Rate and False Positive Rate of the ensemble trained with contrast-significant-features method using for FGSM, BIM, PGD and CW attacks considering MNIST, CIFAR-10 and CIFAR-100}\label{table:tpr_fpr_second}
\resizebox{0.85\linewidth}{!}{
\begin{tabular}{c|c|c|c|c|c|c|c|c|}
\cline{2-9}
 & \multicolumn{2}{c|}{\textbf{FGSM}} & \multicolumn{2}{c|}{\textbf{BIM}} & \multicolumn{2}{c|}{\textbf{PGD}} & \multicolumn{2}{c|}{\textbf{CW}} \\ \cline{2-9} 
 & \textbf{TPR} & \textbf{FPR} & \textbf{TPR} & \textbf{FPR} & \textbf{TPR} & \textbf{FPR} & \textbf{TPR} & \textbf{FPR} \\ \hline
\multicolumn{1}{|c|}{\textbf{MNIST}} & 0.62 & 0.07 & 0.61 & 0.08 & 0.65 & 0.07 & 0.66 & 0.06 \\ \hline
\multicolumn{1}{|c|}{\textbf{CIFAR-10}} & 0.64 & 0.28 & 0.67 & 0.31 & 0.64 & 0.29 & 0.68 & 0.28 \\ \hline
\multicolumn{1}{|c|}{\textbf{CIFAR-100}} & 0.76 & 0.34 & 0.79 & 0.41 & 0.79 & 0.32 & 0.81 & 0.36 \\ \hline
\end{tabular}}\vspace{-0.2cm}
\end{table}

\subsubsection{Evaluation against related works}
We study the methods proposed by Kariyappa et al.~\cite{DBLP:journals/corr/abs-1901-09981}, Tramer et al.~\cite{DBLP:conf/iclr/TramerKPGBM18}, and Strauss et al.~\cite{DBLP:journals/corr/abs-1709-03423} to compare the performance of both split-and-shuffle transformation and contrast-significant-method. For the method proposed by Kariyappa et al.~\cite{DBLP:journals/corr/abs-1901-09981}, we trained two models using the Gradient Alignment Loss, and used one model as the unprotected model and the other as the detector model. For the method proposed by Tramer et al.~\cite{DBLP:conf/iclr/TramerKPGBM18}, we generate adversarial examples on the unprotected model and used those adversarial examples to train the detector model. For the method proposed by Strauss et al.~\cite{DBLP:journals/corr/abs-1709-03423}, we train two different randomly initialized models and use one as the unprotected model and the other as the detector model. We investigated the performance of FGSM, BIM, PGD, and CW attacks on these models, along with the split-and-shuffle transformation and contrast-significant-method proposed in this paper. We used the success rate to present a legitimate comparison in Table~\ref{table:comp_zero}. The specific attack parameters used for a balanced comparison are mentioned in the caption of the table. For the split-and-shuffle transformation, we considered the $\mathcal{M}_{\mathcal{U}} + \mathcal{M}_{\mathcal{D}}^{\mathcal{T}_9}$ ensemble. The boldfaced values in each column of Table~\ref{table:comp_zero} represent the best success rate obtained for that category. We can observe that our proposed method performs better for most circumstances than the three methods discussed previously. Moreover, the authors didn't evaluate their strategy against a more potent adversary, which we discuss next.\vspace{-0.3cm}

\begin{table*}[!t]
\centering
\caption{Comparison with related works using success rates considering zero-knowledge adversary $\mathcal{A}_\mathcal{Z}$. The attack parameters are -- FGSM ($\eta = 0.2$), BIM and PGD ($\eta = 0.3, \alpha = 1$, iteration=50), CW ($c = 0.1$, $\kappa = 0.2$, iteration=1000) for MNIST; FGSM ($\eta = 0.02$), BIM and PGD ($\eta = 0.02, \alpha = 1$, iteration=50), CW ($c = 0.1$, $\kappa = 0.2$, iteration=1000) for both CIFAR-10 and CIFAR-100}\label{table:comp_zero}
\resizebox{\linewidth}{!}{
\begin{tabular}{c|c|c|c|c|c|c|c|c|c|c|c|c|}
\cline{2-13}
\textbf{} & \multicolumn{3}{c|}{\textbf{FGSM}} & \multicolumn{3}{c|}{\textbf{BIM}} & \multicolumn{3}{c|}{\textbf{PGD}} & \multicolumn{3}{c|}{\textbf{CW}} \\ \cline{2-13} 
 & \textbf{MNIST} & \textbf{CIFAR-10} & \textbf{CIFAR-100} & \textbf{MNIST} & \textbf{CIFAR-10} & \textbf{CIFAR-100} & \textbf{MNIST} & \textbf{CIFAR-10} & \textbf{CIFAR-100} & \textbf{MNIST} & \textbf{CIFAR-10} & \textbf{CIFAR-100} \\ \hline
\multicolumn{1}{|c|}{\textbf{Kariyappa et al.~\cite{DBLP:journals/corr/abs-1901-09981}}} & 35.6 & 48.9 & 33.1 & 53.3 & 79.1 & 21.1 & 67.9 & 75.5 & 59.1 & 45.6 & 68.1 & 31.1 \\ \hline
\multicolumn{1}{|c|}{\textbf{Tramer et al.~\cite{DBLP:conf/iclr/TramerKPGBM18}}} & 31.1 & 32.2 & \textbf{14.9} & \textbf{19.1} & \textbf{20.8} & \textbf{11.9} & 37.6 & 48.3 & 29.2 & 32.7 & 38.6 & 30.4 \\ \hline
\multicolumn{1}{|c|}{\textbf{Strauss et al.~\cite{DBLP:journals/corr/abs-1709-03423}}} & 48.0 & 14.4 & 16.2 & 62.3 & 34.7 & 21.3 & 71.3 & 37.8 & 48.3 & 49.8 & 63.7 & 28.6 \\ \hline
\multicolumn{1}{|c|}{\textbf{Split-and-Shuffle Transformation}} & \textbf{14.6} & 19.1 & 16.7 & 23.1 & 25.6 & 16.7 & \textbf{23.5} & \textbf{27.8} & \textbf{16.9} & \textbf{24.3} & \textbf{31.9} & \textbf{20.9} \\ \hline
\multicolumn{1}{|c|}{\textbf{Contrast-Significant-Feature}} & 32.4 & \textbf{13.8} & 18.0 & 35.1 & 30.6 & 20.9 & 35.3 & 34.4 & 20.3 & 31.7 & 32.7 & 25.7 \\ \hline
\end{tabular}}\vspace{-0.3cm}
\end{table*}

\subsection{Evaluation in the presence of perfect knowledge adversary $\mathcal{A}_\mathcal{P}$}
\subsubsection{Evaluation for split-and-shuffle input transformation}\label{sec:results:perfect:input_transformation}
In order to analyze the robustness of split-and-shuffle transformation in the presence of $\mathcal{A}_\mathcal{P}$, we consider the same dataset and ensembles, as discussed in Section~\ref{sec:results_input}. The objective of an adversary in this scenario is to create adversarial examples considering parameters of both the original and the detector model, unlike $\mathcal{A}_\mathcal{Z}$, which considers parameters only from the original model. In order to achieve the objective, we consider a modified version of the FGSM attack to create adversarial examples considering both the original and the detector model. The adversarial example $x_{adv}$ in this scenario can be created from a clean image $x_{init}$ using the following equation\vspace{-0.1cm}
\begin{align*}
    x_{adv} = x_{init} + \eta \cdot \{\beta \cdot sign(\nabla_x J_{1}(x_{init}, w)) + \\(1-\beta) \cdot sign(\mathcal{T}_n^{-1}(\nabla_x J_2(\mathcal{T}_n(x_{init}), \overline{w})))\}
\end{align*}
where $J_1(\cdot)$, $J_2(\cdot)$ are the loss functions and $w$, $\overline{w}$ are the learned parameters of the original model and the detector model respectively. We apply the split-and-shuffle transformation $\mathcal{T}_n(\cdot)$ on the input data point before computing the loss function $J_2(\cdot)$, as the detector model is trained on the transformed data points. We apply an inverse transformation $\mathcal{T}_n^{-1}(\cdot)$ on the gradients computed from the detector model before adding it to the original image such that the overall perturbation on the original image can have an effect on both the models. $\mathcal{T}_n^{-1}(\cdot)$ is the inverse of the shuffling operation used for $\mathcal{T}_n(\cdot)$. The parameter $\beta \in [0, 1]$ controls the effect of perturbations on both the models. A higher value of $\beta$ generates perturbation in such a way that the adversarial sample is more prone to fool the original model. The parameter $\eta > 0$ is the attack strength. In the following analysis, we have considered $\beta = 0.5$ to generate perturbations assigning equal importance to both the original and detector model.

\begin{figure}
    \centering
    \includegraphics[width=\linewidth]{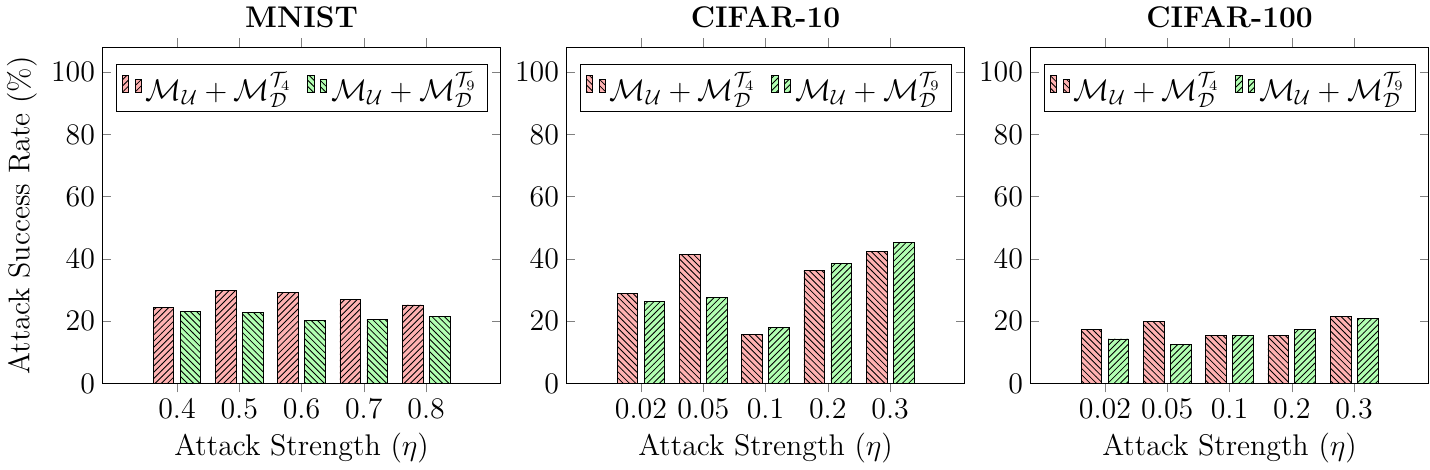}
    \caption{Success Rate of Perfect Knowledge of Adversary for different attack strength considering MNIST, CIFAR-10 and CIFAR-100 on two ensembles $\mathcal{M}_{\mathcal{U}} + \mathcal{M}_{\mathcal{D}}^{\mathcal{T}_4}$ and $\mathcal{M}_{\mathcal{U}} + \mathcal{M}_{\mathcal{D}}^{\mathcal{T}_9}$ trained with different split-and-shuffle input transformation\vspace{-0.2cm}}
    \label{fig:combine_attack}
\end{figure}

The attack success rate for different attack strengths on ensembles $\mathcal{M}_\mathcal{U} + \mathcal{M}_\mathcal{D}^{\mathcal{T}_4}$ and $\mathcal{M}_\mathcal{U} + \mathcal{M}_\mathcal{D}^{\mathcal{T}_9}$ considering MNIST, CIFAR-10 and CIFAR-100 is shown in Fig.~\ref{fig:combine_attack}. We can observe that with an increase in attack strength, the success rate does not increase beyond the zero knowledge adversary $(\mathcal{A}_\mathcal{Z})$ for MNIST but gets better for CIFAR-10 and CIFAR-100. However, the increase in attack strength also increases the amount of perturbation added to the clean image. To show the effect of perturbations on the input images, we present clean and corresponding adversarial examples for each combination of attack strength, ensembles used, and dataset in Table~\ref{table:combined_attack}. The adversarial examples presented in the table have the minimum perturbation, among other examples, within the same combination. The table also shows the $L_2$-norm of the perturbation (i.e., $|x_{adv} - x_{init}|$) for each example. We can visually distinguish between a clean and the corresponding adversarial examples as the attack strength increases, which fails the primary motive of adversarial example generation.\vspace{-0.2cm}

\begin{table*}[!t]
\centering
\caption{The clean image and corresponding adversarial example with $L_2$-norm of perturbation for different attack strength considering ensembles $\mathcal{M}_{\mathcal{U}} + \mathcal{M}_{\mathcal{D}}^{\mathcal{T}_4}$ and $\mathcal{M}_{\mathcal{U}} + \mathcal{M}_{\mathcal{D}}^{\mathcal{T}_9}$ trained with different split-and-shuffle input transformation for MNIST, CIFAR-10 and CIFAR-100}\label{table:combined_attack}
\resizebox{0.9\linewidth}{!}{
\begin{tabular}{cc|c|c|c|c|c|}
\cline{3-7}
\multicolumn{1}{c}{} & \multicolumn{1}{c|}{} & \multicolumn{1}{c|}{$\eta = 0.4$} & \multicolumn{1}{c|}{$\eta = 0.5$} & \multicolumn{1}{c|}{$\eta = 0.6$} & \multicolumn{1}{c|}{$\eta = 0.7$} & \multicolumn{1}{c|}{$\eta = 0.8$} \\ \hline
\multicolumn{1}{|c|}{\multirow{6}{*}{\textbf{MNIST}}} & $\mathcal{M}_{\mathcal{U}} + \mathcal{M}_{\mathcal{D}}^{\mathcal{T}_4}$
& \begin{tabular}[c]{@{}c@{}}\subfloat{\includegraphics[width=0.06\linewidth]{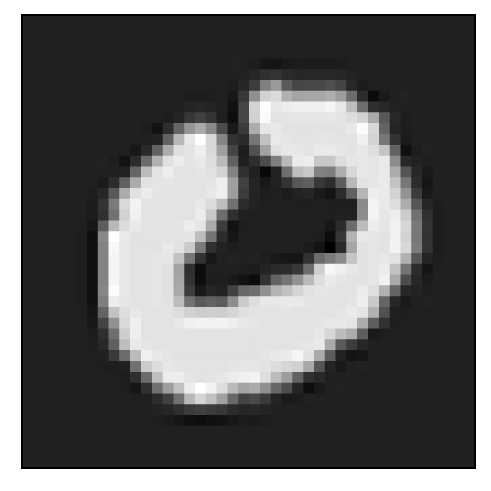}}\subfloat{\includegraphics[width=0.06\linewidth]{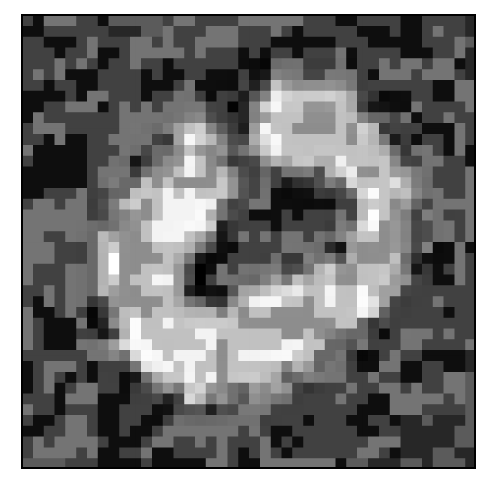}} \\ $L_2$-norm: 6.87\end{tabular}
& \begin{tabular}[c]{@{}c@{}}\subfloat{\includegraphics[width=0.06\linewidth]{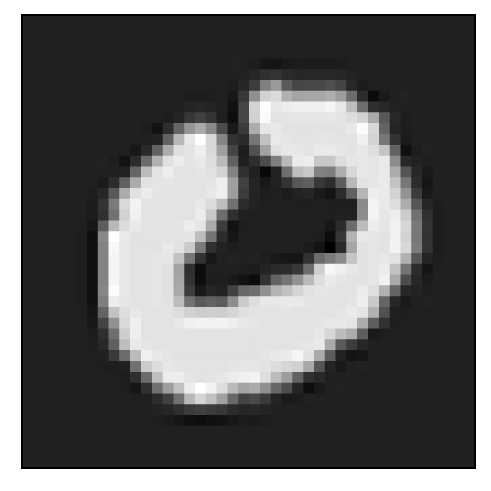}} \subfloat{\includegraphics[width=0.06\linewidth]{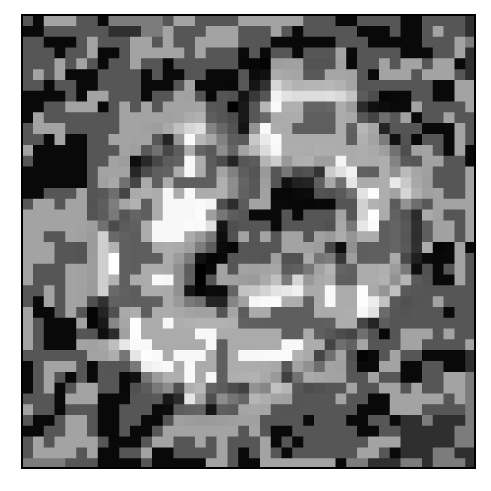}}\\ $L_2$-norm: 15.47\end{tabular} & \begin{tabular}[c]{@{}c@{}}\subfloat{\includegraphics[width=0.06\linewidth]{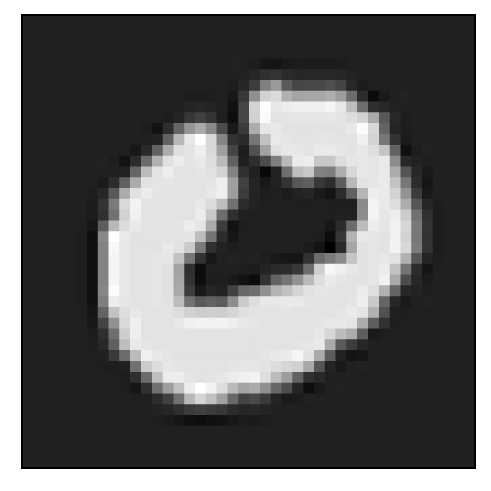}} \subfloat{\includegraphics[width=0.06\linewidth]{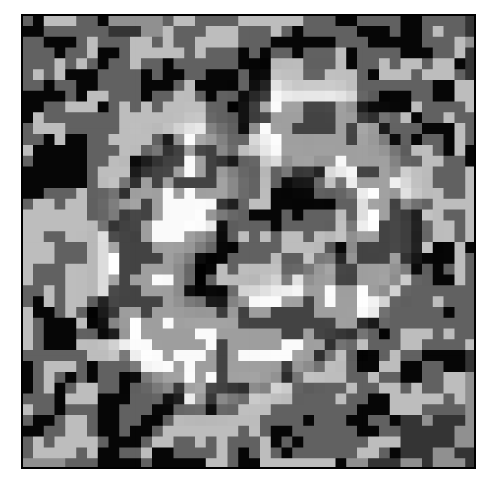}} \\ $L_2$-norm: 25.79\end{tabular} & \begin{tabular}[c]{@{}c@{}}\subfloat{\includegraphics[width=0.06\linewidth]{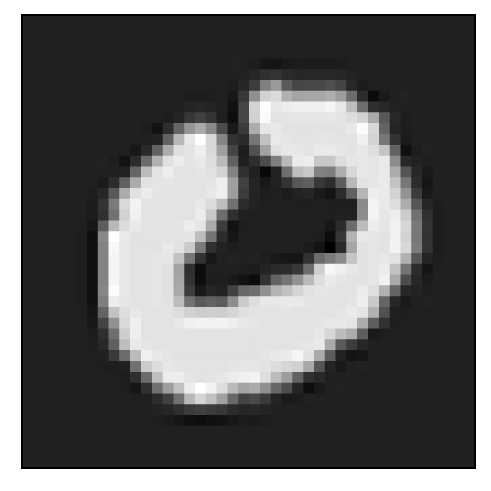}} \subfloat{\includegraphics[width=0.06\linewidth]{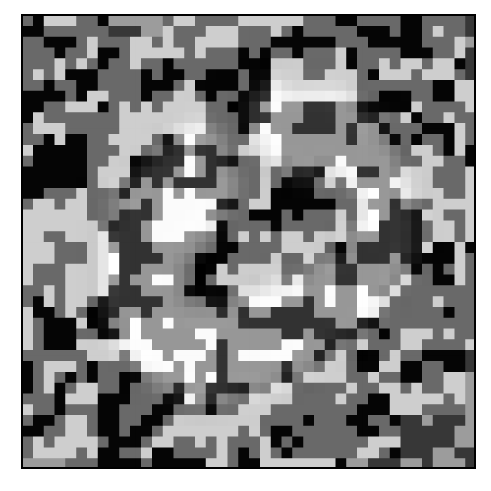}} \\$L_2$-norm: 37.82\end{tabular} &\begin{tabular}[c]{@{}c@{}}\subfloat{\includegraphics[width=0.06\linewidth]{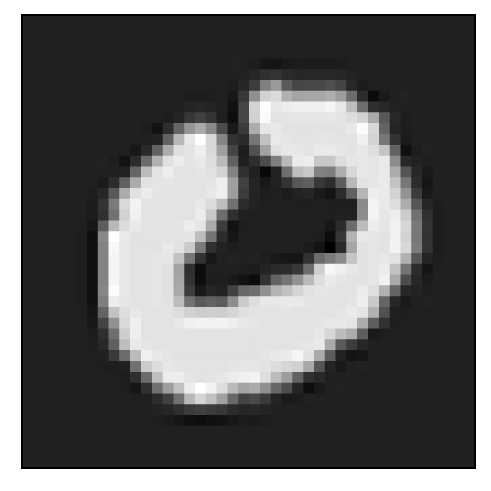}} \subfloat{\includegraphics[width=0.06\linewidth]{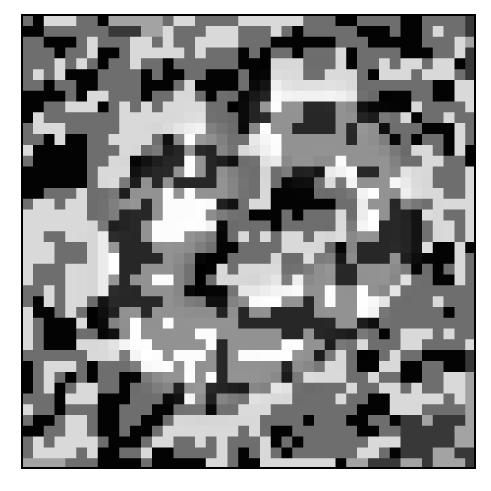}}\\$L_2$-norm: 51.58\end{tabular}\\ \cline{2-7} 
\multicolumn{1}{|c|}{} & $\mathcal{M}_{\mathcal{U}} + \mathcal{M}_{\mathcal{D}}^{\mathcal{T}_9}$
& \begin{tabular}[c]{@{}c@{}}\subfloat{\includegraphics[width=0.06\linewidth]{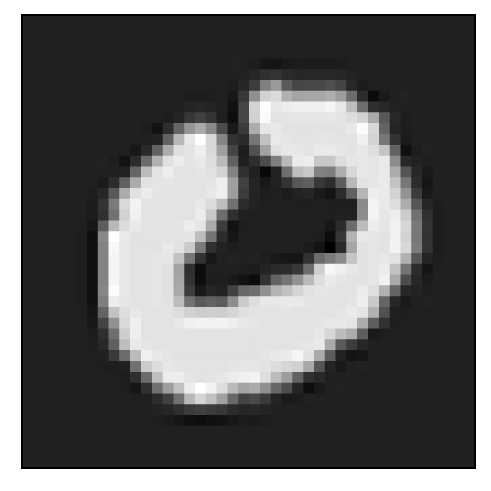}} \subfloat{\includegraphics[width=0.06\linewidth]{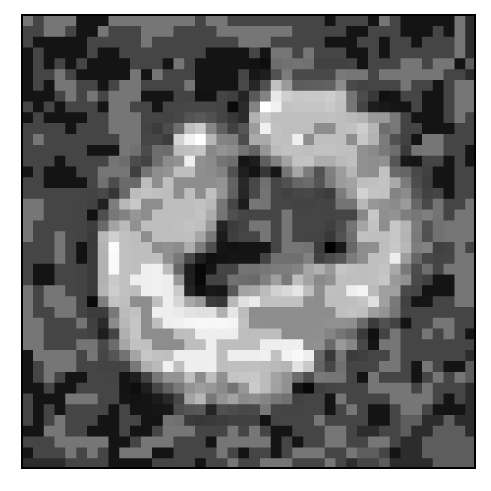}} \\ $L_2$-norm: 7.25\end{tabular}& \begin{tabular}[c]{@{}c@{}}\subfloat{\includegraphics[width=0.06\linewidth]{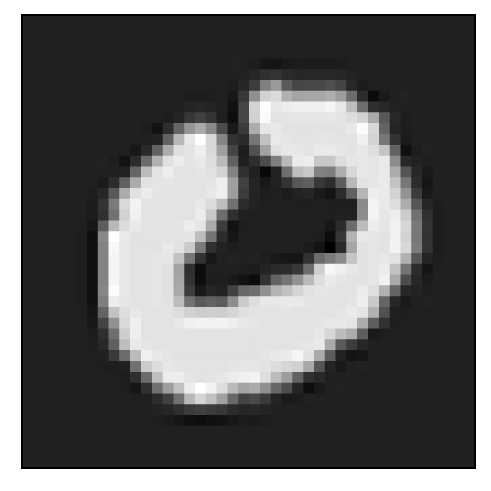}} \subfloat{\includegraphics[width=0.06\linewidth]{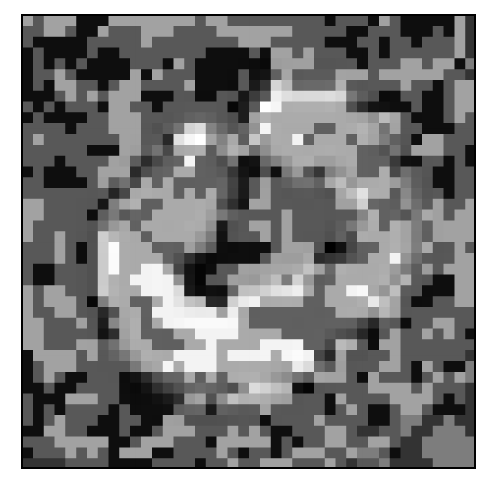}} \\ $L_2$-norm: 16.32\end{tabular}& \begin{tabular}[c]{@{}c@{}}\subfloat{\includegraphics[width=0.06\linewidth]{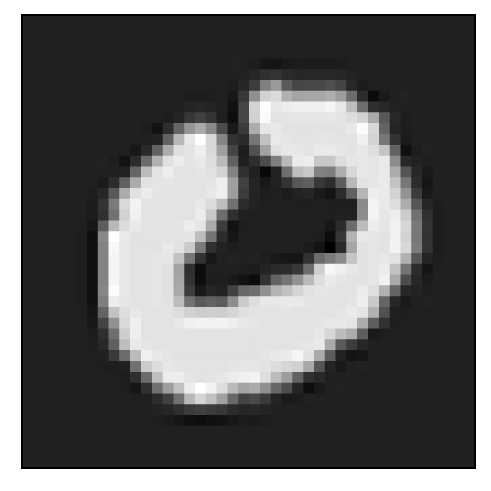}} \subfloat{\includegraphics[width=0.06\linewidth]{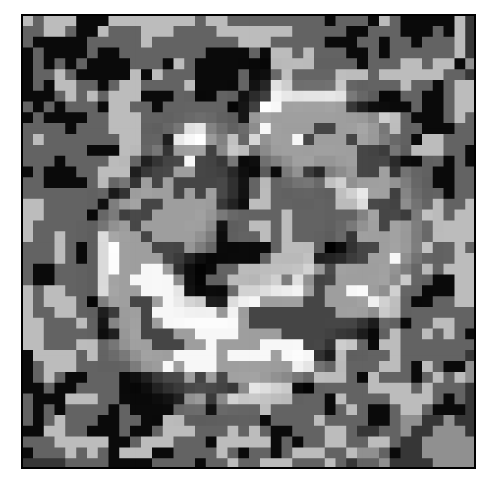}} \\ $L_2$-norm: 27.21\end{tabular}& \begin{tabular}[c]{@{}c@{}}\subfloat{\includegraphics[width=0.06\linewidth]{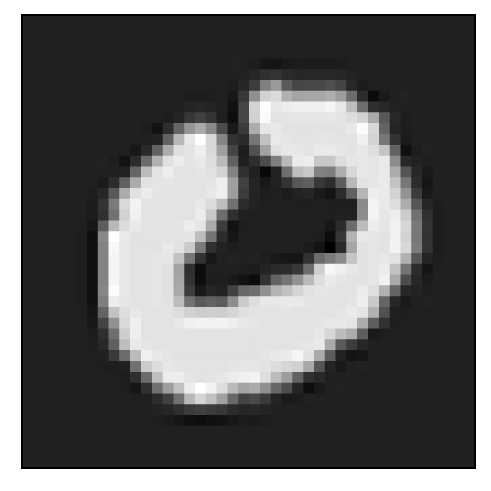}} \subfloat{\includegraphics[width=0.06\linewidth]{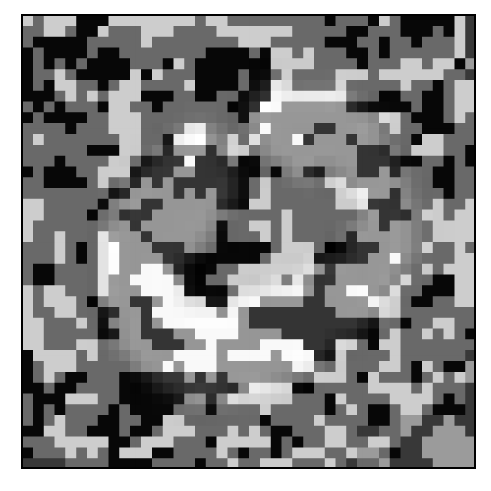}} \\ $L_2$-norm: 39.91\end{tabular}& \begin{tabular}[c]{@{}c@{}}\subfloat{\includegraphics[width=0.06\linewidth]{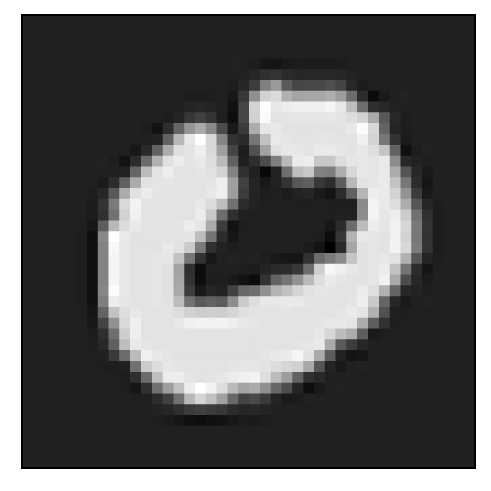}} \subfloat{\includegraphics[width=0.06\linewidth]{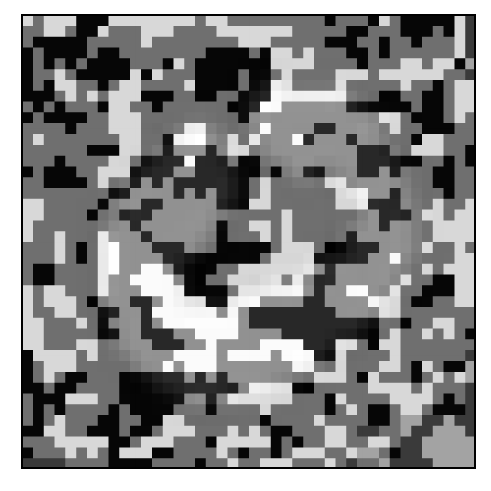}} \\ $L_2$-norm: 54.41\end{tabular}\\ \hline \\
\cline{3-7}
\multicolumn{1}{c}{} & \multicolumn{1}{c|}{} & \multicolumn{1}{c|}{$\eta = 0.02$} & \multicolumn{1}{c|}{$\eta = 0.05$} & \multicolumn{1}{c|}{$\eta = 0.1$} & \multicolumn{1}{c|}{$\eta = 0.2$} & \multicolumn{1}{c|}{$\eta = 0.3$} \\ \hline
\multicolumn{1}{|c|}{\multirow{6}{*}{\textbf{CIFAR-10}}} & $\mathcal{M}_{\mathcal{U}} + \mathcal{M}_{\mathcal{D}}^{\mathcal{T}_4}$
& \begin{tabular}[c]{@{}c@{}}\subfloat{\includegraphics[width=0.06\linewidth]{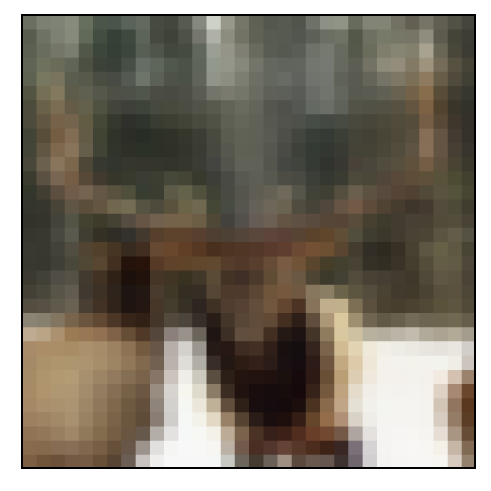}} \subfloat{\includegraphics[width=0.06\linewidth]{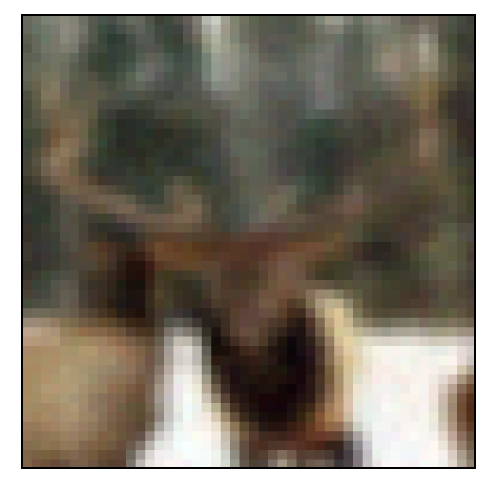}} \\ $L_2$-norm: 0.73\end{tabular}& \begin{tabular}[c]{@{}c@{}}\subfloat{\includegraphics[width=0.06\linewidth]{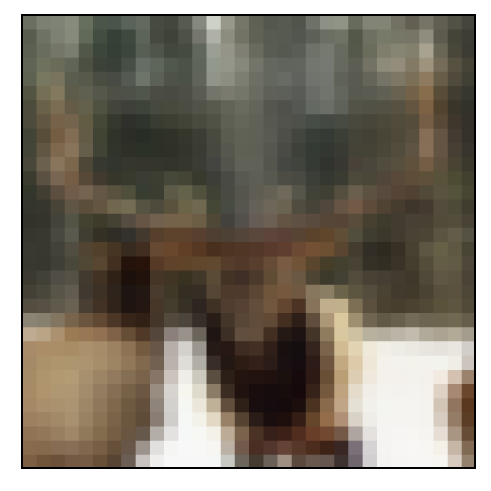}} \subfloat{\includegraphics[width=0.06\linewidth]{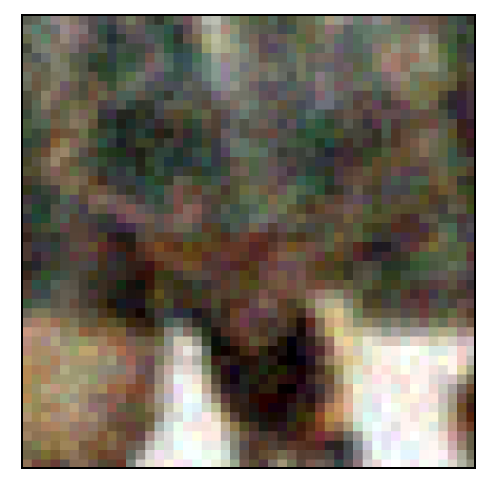}} \\ $L_2$-norm: 2.57\end{tabular}& \begin{tabular}[c]{@{}c@{}}\subfloat{\includegraphics[width=0.06\linewidth]{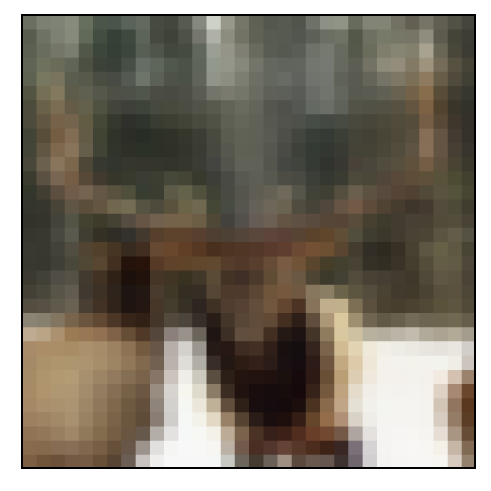}} \subfloat{\includegraphics[width=0.06\linewidth]{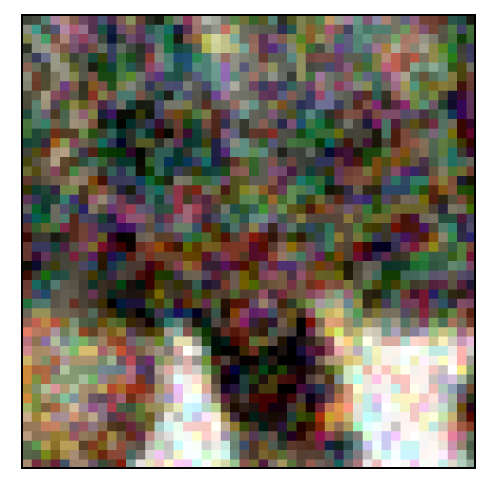}} \\ $L_2$-norm: 6.24\end{tabular}& \begin{tabular}[c]{@{}c@{}}\subfloat{\includegraphics[width=0.06\linewidth]{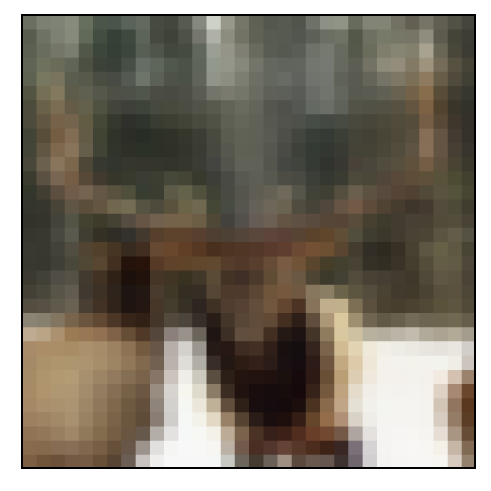}} \subfloat{\includegraphics[width=0.06\linewidth]{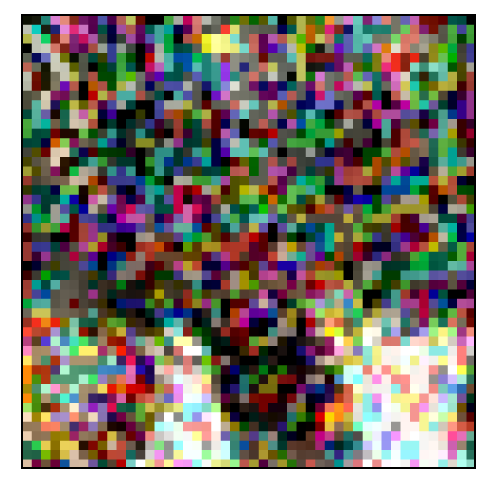}} \\ $L_2$-norm: 13.59\end{tabular}& \begin{tabular}[c]{@{}c@{}}\subfloat{\includegraphics[width=0.06\linewidth]{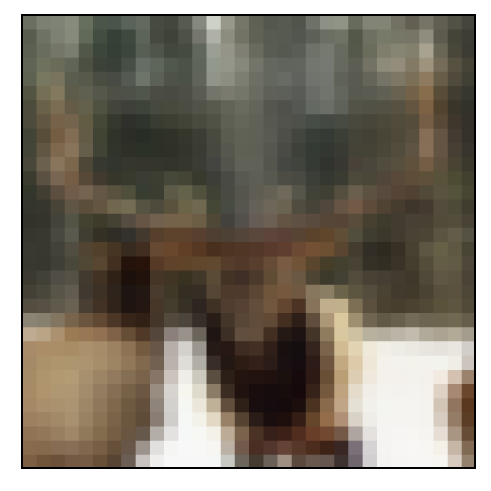}} \subfloat{\includegraphics[width=0.06\linewidth]{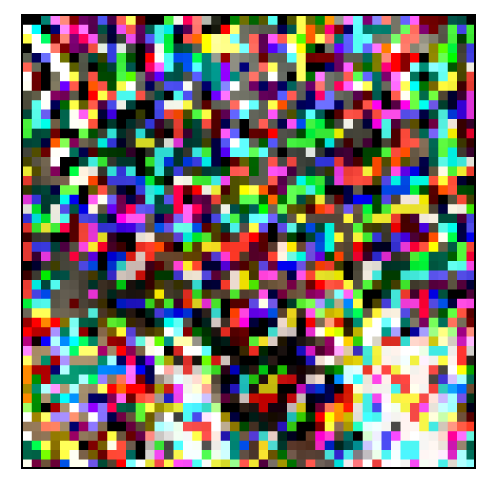}} \\ $L_2$-norm: 24.61\end{tabular}\\ \cline{2-7} 
\multicolumn{1}{|c|}{} & $\mathcal{M}_{\mathcal{U}} + \mathcal{M}_{\mathcal{D}}^{\mathcal{T}_9}$
& \begin{tabular}[c]{@{}c@{}} \subfloat{\includegraphics[width=0.06\linewidth]{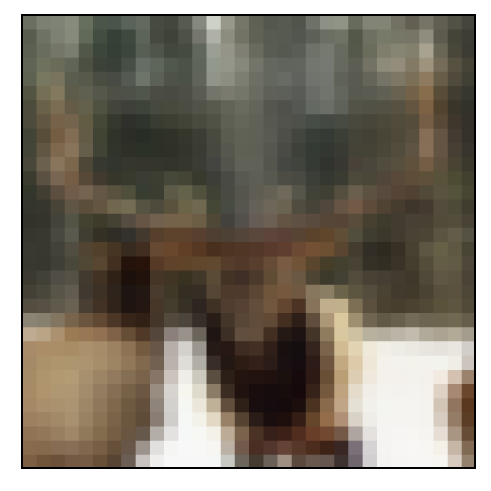}} \subfloat{\includegraphics[width=0.06\linewidth]{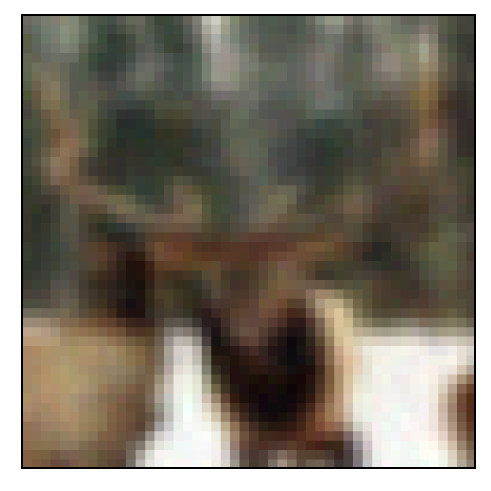}} \\ $L_2$-norm: 0.74\end{tabular}& \begin{tabular}[c]{@{}c@{}}\subfloat{\includegraphics[width=0.06\linewidth]{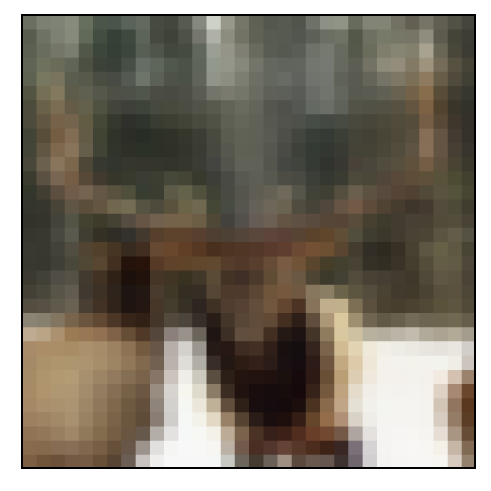}} \subfloat{\includegraphics[width=0.06\linewidth]{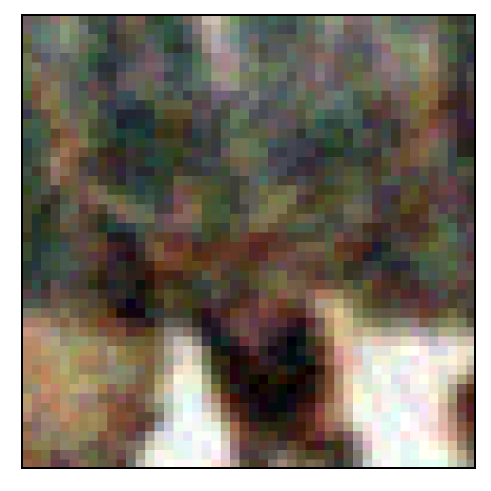}} \\ $L_2$-norm: 2.59\end{tabular}& \begin{tabular}[c]{@{}c@{}}\subfloat{\includegraphics[width=0.06\linewidth]{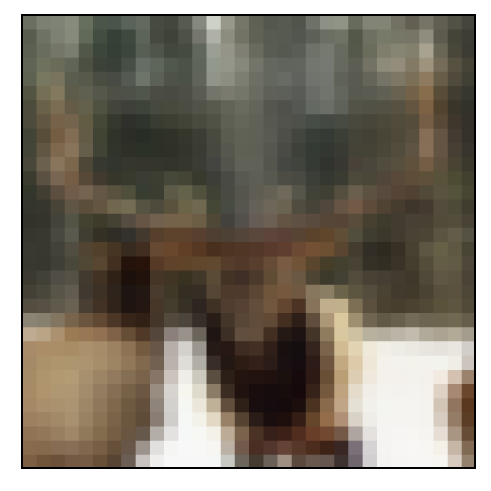}} \subfloat{\includegraphics[width=0.06\linewidth]{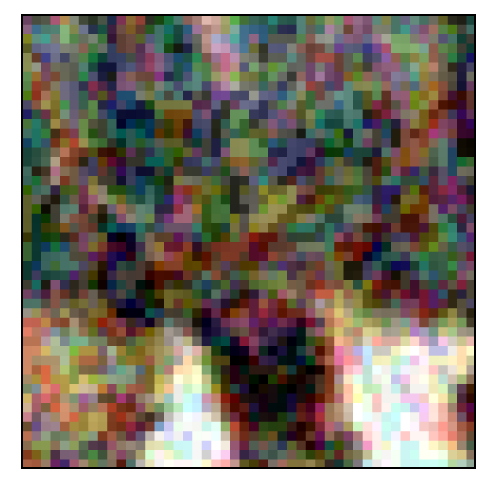}} \\ $L_2$-norm: 6.31\end{tabular}& \begin{tabular}[c]{@{}c@{}}\subfloat{\includegraphics[width=0.06\linewidth]{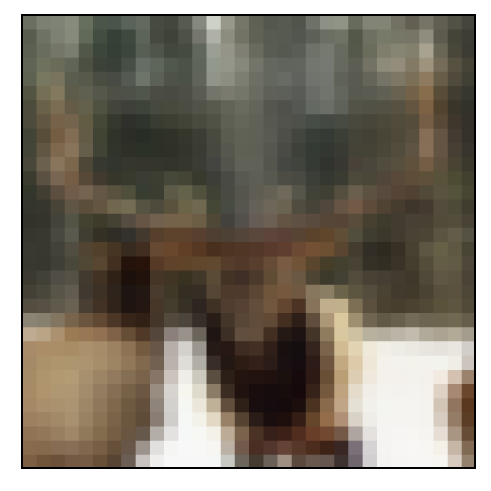}} \subfloat{\includegraphics[width=0.06\linewidth]{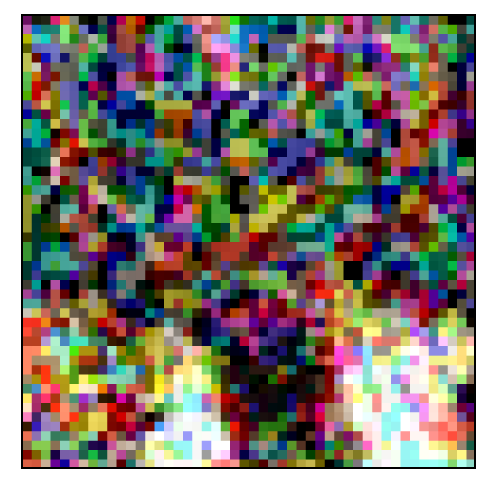}} \\ $L_2$-norm: 13.73\end{tabular}& \begin{tabular}[c]{@{}c@{}}\subfloat{\includegraphics[width=0.06\linewidth]{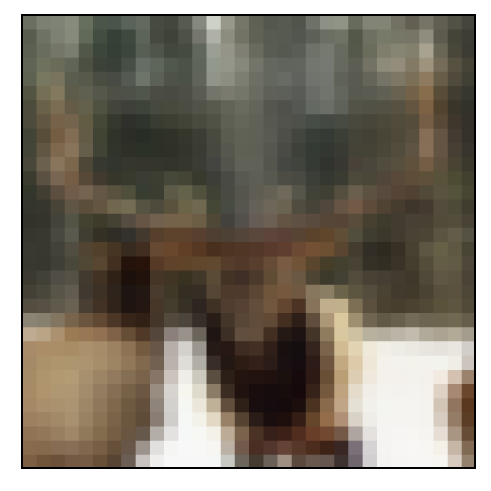}} \subfloat{\includegraphics[width=0.06\linewidth]{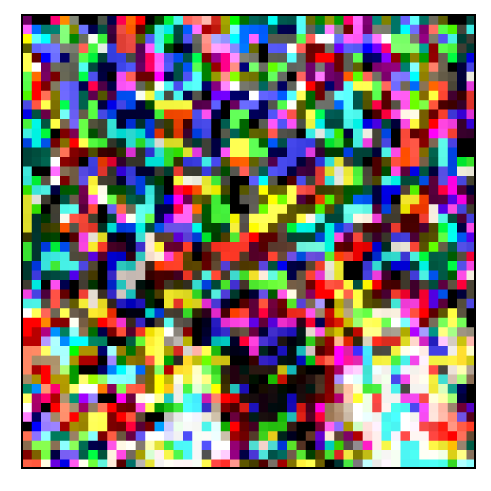}} \\ $L_2$-norm: 24.86\end{tabular}\\ \hline \\
\cline{3-7}
\multicolumn{1}{c}{} & \multicolumn{1}{c|}{} & \multicolumn{1}{c|}{$\eta = 0.02$} & \multicolumn{1}{c|}{$\eta = 0.05$} & \multicolumn{1}{c|}{$\eta = 0.1$} & \multicolumn{1}{c|}{$\eta = 0.2$} & \multicolumn{1}{c|}{$\eta = 0.3$} \\ \hline
\multicolumn{1}{|c|}{\multirow{6}{*}{\textbf{CIFAR-100}}} & $\mathcal{M}_{\mathcal{U}} + \mathcal{M}_{\mathcal{D}}^{\mathcal{T}_4}$
& \begin{tabular}[c]{@{}c@{}}\subfloat{\includegraphics[width=0.06\linewidth]{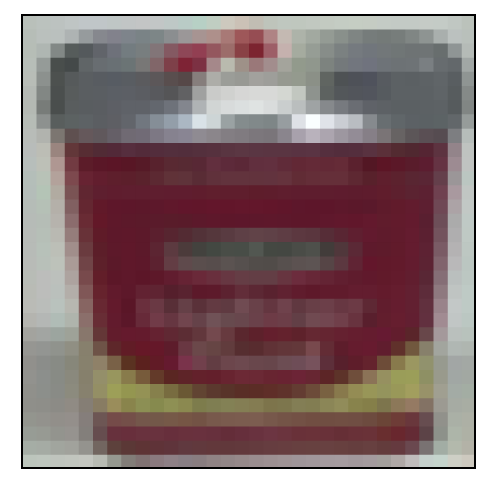}} \subfloat{\includegraphics[width=0.06\linewidth]{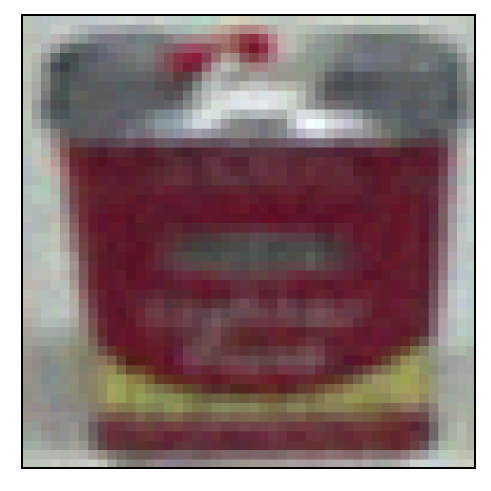}} \\ $L_2$-norm: 0.74\end{tabular}& \begin{tabular}[c]{@{}c@{}}\subfloat{\includegraphics[width=0.06\linewidth]{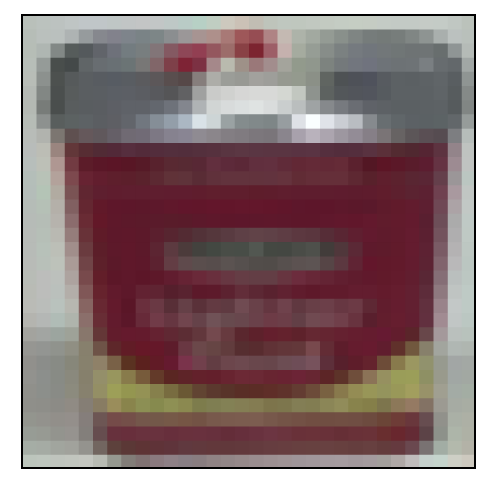}} \subfloat{\includegraphics[width=0.06\linewidth]{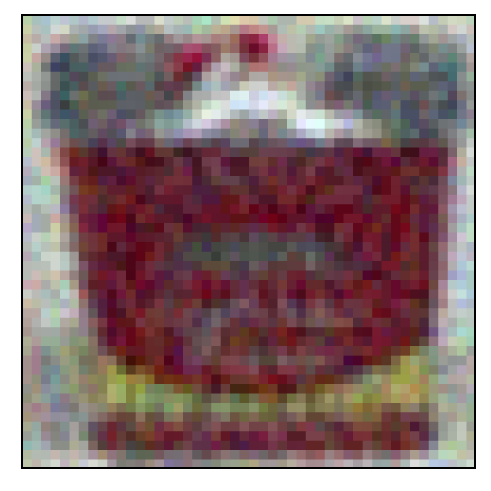}} \\ $L_2$-norm: 2.61\end{tabular}& \begin{tabular}[c]{@{}c@{}}\subfloat{\includegraphics[width=0.06\linewidth]{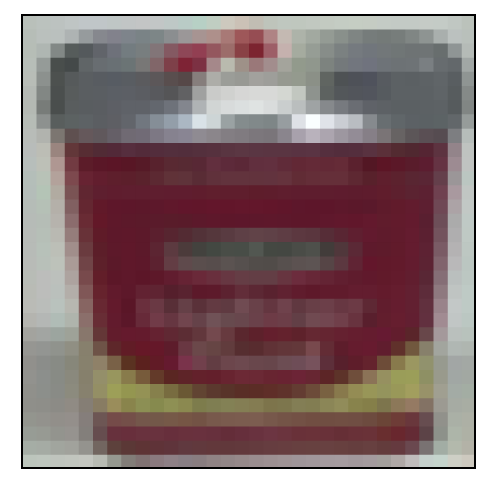}} \subfloat{\includegraphics[width=0.06\linewidth]{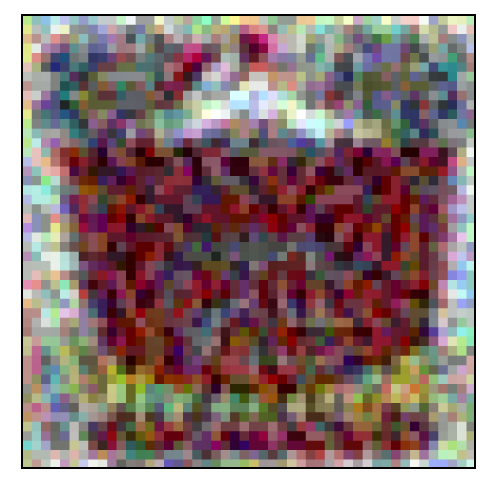}} \\ $L_2$-norm: 6.35\end{tabular}& \begin{tabular}[c]{@{}c@{}}\subfloat{\includegraphics[width=0.06\linewidth]{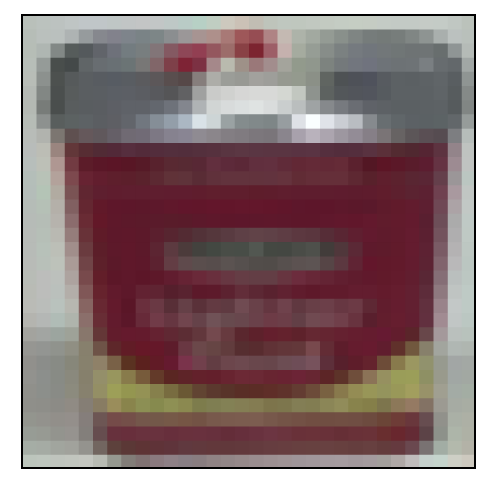}} \subfloat{\includegraphics[width=0.06\linewidth]{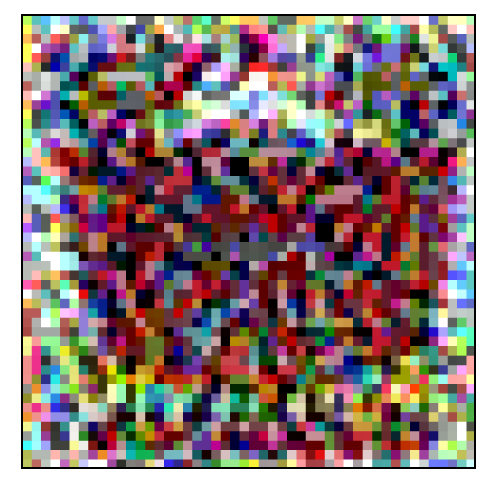}} \\ $L_2$-norm: 13.83\end{tabular}& \begin{tabular}[c]{@{}c@{}}\subfloat{\includegraphics[width=0.06\linewidth]{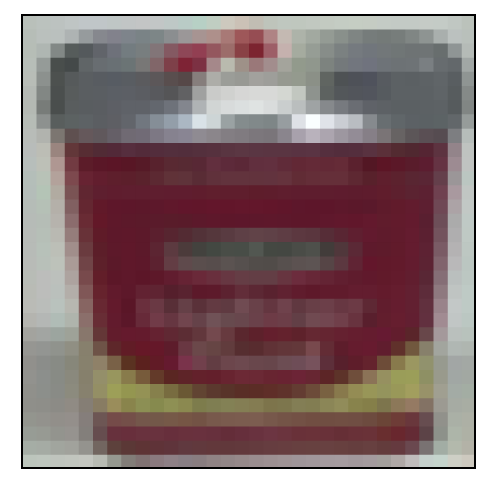}} \subfloat{\includegraphics[width=0.06\linewidth]{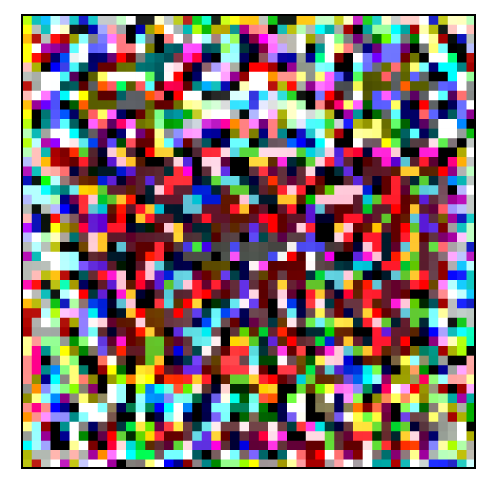}} \\ $L_2$-norm: 25.04\end{tabular}\\ \cline{2-7} 
\multicolumn{1}{|c|}{} & $\mathcal{M}_{\mathcal{U}} + \mathcal{M}_{\mathcal{D}}^{\mathcal{T}_9}$
& \begin{tabular}[c]{@{}c@{}} \subfloat{\includegraphics[width=0.06\linewidth]{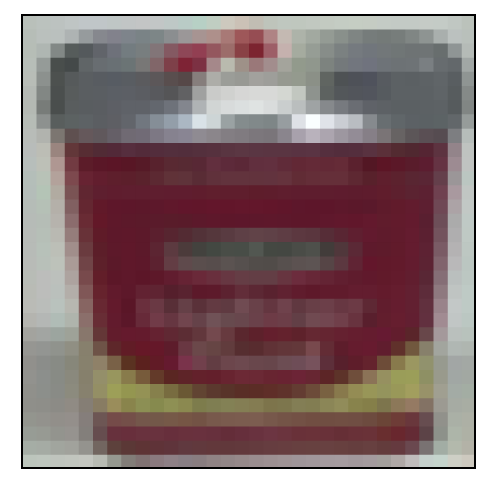}} \subfloat{\includegraphics[width=0.06\linewidth]{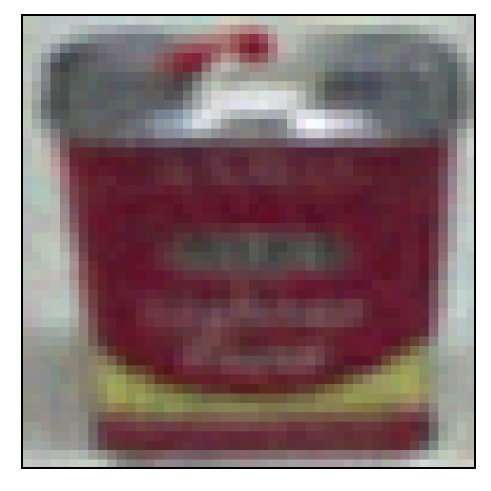}} \\ $L_2$-norm: 0.77\end{tabular}& \begin{tabular}[c]{@{}c@{}}\subfloat{\includegraphics[width=0.06\linewidth]{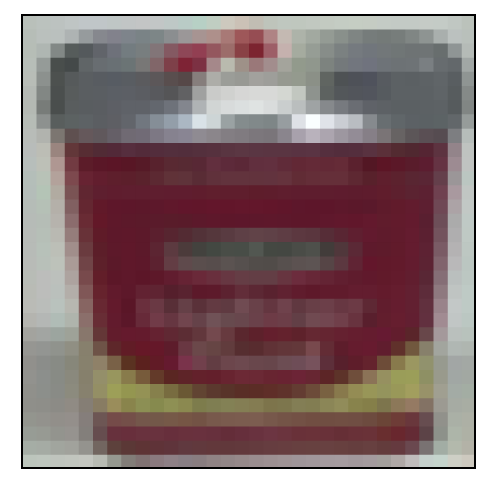}} \subfloat{\includegraphics[width=0.06\linewidth]{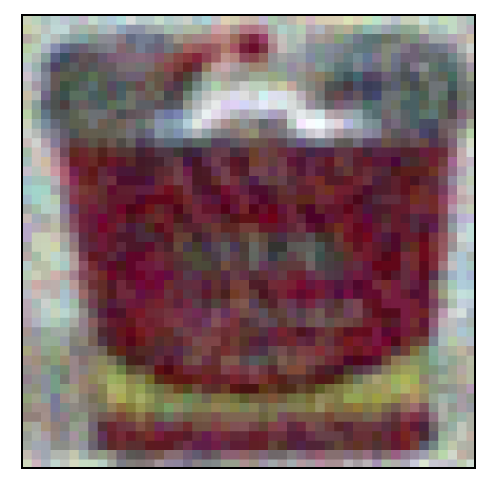}} \\ $L_2$-norm: 2.71\end{tabular}& \begin{tabular}[c]{@{}c@{}}\subfloat{\includegraphics[width=0.06\linewidth]{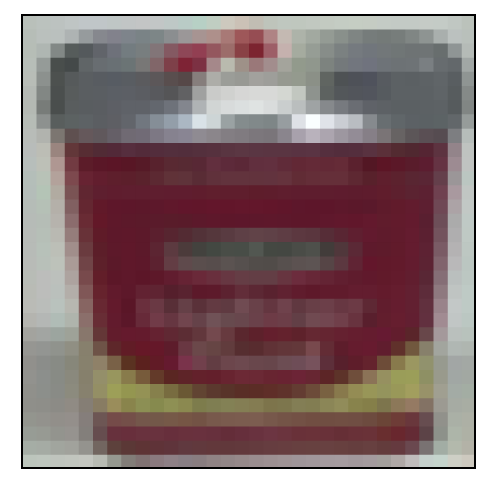}} \subfloat{\includegraphics[width=0.06\linewidth]{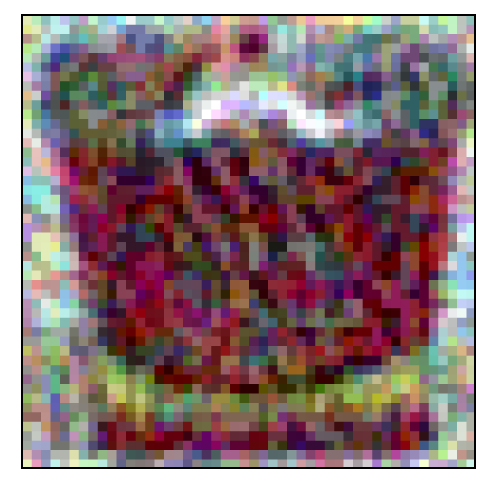}} \\ $L_2$-norm: 6.56\end{tabular}& \begin{tabular}[c]{@{}c@{}}\subfloat{\includegraphics[width=0.06\linewidth]{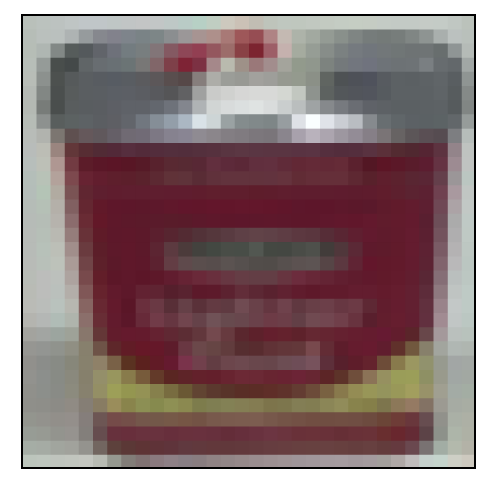}} \subfloat{\includegraphics[width=0.06\linewidth]{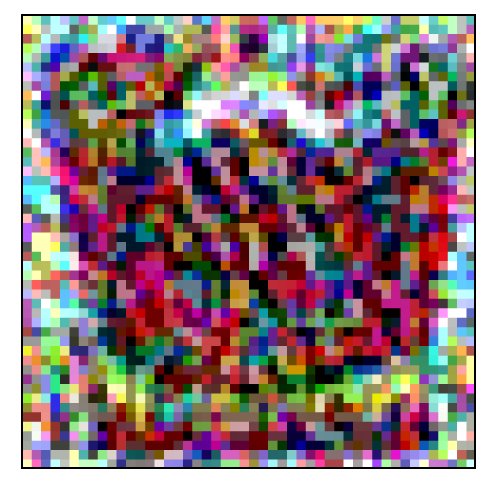}} \\ $L_2$-norm: 14.28\end{tabular}& \begin{tabular}[c]{@{}c@{}}\subfloat{\includegraphics[width=0.06\linewidth]{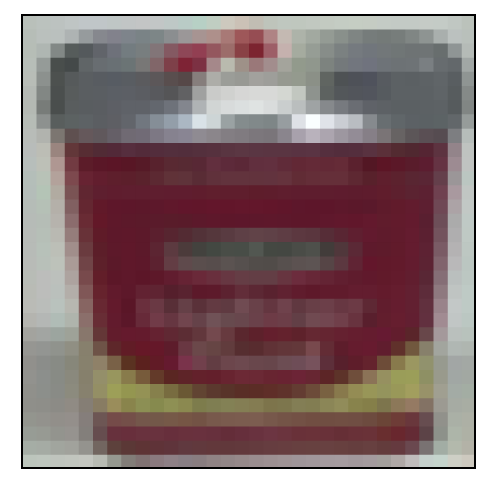}} \subfloat{\includegraphics[width=0.06\linewidth]{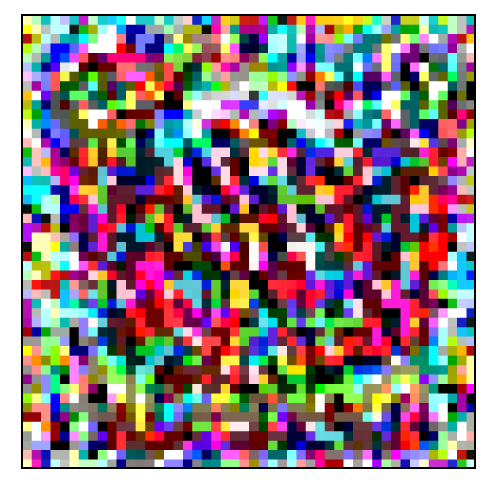}} \\ $L_2$-norm: 25.87\end{tabular} \\ \hline
\end{tabular}}\vspace{-0.3cm}
\end{table*}

\subsubsection{Evaluation for contrast-significant-features method}
\begin{figure}[!t]
    \centering
    \includegraphics[width=\linewidth]{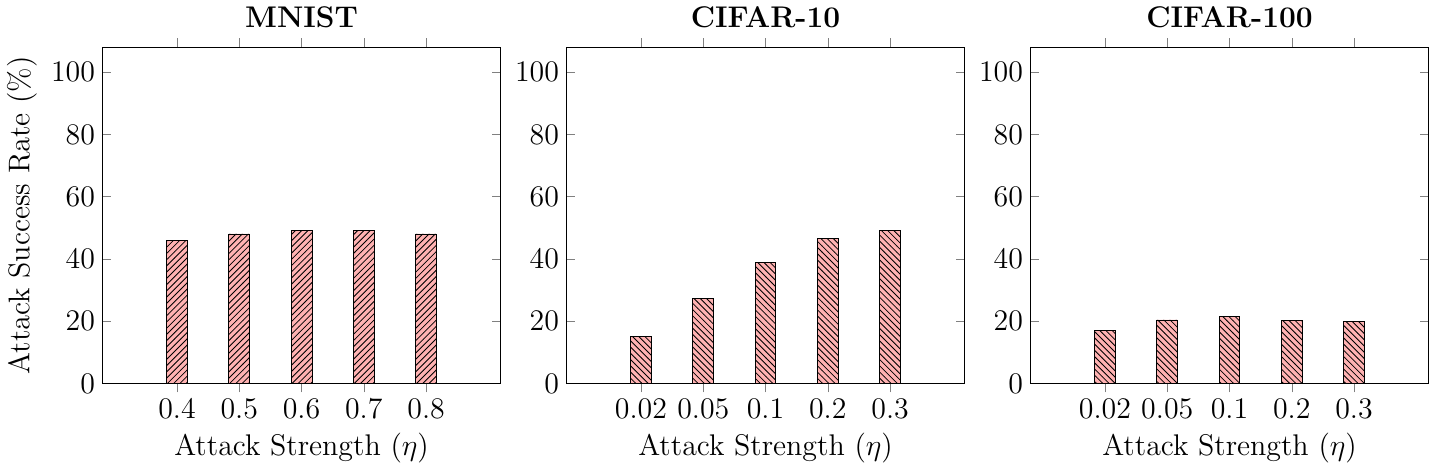}
    \caption{Success Rate of Perfect Knowledge of Adversary for different attack strength considering MNIST, CIFAR-10 and CIFAR-100 on the ensemble $\mathcal{M}_{\mathcal{U}} + \mathcal{M}_{\mathcal{D}}$ trained with contrast-significant-features method\vspace{-0.3cm}}
    \label{fig:combine_attack_2}
\end{figure}

In order to analyze the robustness of contrast-significant-features method, we follow the same experiments as described in Section~\ref{sec:results:perfect:input_transformation}. In this scenario, the adversarial example $x_{adv}$ is created from a clean image $x_{init}$ using the following equation\vspace{-0.1cm}
\begin{align*}
    x_{adv} = x_{init} + \eta \cdot \{\beta \cdot sign(\nabla_x J_{1}(x_{init}, w)) + \\(1-\beta) \cdot sign(\nabla_x J_2(x_{init}, \overline{w}))\}
\end{align*}
where $J_1(\cdot)$, $J_2(\cdot)$ are the loss functions and $w$, $\overline{w}$ are the learned parameters of the original model and the detector model respectively. Like before, we use $\beta = 0.5$ to give equal importance to both original and detector models while creating adversarial examples. We increase the attack strength $\eta$ gradually and observe the success rate of attack on the ensemble $\mathcal{M}_\mathcal{U} + \mathcal{M}_\mathcal{D}$ for MNIST, CIFAR-10 and CIFAR-100. The observation is shown in Fig.~\ref{fig:combine_attack_2}. Also, in this scenario, we can observe that with an increase in attack strength, the success rate gets better than the zero knowledge adversary $(\mathcal{A}_\mathcal{Z})$ for the CIFAR-10 dataset. However, the increase in attack strength also increases the amount of perturbation added to the clean image. To show the effect of perturbations on the input images, we present clean and corresponding adversarial examples for each combination of attack strength and dataset in Table~\ref{table:combined_attack_2}. The table also shows the $L_2$-norm of the perturbation for each example. We can visually distinguish between a clean and the corresponding adversarial examples as the attack strength increases, which again fails the primary motive of generating adversarial examples.\vspace{-0.3cm}

\begin{table*}[!t]
\centering
\caption{The clean image and corresponding adversarial example with $L_2$-norm of perturbation for different attack strength considering the ensemble $\mathcal{M}_{\mathcal{U}} + \mathcal{M}_{\mathcal{D}}$ trained with contrast-significant-features method for MNIST, CIFAR-10 and CIFAR-100\label{table:combined_attack_2}}
\begin{tabular}{cccccc}
\cline{2-6}
\multicolumn{1}{c|}{} & \multicolumn{1}{c|}{$\eta = 0.4$} & \multicolumn{1}{c|}{$\eta = 0.5$} & \multicolumn{1}{c|}{$\eta = 0.6$} & \multicolumn{1}{c|}{$\eta = 0.7$} & \multicolumn{1}{c|}{$\eta = 0.8$} \\ \hline
\multicolumn{1}{|c|}{\textbf{MNIST}}
& \multicolumn{1}{c|}{\begin{tabular}[c]{@{}c@{}}
\subfloat{\includegraphics[width=0.06\linewidth]{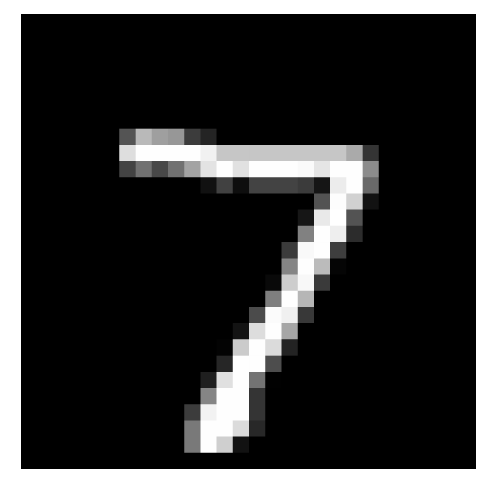}} \subfloat{\includegraphics[width=0.06\linewidth]{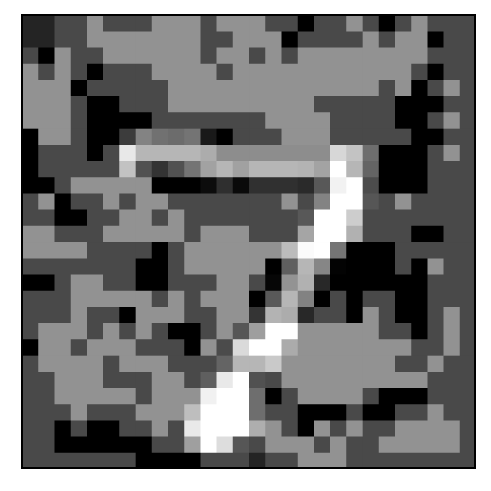}} \\ $L_2$-norm: 8.29\end{tabular}
}
& \multicolumn{1}{c|}{\begin{tabular}[c]{@{}c@{}}
\subfloat{\includegraphics[width=0.06\linewidth]{figures/2combined_sample/mnist/7.pdf}} \subfloat{\includegraphics[width=0.06\linewidth]{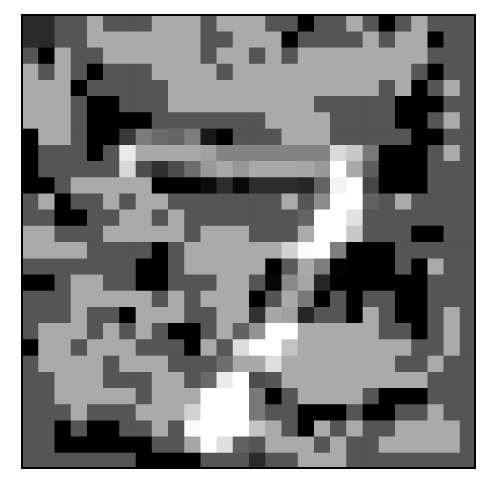}} \\ $L_2$-norm: 10.36\end{tabular}
}
& \multicolumn{1}{c|}{\begin{tabular}[c]{@{}c@{}}
\subfloat{\includegraphics[width=0.06\linewidth]{figures/2combined_sample/mnist/7.pdf}} \subfloat{\includegraphics[width=0.06\linewidth]{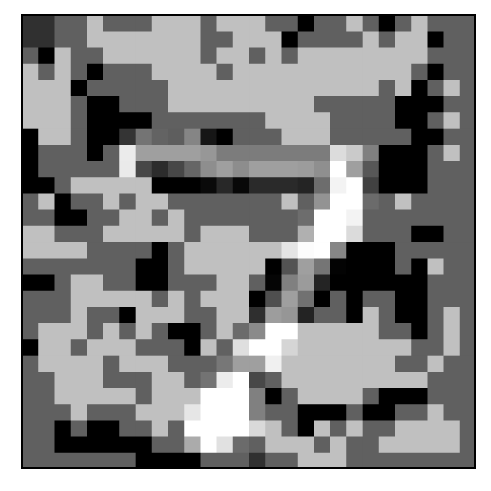}}\\ $L_2$-norm: 12.43\end{tabular}
}
& \multicolumn{1}{c|}{\begin{tabular}[c]{@{}c@{}}
\subfloat{\includegraphics[width=0.06\linewidth]{figures/2combined_sample/mnist/7.pdf}} \subfloat{\includegraphics[width=0.06\linewidth]{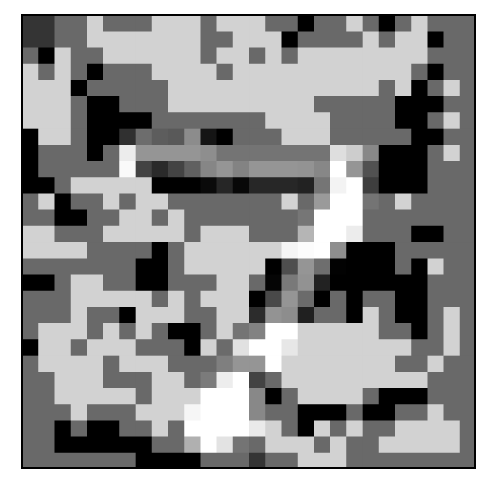}}\\ $L_2$-norm: 14.49\end{tabular}
}
& \multicolumn{1}{c|}{\begin{tabular}[c]{@{}c@{}}
\subfloat{\includegraphics[width=0.06\linewidth]{figures/2combined_sample/mnist/7.pdf}} \subfloat{\includegraphics[width=0.06\linewidth]{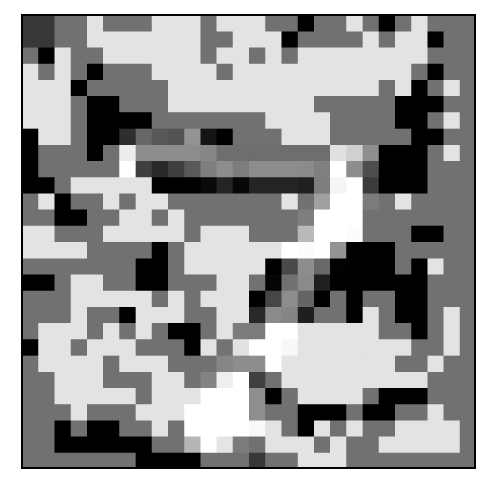}}\\ $L_2$-norm: 16.54\end{tabular}
} \\ \hline
\multicolumn{1}{l}{} & \multicolumn{1}{l}{} & \multicolumn{1}{l}{} & \multicolumn{1}{l}{} & \multicolumn{1}{l}{} & \multicolumn{1}{l}{} \\ \cline{2-6} 
\multicolumn{1}{c|}{} & \multicolumn{1}{c|}{$\eta = 0.02$} & \multicolumn{1}{c|}{$\eta = 0.05$} & \multicolumn{1}{c|}{$\eta = 0.1$} & \multicolumn{1}{c|}{$\eta = 0.2$} & \multicolumn{1}{c|}{$\eta = 0.3$} \\ \hline
\multicolumn{1}{|c|}{\textbf{CIFAR-10}} 
& \multicolumn{1}{c|}{\begin{tabular}[c]{@{}c@{}}
\subfloat{\includegraphics[width=0.06\linewidth]{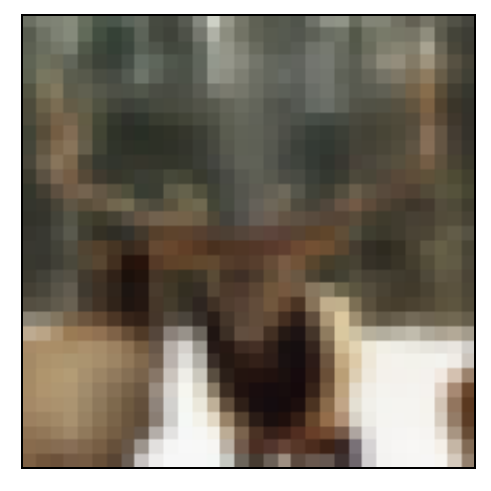}} \subfloat{\includegraphics[width=0.06\linewidth]{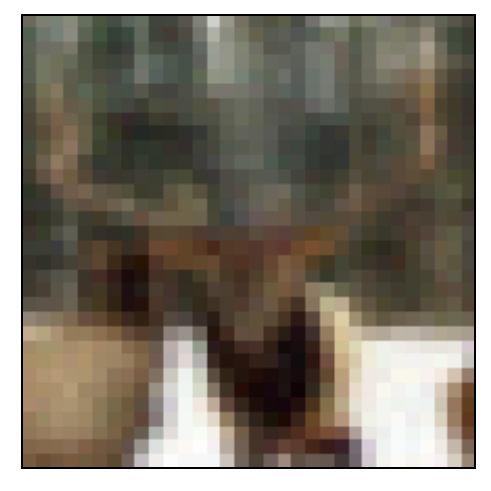}} \\ $L_2$-norm: 0.80\end{tabular}
} 
& \multicolumn{1}{c|}{\begin{tabular}[c]{@{}c@{}}
\subfloat{\includegraphics[width=0.06\linewidth]{figures/cifar10_csf_sample_combined_attack/cifar10_org_data.pdf}} \subfloat{\includegraphics[width=0.06\linewidth]{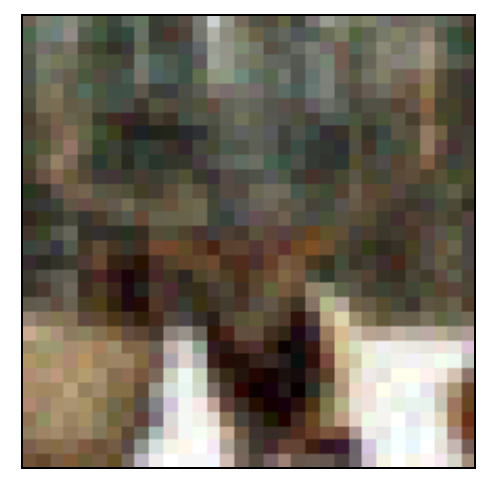}}\\ $L_2$-norm: 2.00\end{tabular}
} 
& \multicolumn{1}{c|}{\begin{tabular}[c]{@{}c@{}}
\subfloat{\includegraphics[width=0.06\linewidth]{figures/cifar10_csf_sample_combined_attack/cifar10_org_data.pdf}} \subfloat{\includegraphics[width=0.06\linewidth]{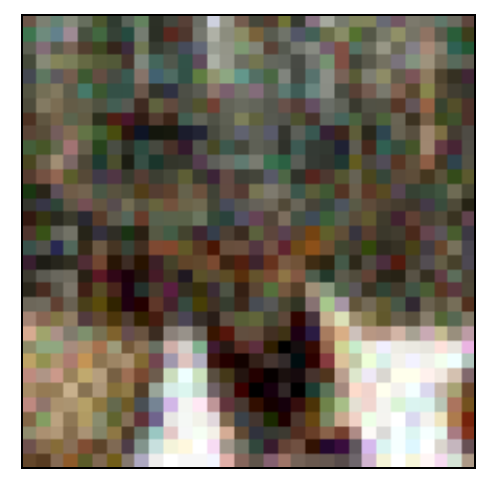}}\\ $L_2$-norm: 4.01\end{tabular}
} 
& \multicolumn{1}{c|}{\begin{tabular}[c]{@{}c@{}}
\subfloat{\includegraphics[width=0.06\linewidth]{figures/cifar10_csf_sample_combined_attack/cifar10_org_data.pdf}} \subfloat{\includegraphics[width=0.06\linewidth]{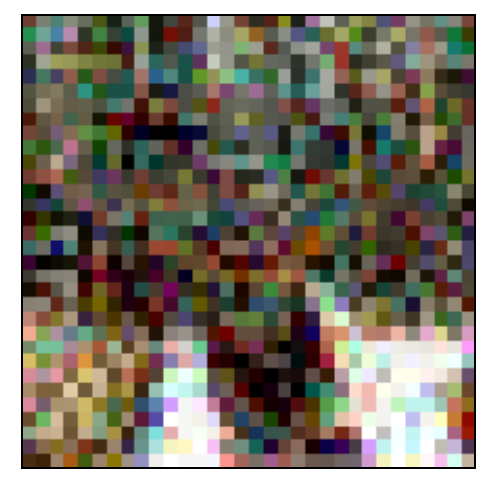}}\\ $L_2$-norm: 8.01\end{tabular}
} 
& \multicolumn{1}{c|}{\begin{tabular}[c]{@{}c@{}}
\subfloat{\includegraphics[width=0.06\linewidth]{figures/cifar10_csf_sample_combined_attack/cifar10_org_data.pdf}} \subfloat{\includegraphics[width=0.06\linewidth]{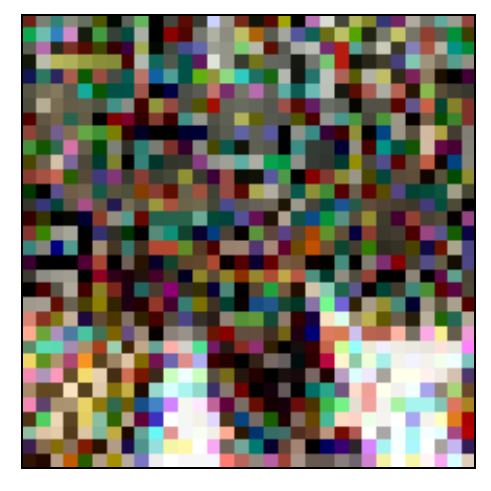}}\\ $L_2$-norm: 12.02\end{tabular}
} \\ \hline
\multicolumn{1}{l}{} & \multicolumn{1}{l}{} & \multicolumn{1}{l}{} & \multicolumn{1}{l}{} & \multicolumn{1}{l}{} & \multicolumn{1}{l}{} \\ \cline{2-6} 
\multicolumn{1}{c|}{} & \multicolumn{1}{c|}{$\eta = 0.02$} & \multicolumn{1}{c|}{$\eta = 0.05$} & \multicolumn{1}{c|}{$\eta = 0.1$} & \multicolumn{1}{c|}{$\eta = 0.2$} & \multicolumn{1}{c|}{$\eta = 0.3$} \\ \hline
\multicolumn{1}{|c|}{\textbf{CIFAR-100}} 
& \multicolumn{1}{c|}{\begin{tabular}[c]{@{}c@{}}
\subfloat{\includegraphics[width=0.06\linewidth]{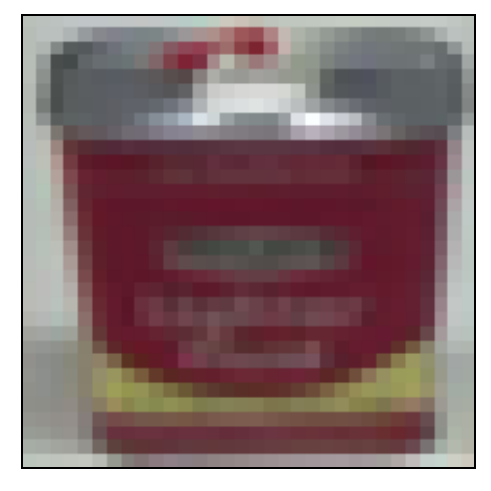}} \subfloat{\includegraphics[width=0.06\linewidth]{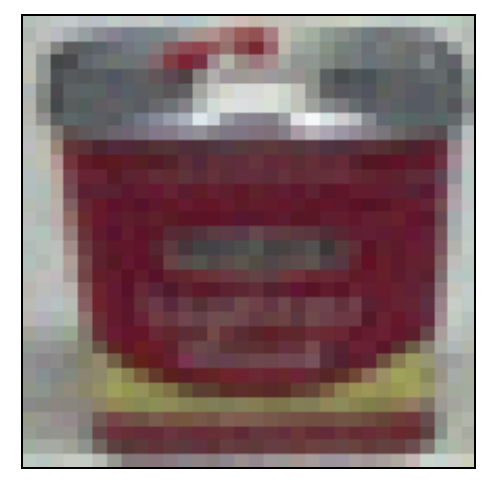}} \\ $L_2$-norm: 0.80\end{tabular}
} 
& \multicolumn{1}{c|}{\begin{tabular}[c]{@{}c@{}}
\subfloat{\includegraphics[width=0.06\linewidth]{figures/cifar100_csf_sample_combined_attack/cifar100_org_data.pdf}} \subfloat{\includegraphics[width=0.06\linewidth]{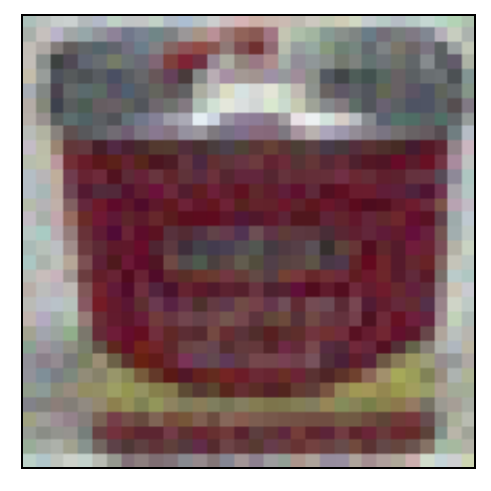}}\\ $L_2$-norm: 2.01\end{tabular}
} 
& \multicolumn{1}{c|}{\begin{tabular}[c]{@{}c@{}}
\subfloat{\includegraphics[width=0.06\linewidth]{figures/cifar100_csf_sample_combined_attack/cifar100_org_data.pdf}} \subfloat{\includegraphics[width=0.06\linewidth]{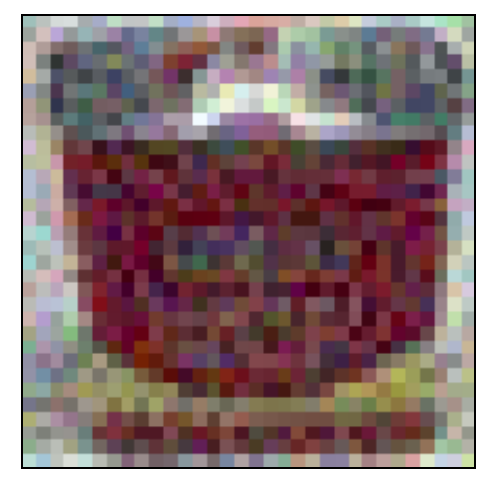}}\\ $L_2$-norm: 4.02\end{tabular}
} 
& \multicolumn{1}{c|}{\begin{tabular}[c]{@{}c@{}}
\subfloat{\includegraphics[width=0.06\linewidth]{figures/cifar100_csf_sample_combined_attack/cifar100_org_data.pdf}} \subfloat{\includegraphics[width=0.06\linewidth]{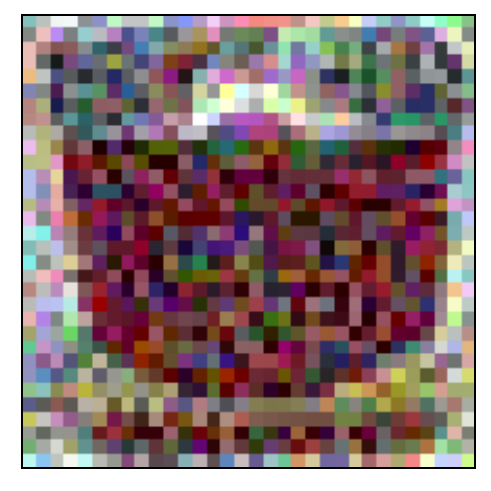}}\\ $L_2$-norm: 8.04\end{tabular}
} 
& \multicolumn{1}{c|}{\begin{tabular}[c]{@{}c@{}}
\subfloat{\includegraphics[width=0.06\linewidth]{figures/cifar100_csf_sample_combined_attack/cifar100_org_data.pdf}} \subfloat{\includegraphics[width=0.06\linewidth]{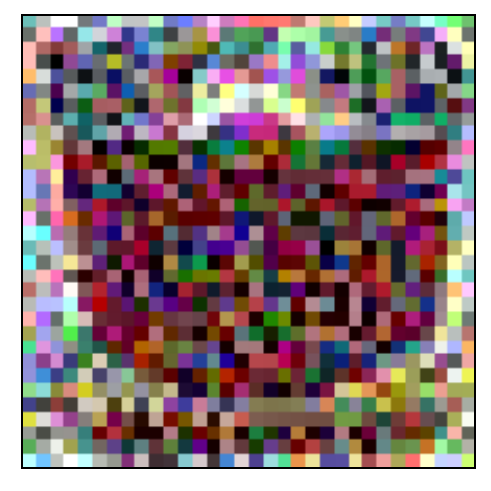}}\\ $L_2$-norm: 12.07\end{tabular}
} \\ \hline
\end{tabular}
\end{table*}

\section{Conclusion}\label{sec:conclusion}\vspace{-0.1cm}
This paper proposes a new ensemble-based solution by constructing defender models with diverse decision boundaries with respect to the original model. The defender models constructed by {\em Split-and-Shuffle} transformation and {\em Contrast-Significant-Features} method reduces the chance of transferring adversarial examples from the original to the defender model targeting the same class. The experimental results show the robustness of the proposed methodology against state-of-the-art adversarial attacks, even in the presence of a stronger adversary targeting both the models within the ensemble simultaneously. Exploration of further techniques to construct machines with diverse decision boundaries adhering to the principle outlined in the paper, and techniques to unify the defender machines can be very interesting future directions of research.\vspace{-0.3cm}
\ifCLASSOPTIONcaptionsoff
  \newpage
\fi



\bibliographystyle{IEEEtran}
\bibliography{IEEEabrv,references.bib}

%



%
\vspace{-1.2cm}
\begin{IEEEbiography}{Manaar Alam} has been pursuing PhD in the Department of Computer Science and Engineering, Indian Institute of Technology Kharagpur, since 2016. His primary research interests include application of Deep Learning in Hardware and Software security, Security Evaluation of Deep Learning Algorithms and Implementations.
\end{IEEEbiography}\vspace{-1.1cm}

\begin{IEEEbiography}{Shubhajit Datta}
has been pursuing PhD in the Centre of Excellence in Artificial Intelligence, Indian Institute of Technology Kharagpur, since 2020. His research interests includes Deep Learning, Computer Vision, Security Evaluation of Deep Learning.
\end{IEEEbiography}\vspace{-1.1cm}

\begin{IEEEbiography}{Debdeep Mukhopadhyay}
received his PhD from the Department of Computer Science and Engineering, Indian Institute of Technology Kharagpur in 2007, where he is presently a Professor. His research interests include Cryptography, VLSI of Cryptographic Algorithms, Hardware Security and Side-Channel Analysis. He is a senior member of the ACM and IEEE.
\end{IEEEbiography}\vspace{-1.1cm}

\begin{IEEEbiography}{Arijit Mondal}
received his PhD from the Department of Computer Science and Engineering, Indian Institute of Technology Kharagpur in 2010. Currently, he is working as an Assistant Professor in Center of Excellence in Artificial Intelligence, Indian Institute of Technology Kharagpur. His research interests include Embedded Control Systems, CAD for VLSI, Smart Grid, and Deep Learning. He is a member of the IEEE.
\end{IEEEbiography}\vspace{-1.1cm}

\begin{IEEEbiography}{Partha Pratim Chakrabarti}
received his PhD from the Department of Computer Science and Engineering, Indian Institute of Technology Kharagpur, in 1988, where he is presently a Professor. He is also associated with the Centre of Excellence in Artificial Intelligence, Indian Institute of Technology, Kharagpur. His areas of interest include AI, CAD for VLSI, Embedded Systems, Algorithm Design, and Reliable and Fault Tolerant Systems.
\end{IEEEbiography}







\end{document}